\pdfoutput=1
\documentclass[11pt]{article}
\usepackage[margin=1in]{geometry}
\usepackage[T1]{fontenc}
\usepackage{newtxtext}
\usepackage{amsmath,amssymb,amsfonts,amsthm}
\usepackage{natbib}
\usepackage{booktabs}
\usepackage{enumitem}
\usepackage{graphicx}
\usepackage{url}
\usepackage{xcolor}
\usepackage{tikz}
\usetikzlibrary{arrows.meta,decorations.pathreplacing}
\usepackage[hidelinks,pdftitle={What Does a Copyright Budget Certify?},pdfauthor={J. Vijayavallabh}]{hyperref}

\newtheorem{proposition}{Proposition}
\newtheorem{lemma}{Lemma}
\newtheorem{remark}{Remark}
\graphicspath{{figures/}}

\newcommand{\codeurl}{\url{https://github.com/Vijayavallabh/DA5001_Project}}

\title{What Does a Copyright Budget Certify?\\
An Exact Threshold for Per-Token Meters, and a Length-Independent Certificate for Selection}

\author{J. Vi{}jayavallabh}
\date{}

\begin{document}
\maketitle

\begin{abstract}
Certified inference-time copyright defences publish a divergence budget between the served model and a
safe \emph{anchor} trained without the protected works. We ask what such a budget certifies.
\emph{Proved:} a budget of $K$ nats says nothing about a work whose surprisal under the anchor is at
most $K$, at every R\'enyi order. Anchored decoding charges per token, so its budget grows with the
output. \emph{Measured:} at its authors' setting, the budget exceeds the anchor's surprisal of every one of
the $758$ protected passages we test, so it certifies nothing about them.
\emph{Proved:} serving one of $n$ anchor draws, by any rule, is certified at $\log n$ nats for the whole
output, at any length; in installments, a window is certified only given the text up to a block boundary.
\emph{Measured:} over a vetted anchor, with an adversarial scorer, selection reproduced no protected
passage near-verbatim. Its draws cost $21.8\times$ a metered decode on an anchor-only server.
\emph{Judged:} in the headline configuration, at the only published meter budget that bounds a
$50$-token window, selection's gain over one anchor draw is reliably positive and the meter's is, by a
post-hoc equivalence test, zero. At matched compute selection loses. At the authors' sampling settings on
the $8$B, the two tie at that budget. Where the meter's certificate is vacuous, the ranking depends on how
the stronger model is served. It reverses under a chat template, at the authors' sampling settings on the
$8$B, and on AlpacaEval.
\end{abstract}

\section{Introduction}\label{sec:intro}

Language models memorise their training data. \texttt{Llama-3.1-70B}, for example, reproduces
\emph{Harry Potter and the Sorcerer's Stone} almost verbatim from its first few words
\citep{cooper2025extracting}. Retraining that memory away is impractical against a stream of takedown
requests. Certified decoders need no retraining. They serve a law $q$ that stays within a divergence
budget $K$ of a \emph{safe} model $p_s$, the \emph{anchor}, trained without the protected work. This
property is $k$-near-access-freeness \citep{vyas2023provable}. The deployed instance, anchored decoding
\citep{he2026anchored}, tilts each token from the anchor toward a stronger \emph{risky} model $p_r$. It
charges the divergence per token at rate $k$, so a $T$-token response is certified at $K = kT$.

This paper asks what such a published budget certifies. We study \emph{certified} inference-time defences
only. Heuristic filters such as $n$-gram blocklists, and training-time methods such as unlearning, are
different tools, which Section~\ref{sec:related} places.

\begin{figure}[t]
\centering
\begin{tikzpicture}[x=1cm, y=1cm, font=\sffamily\scriptsize,
  tok/.style={draw=black!55, fill=white, minimum width=0.62cm, minimum height=0.42cm, inner sep=0pt,
              rounded corners=1pt},
  mdl/.style={circle, draw=black!55, minimum size=0.6cm, inner sep=0pt},
  arr/.style={-{Stealth[length=1.5mm]}, black!55, line width=0.5pt}]
\node[anchor=west, font=\sffamily\footnotesize, inner sep=0pt] at (0,2.62)
  {\textcolor{red!55!black}{\bfseries (a) budget spent per token}\enspace\textcolor{black!55}{metered decoding}};
\node[mdl, fill=blue!12] (ps) at (0.35,1.80) {$p_s$};
\node[mdl, fill=red!12] (pr) at (0.35,0.72) {$p_r$};
\node[draw=black!55, rounded corners=1pt, fill=black!4, inner sep=2pt] (mix) at (1.42,1.26) {$p_{\theta_t}$};
\draw[arr] (ps) -- (mix); \draw[arr] (pr) -- (mix);
\foreach \i/\lab in {1/$y_1$,2/$y_2$,3/$y_3$,4/$y_4$} {
  \node[tok] (t\i) at ({1.62+\i*0.72},1.26) {\lab};
}
\node at ({1.62+5*0.72},1.26) {$\cdots$};
\node[tok] (tT) at ({1.62+6*0.72},1.26) {$y_T$};
\draw[arr] (mix) -- (t1);
\foreach \i/\h in {1/0.30,2/0.50,3/0.22,4/0.44,6/0.36} {
  \fill[red!55!black, opacity=0.75] ({1.62+\i*0.72-0.13},1.53) rectangle ++(0.26,\h);
}
\node[text=red!55!black] at (1.15,2.16) {charge $\le$ bucket};
\draw[decorate, decoration={brace, mirror, amplitude=3pt}, black!55] (2.02,0.98) -- (6.24,0.98);
\node[anchor=north] at (3.95,0.86) {certificate $K = kT$ nats: grows with the output};
\node[anchor=north, text=red!55!black] at (3.95,0.50) {vacuous for a work $x$ once $k \ge s(x)$, at any length};
\begin{scope}[xshift=7.45cm]
\node[anchor=west, font=\sffamily\footnotesize, inner sep=0pt] at (0,2.62)
  {\textcolor{blue!55!black}{\bfseries (b) budget spent once, on the draw}\enspace\textcolor{black!55}{ours}};
\node[mdl, fill=blue!12] (qs) at (0.35,1.26) {$p_s$};
\foreach \j/\yy/\lab in {1/2.02/$y^{(1)}$,2/1.60/$y^{(2)}$,3/0.72/$y^{(n)}$} {
  \node[draw=black!55, fill=blue!8, rounded corners=1pt, minimum width=1.8cm, minimum height=0.30cm,
        inner sep=0pt] (d\j) at (2.05,\yy) {\lab};
  \draw[arr] (qs) -- (d\j.west);
}
\node at (2.05,1.20) {$\vdots$};
\node[draw=black!55, rounded corners=1pt, fill=black!4, align=center, inner sep=2pt] (f) at (4.15,1.37)
  {any score $f$\\[-1pt]{\tiny reward, vote, adversary}};
\foreach \j in {1,2,3} { \draw[arr] (d\j.east) -- (f.west); }
\node[draw=blue!55!black, rounded corners=1pt, fill=blue!12, align=center, inner sep=2pt] (o) at (5.95,1.37)
  {serve\\[-1pt]one draw};
\draw[arr] (f) -- (o);
\draw[decorate, decoration={brace, mirror, amplitude=3pt}, black!55] (1.10,0.49) -- (4.85,0.49);
\node[anchor=north] at (2.97,0.37) {certificate $q(y) \le n\,p_s(y)$: $\log n$ nats, for any $f$};
\node[anchor=north, text=blue!55!black] at (2.97,0.01) {$n=8$: $2.08$ nats at any length};
\end{scope}
\end{tikzpicture}\\[3pt]
\includegraphics[width=0.8\textwidth]{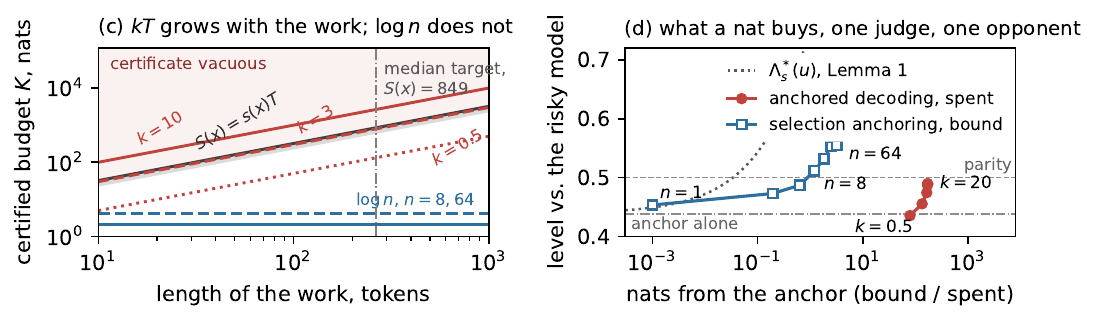}
\caption{\textbf{Where the budget is spent decides what it certifies.} \textbf{(a)} A meter tilts each
token from the anchor $p_s$ toward the risky model $p_r$ and charges a bucket. \textbf{(b)} Selection draws
$n$ outputs from $p_s$ and serves one. Only the choice is charged. \textbf{(c)} For a work served at its
own length, the meter's $K = kT$ (red) and the work's price $S(x) = s(x)T$ (grey, nine pairs) are both
rays. Vacuity ($K \ge S(x)$, shaded) is decided by $k/s(x)$. Selection's $\log n$ (blue) is flat.
\textbf{(d)} One judge pass: level against the risky model's own draw ($0.5$ is parity), against the
divergence the meter \emph{spent} (filled) or selection is \emph{certified} not to exceed (open, a KL bound;
Appendix~\ref{app:realisedkl}). Arms at $0$ nats drawn at $10^{-3}$. Dotted: Lemma~\ref{thm:nfl}'s frontier.}
\label{fig:selection}
\end{figure}

\paragraph{Two answers.} First, a budget of $K$ nats says nothing about a work whose surprisal under the
anchor is at most $K$, at every R\'enyi order (Proposition~\ref{prop:threshold}). A per-token budget grows
with the output, so at the authors' setting it is vacuous for every protected passage we test. Second, a
budget need not grow. Draw $n$ completions from the anchor, score them with anything, and serve one
(Figure~\ref{fig:selection}b). This \emph{selection} rule is certified at $\log n$ nats for the whole
output, at any length and for any scorer (Proposition~\ref{prop:selection}). It buys a choice among the
anchor's own draws, whereas a meter buys the risky model's tokens and pays for each one it imitates.

\paragraph{The deployment regime.} Selection never runs the risky model, so it is not a substitute for
serving a frontier model. It is a certified way to improve a model the deployer may already serve: a
licensed, vetted anchor. The binding constraint is an anchor of sufficient quality that passes vetting. Of
the openly licensed anchors we test, the largest to pass is Comma-7B, far below frontier scale. We do not
know where a frontier-quality vetted anchor would come from. No $n$ helps an anchor that cannot do the task
(Section~\ref{sec:cost}).

\paragraph{Contributions.} We keep three kinds of claim apart. \textsc{Proved} claims are theorems.
\textsc{Measured} claims are leakage and cost. \textsc{Judged} claims are utility as scored by
language-model judges.
\begin{itemize}[leftmargin=1.2em,itemsep=2pt,topsep=2pt,parsep=0pt]
\item \textbf{\textsc{Proved}} (Sections~\ref{sec:method}--\ref{sec:frontier}). The vacuity threshold
(Proposition~\ref{prop:threshold}); the $\log n$ certificate for any selection rule, its exact composition
over $m$ queries, and its corrected form in installments (Propositions~\ref{prop:selection}
and~\ref{prop:installments}); and two chain-rule limits on per-token meters
(Propositions~\ref{prop:sparse} and~\ref{prop:imitation}). \emph{Borrowed vs new:} the first two
propositions are elementary \citep{vanerven2014renyi,beirami2025bestofn}. New are reading deployed
budgets against the threshold, and the observation that any selection rule is a certified near-access-free
mechanism, auditable by sampling the anchor.
\item \textbf{\textsc{Measured}} (Sections~\ref{sec:leakage} and~\ref{sec:cost}). Anchored decoding
leaks only where its certificate is void. Over a vetted anchor, selection with an adversarial scorer
reproduced no protected passage, but it multiplies what a contaminated anchor leaks, and its draws are
costly. \emph{New:} these measurements.
\item \textbf{\textsc{Judged}} (Section~\ref{sec:utility}). At the one published meter budget whose
certificate bounds a $50$-token window, selection's gain over one anchor draw is reliably positive and the
meter's is equivalent to zero (post hoc). At a matched whole-output certificate, selection and its
installments lead the two meter variants we could match. Selection ties at the authors' sampling settings on the $8$B and
loses at matched compute. Where the meter's certificate is vacuous, the ranking depends on how the risky
model is served, and it reverses in three settings; a windowed meter whose certificate is too large to
audit also wins under the chat template. \emph{New:} the comparison; the judging protocol is
standard.
\end{itemize}

\paragraph{Correction note.} An earlier draft of this work stated the installment certificate for a
window conditional on all of the text before it. That is false for a window that starts inside a block,
as a reviewer of that draft showed. Proposition~\ref{prop:installments} gives the corrected,
boundary-conditioned statement, and every comparison that used it now names the certificate type.

\section{What a budget certifies}\label{sec:method}\label{sec:selection}

This section is \textsc{proved}. It states two elementary propositions and reads deployed budgets against
them; Table~\ref{tab:certificate} evaluates them at measured surprisals.

\paragraph{Setup.} Fix a prompt. The anchor $p_s$ is trained without the protected work; $p_r$ is the
risky model a deployer would rather serve. A defence serves a law $q$ and certifies
$D(q\,\|\,p_s) \le K$ for some divergence $D$. For a protected work $x$, $S(x) = -\log p_s(x)$ is its
surprisal under the anchor and $s(x)$ that surprisal per token (appendix Table~\ref{tab:notation}). We protect
CopyBench \emph{passages} \citep{chen2024copybench}: $64$ tokens, median $204.8$ nats over the $758$ we
test. A \emph{window} is $50$ protected tokens given the true text before them, median $S_w = 159.8$ nats.
An extraction \emph{target} is a passage's text after a $20$-token seed, median $849$ nats.

\subsection{The vacuity threshold}

\begin{proposition}[The vacuity threshold does not depend on the order]\label{prop:threshold}
Let $E$ be an event with $\Pr_{p_s}[E] = e^{-S}$. For $\alpha \in [1,\infty]$ let
$b_\alpha(K) = \sup\{\Pr_q[E] : D_\alpha(q\,\|\,p_s) \le K\}$. This is the most that a budget of $K$ nats
at order $\alpha$ can say about $E$. Then $b_\alpha(K) = \max\{a : d_\alpha(a\,\|\,e^{-S}) \le K\}$, where
$d_\alpha$ is the R\'enyi divergence between Bernoulli laws. Moreover $b_\alpha(K) = 1$ if and only if
$K \ge S$, at every order.
\end{proposition}

The proof is data processing, a law that reweights $p_s$ uniformly on $E$ and its complement, and
$d_\alpha(1\,\|\,e^{-S}) = S$ (Appendix~\ref{app:proofs}). Below the threshold the orders differ. The
pathwise ($\alpha=\infty$) bound is $e^{K-S}$, while the KL bound falls only as about $K/S$: a KL budget
of half a work's surprisal still permits the work half the time.

Length makes this worse for a per-token meter. A work served at its own length has $K = kT$ and
$S(x) = s(x)T$, so the KL bound, about $k/s(x)$, does not improve with length. A $w$-token window inside a
$T_{\max}$-token response is unprotected once $k \ge s(x)\,w/T_{\max}$: a quarter of $s(x)$ for $50$
tokens at $T_{\max} = 200$ (appendix Table~\ref{tab:windowvacuity}). \emph{What this adds:} the identity is known
\citep{vanerven2014renyi,vyas2023provable}; what is new is reading \emph{deployed} budgets against it,
which is why we suggest publishing $k/s(x)$ rather than $k$.

\subsection{A budget that does not grow: selection}

A \emph{selection rule} draws $y_1,\dots,y_n \sim p_s$ i.i.d., scores them with any function $f$, and
serves one. Every served token was sampled from $p_s$; only the choice of draw is charged.

\begin{proposition}[The budget of a selection rule]\label{prop:selection}
Let $y_1,\dots,y_n$ be i.i.d.\ $p_s$, and let $q$ be the law of \emph{any} rule that serves one of them.
Then $q(y) \le n\,p_s(y)$ for every $y$. Hence $D_\infty(q \,\|\, p_s) \le \log n$, so
$D_\alpha(q\,\|\,p_s) \le \log n$ for every $\alpha \in [1,\infty]$, and $\Pr_q[E] \le n \Pr_{p_s}[E]$
for every event $E$. If the rule is the argmax of a score with no ties, the KL is smaller still:
$D_{\mathrm{KL}}(q \,\|\, p_s) \le \log n - (n-1)/n$.
\end{proposition}

\begin{proof}
If $y$ is served, then $y$ was drawn. So
$q(y) \le \Pr[\exists i : y_i = y] \le \sum_i \Pr[y_i = y] = n\,p_s(y)$. Taking a logarithm and a maximum
over $y$ gives the $\alpha = \infty$ statement, which dominates every lower order. Summing over $y \in E$
gives the event form. Only the union bound is used, so the rule may consult the scores, the strings,
external randomness or nothing. The KL form is due to \citet{beirami2025bestofn}.
\end{proof}

The bound is tight, survives post-processing and early stopping, and covers rules with no score, such as
self-consistency \citep{wang2023selfconsistency} (Appendix~\ref{app:certdetail}). It is \emph{relative}
to the anchor as sampled, so it is only as good as the anchor: self-consistency over a web-trained model
satisfies it and certifies nothing about copyright. \emph{What this adds:} the inequality and its KL form
are known. What is new is what it certifies. Any selection rule over a vetted anchor is a certified
near-access-free mechanism, whatever its scorer, with a pathwise budget independent of output length.

\paragraph{One response, one window.} Table~\ref{tab:certificate} reads both propositions against the
median window. A meter at the authors' $k=3$, or at $k=10$, certifies nothing about it. At $k=0.5$ it
bounds the window by $0.630$. Selection at $n=64$ bounds it by about $10^{-67.6}$. This closed form prices
the exact window at the start of the response. The window at any later position, or after served text, is
priced by the same joint bound but has no closed form (Appendix~\ref{app:certdetail}).

\begin{table}[t]
\centering\footnotesize
\caption{\textbf{What one response's certificate says about a protected $50$-token window}
($S_w = 159.8$ nats, median over $758$ passages), priced at the start of the response with the work's
true preceding text as the prompt. \emph{Meter}: KL charge $K = kT_{\max}$ at $T_{\max} = 200$ (measured
mean spend in brackets); bound from Proposition~\ref{prop:threshold}, where $1$ excludes nothing.
\emph{Selection}: pathwise charge $\log n$, bound $n e^{-S_w}$. \emph{queries}: adaptive queries whose
composed certificate stays below $S_w$, which matters only for a colluding scorer. \emph{draws}: clean
anchor draws that certify near-verbatim reproduction of one work below $1\%$ per response by the $3n/N$
audit (one prompt, $95\%$, zero hits; Section~\ref{sec:compose}).}
\label{tab:certificate}
\setlength{\tabcolsep}{4.5pt}
\begin{tabular}{@{}llrrr@{\hspace{14pt}}rr@{}}
\toprule
mechanism & charge & $K$ (spent), nats & $K/S_w$ & $\Pr_q[\text{window}] \le$ & queries & draws \\
\midrule
meter, $k=0.5$ & KL & $100$ ($78.5$) & $0.63$ & $0.630$ & $1$ & --- \\
meter, $k=3$ (authors') & KL & $600$ ($165.0$) & $3.75$ & $1$ & $0$ & --- \\
meter, $k=10$ & KL & $2000$ ($171.3$) & $12.5$ & $1$ & $0$ & --- \\
\midrule
selection, $n=8$ & pathwise & $2.08$ & $0.013$ & $10^{-68.5}$ & $76$ & $2{,}400$ \\
selection, $n=64$ & pathwise & $4.16$ & $0.026$ & $10^{-67.6}$ & $38$ & $19{,}200$ \\
\bottomrule
\end{tabular}
\end{table}

\section{What a certificate covers across queries, and how it is audited}\label{sec:compose}

\paragraph{Composition (\textsc{proved}).} Let a user issue $m$ queries, each answered by a selection rule
over $n$ fresh draws and each chosen in light of everything served so far. Proposition~\ref{prop:selection}
applies round by round, so the transcript satisfies $D_\infty(Q\,\|\,P_s) \le m\log n$ exactly, with no
slack from adaptivity (Appendix~\ref{app:compose}). Per-query access-freeness does not compose like this
in general \citep{cohen2025blameless}, and a per-token meter composes to $mkT$. That a single response
reproduces the window has probability at most $m\,n\,e^{-S_w}$, below $10^{-60}$ after $10^4$ queries at
$n=64$. A transcript that carries the window piecewise is bounded only by $m\log n$, which stays below
$S_w$ for $38$ queries.

\paragraph{The $3n/N$ audit and its scope (\textsc{measured}).} A near-verbatim variant of a window has no
closed-form probability, so a rights-holder must sample the anchor, as privacy is audited
\citep{jagielski2020auditing}. If $N$ clean anchor draws show the event zero times, then
$\Pr_{p_s}[E] \le 3/N$ at $95\%$ confidence, and selection certifies $\Pr_q[E] \le 3n/N$ per response. The
scope is narrow. The bound covers one fixed prompt distribution (the work's own prefix, at a stated length
and temperature), holds at $95\%$ confidence, and assumes zero observed hits. Extending it to other prompts
requires auditing those prompts, for instance with a union bound over a finite family fixed in advance.
Pooled adaptive queries are covered by the $m\log n$ composition, not by the audit
(Appendix~\ref{app:auditscope}). No feasible sample audits a KL certificate of $100$ nats or more.

\paragraph{The scorer (\textsc{proved}, plus an operational condition).} A scorer that colludes with the
user can spend the whole composed bound, for instance by steganography (Appendix~\ref{app:colluding}, \emph{A colluding scorer}). A deployer who cannot vet the scorer must therefore cap the queries of
every set of users who could pool transcripts below $S/\log n$ for the shortest protected event. That is $38$ queries at $n=64$ for a window, for the
whole deployment, a cap few public services would accept. A
\emph{vetted} scorer, one that neither holds the work nor takes the user's instructions, removes the need
for the cap in practice. This is an operational condition that \emph{the certificate cannot verify}:
formally the transcript bound stays $m\log n$. Our headline scorer, \texttt{Qwen2.5-7B-Instruct}, is not
vetted; a vetted one gains little (Section~\ref{sec:breadth}).

\section{What a per-token meter can certify}\label{sec:frontier}

This section is \textsc{proved}, with the measurements that make each result bite marked. It asks
whether a per-token meter could carry a budget as small as selection's, one that does not grow with the
output.

\begin{proposition}[A bounded budget is sparse]\label{prop:sparse}
Let $q$ be any causal policy, $q(y) = \prod_{t \le T} q_t(y_t \mid y_{<t})$, with sequence budget
$D_{\mathrm{KL}}(q \,\|\, p_s) \le K$, and write
$N_\varepsilon = \#\{t \le T : D_{\mathrm{KL}}(q_t(\cdot \mid y_{<t}) \,\|\, p_{s,t}) > \varepsilon\}$.
Then $\mathbb{E}_q[N_\varepsilon] \le K/\varepsilon$ for every $\varepsilon > 0$.
\end{proposition}

The proof is the chain rule and Markov's inequality (Appendix~\ref{app:proofs}). A per-token allowance
therefore has two options. If it grows with the work, $K = kT$, it is vacuous from $k = s(x)$ on. If it
stays bounded, the policy is the anchor at all but $O(K/\varepsilon)$ steps. Selection's served law
obeys this bound too. Its budget is spent on one choice among complete drafts, which a per-token decode loop
never sees.

\begin{lemma}[The price of utility]\label{thm:nfl}
Let $q$ be \emph{any} law with $D_{\mathrm{KL}}(q \,\|\, p_s) \le K$. For every bounded utility $U$,
\begin{equation}
K \;\ge\; \Lambda^*_s\!\bigl(\mathbb{E}_q[U]\bigr), \qquad
\Lambda^*_s(u) = \sup_{\lambda}\;\bigl\{\lambda u - \log \mathbb{E}_{p_s}[e^{\lambda U}]\bigr\},
\label{eq:nfl}
\end{equation}
the Cram\'er rate function of $U$ under the safe model.
\end{lemma}

This is the Donsker--Varadhan variational principle \citep{donsker1975asymptotic}, and we claim no
novelty for it. Reproducing a work costs $S(x)$, which grows with its length. Buying mean utility $u$
costs $\Lambda^*_s(u)$, which is $O(1)$ at any length for a utility the anchor itself sometimes attains.
So there is room for a mechanism that is both safe and useful. A meter that tilts toward $p_r$ does not
occupy it.

\begin{proposition}[Where its bucket is slack, a meter pays the imitation cost]\label{prop:imitation}
Let a bucket-metered decoder serve, at each step, the largest tilt $p_{\theta_t} \propto
p_s^{1-\theta_t}p_r^{\theta_t}$ whose KL charge fits its bucket. Call a step \emph{slack} when
$\theta_t = 1$, and let $I_T$ be the sum of $D_{\mathrm{KL}}(p_{r,t}\,\|\,p_{s,t})$ over the slack steps.
\begin{enumerate}[label=(\roman*),leftmargin=*,itemsep=1pt,topsep=2pt]
\item $D_{\mathrm{KL}}(q\,\|\,p_s) \ge \mathbb{E}_q[I_T]$, whatever utility the decoder achieves.
\item Suppose $\mathbb{E}_q[I_T] \ge \gamma T$. Let $U$ be a bounded utility that the decoder does not
lower, $\mathbb{E}_q[U] \ge \mathbb{E}_{p_s}[U]$, and suppose $U$ attains its maximum under $p_s$ with
probability at least $\pi \in (0,1)$. Then
$D_{\mathrm{KL}}(q\,\|\,p_s) \ge \Lambda^*_s(\mathbb{E}_q[U])\cdot\gamma T/\log(1/\pi)$.
\end{enumerate}
The proviso on $\pi$ is substantive. Part (ii) separates the meter from Lemma~\ref{thm:nfl}'s frontier by a
factor growing in $T$ only when $\pi$ does not shrink with $T$, that is, when the anchor sometimes attains
$U$'s maximum on its own. For a utility attained only by one exact long string, $\pi = e^{-\Theta(T)}$, the
factor $\gamma T/\log(1/\pi)$ is $\Theta(1)$, and no separation follows.
\end{proposition}

\paragraph{Why Propositions~\ref{prop:sparse} and~\ref{prop:imitation} matter.} Both are short
consequences of the chain rule; their content is what they rule out. Proposition~\ref{prop:sparse} shows
that rescheduling cannot rescue a bounded budget. Proposition~\ref{prop:imitation} shows that a meter
which imitates $p_r$ is charged $\gamma T$ whatever utility the imitation buys. A measurement makes this
bite: on ordinary text $\gamma = 0.857$ nats per token, about $40$ nats for a $50$-token window. Neither
result says that selection's output law is out of reach of a per-token policy, since every law over
sequences factorises token by token. The difference is in what each step may condition on: a meter's step
sees only the prefix, while selection's choice sees whole drafts (Appendix~\ref{app:noseparation}).

\begin{figure}[t]
\centering
\includegraphics[width=0.72\textwidth]{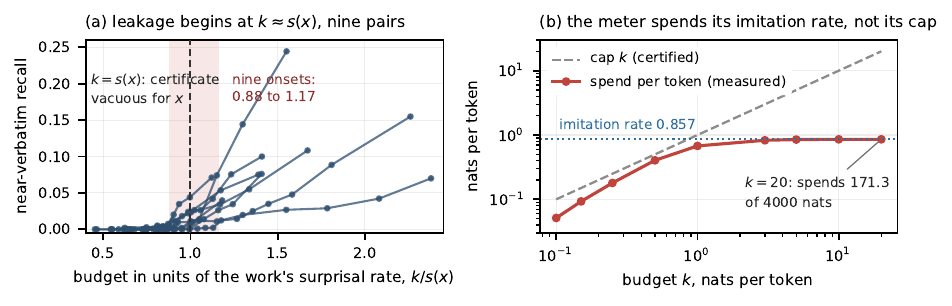}
\caption{\textbf{Where the deployed meter leaks, and what it spends.} \textbf{(a)} Near-verbatim recall of held-out
protected passages against the budget in units of the anchor's surprisal rate, nine (anchor, memoriser)
pairs with nine distinct anchors and six tokenizers. Extraction begins between $0.88$ and $1.17$ of
$s(x)$ (shaded: their range, not an interval), near Proposition~\ref{prop:threshold}'s threshold for the
whole target. Each sweep decodes $T$ tokens, the length of its longest target ($296$ at this anchor). The
certificate is therefore void for the median target from $0.84$ to $0.89$ of $s(x)$, and for a $50$-token
window from $50/T$, that is $0.08$ to $0.17$ of $s(x)$. Wherever extraction occurred, the certificate
excluded nothing.
\textbf{(b)} Anchored decoding at its authors' anchor on ordinary prompts, $16{,}500$ trajectories: the
cap it certifies (dashed) against what it spends (red). Past the imitation rate the
bucket never binds and the spend stays at $171.3$ nats while the cap grows to $4000$.}
\label{fig:horns}
\end{figure}

\paragraph{The deployed meter shows both options (\textsc{measured}; Figure~\ref{fig:horns}).} On nine
anchor--memoriser pairs, extraction begins between $0.88$ and $1.17$ of $s(x)$, where the certificate is
already void. The onset level is robust, though the order of the pairs is not, and two mechanistic
predictions of the onset failed (Appendix~\ref{app:onset}). Where the budget does certify something, the
meter binds and is judged indistinguishable from its own anchor. The share of steps its bucket constrains
falls from $0.954$ at $k=0.1$ to $0.0000$ at $k=20$, where it serves the risky model unchanged. No
per-trajectory violation occurs in over $100{,}000$ budgeted trajectories: the gap is in how $k$ is read,
not in the accounting. Every repair we tried (a sharper charge, a lower cap, a better schedule, a budget
granted up front) either lands on one of the two options or falls short of selection
(Appendix~\ref{app:metertail}).

\paragraph{What the dichotomy does not cover, built and measured.}\label{sec:uncovered}
Two mechanisms escape Proposition~\ref{prop:sparse} by certifying windows instead of the whole
sequence. Their window certificates are of different types.

The first is a \emph{windowed meter} ($w=50$; Appendix~\ref{app:windowed}). It caps each step's
max-divergence so that every span of at most $w$ tokens is certified at $W$ \emph{conditional on all
served text before it}; we mark this type $^{c}$. It covers near-verbatim variants only up to a factor
$e^{W}$, which an audit can check only with more than $3e^{W}$ draws.

The second is selection spent in \emph{installments}: $n$ draws per block of $L$ tokens. An earlier draft
stated its window certificate conditional on the full preceding text, which is false inside a block. The
correct statement follows (proof in Appendix~\ref{app:installmentsproof}).

\begin{proposition}[Selection in installments]\label{prop:installments}
Emit the output in consecutive blocks of $L$ tokens (the last may be shorter, or end at end-of-text), so
block boundaries sit at multiples of $L$. Let $h_{<j}$ be the served text before block $j$, and let
$B = \lceil T_{\max}/L\rceil$ be the maximum number of blocks. At block $j$, draw $n$ continuations i.i.d.\ from $p_s(\cdot \mid h_{<j})$, using fresh
draws independent of all earlier internal randomness. Serve one of them by any rule; the rule may use
scores, rollouts, earlier unserved candidates or external randomness.
\begin{enumerate}[label=(\roman*),leftmargin=*,itemsep=1pt,topsep=2pt]
\item $q(y) \le n^{B}p_s(y)$ for every output $y$, so $D_\infty(q\,\|\,p_s) \le B\log n$. (Indeed
$q(y) \le n^{B(y)}p_s(y)$, with $B(y)$ the number of blocks $y$ actually uses.)
\item For every $j$, every $r \ge 1$, every served history $h_{<j}$ with $\Pr_q[h_{<j}] > 0$, and every event $E$ determined by
blocks $j,\dots,j+r-1$: $\Pr_q[E \mid h_{<j}] \le n^{r}\Pr_{p_s}[E \mid h_{<j}]$.
\item Hence a $w$-token window is certified at $\min(\lceil w/L\rceil + 1, B)\log n$
\emph{conditional on the served text up to the last block boundary at or before the window's start}. A
window that starts exactly on a block boundary needs only $\min(\lceil w/L\rceil, B)\log n$. We mark this type
$^{b}$.
\item No bound of the form $c(n,L,w)$ holds conditional on the served tokens between that boundary and the
window's start: the ratio can be unbounded (Remark~\ref{rem:installments}).
\end{enumerate}
\end{proposition}

\begin{remark}[Conditioning inside a block]\label{rem:installments}
Take one block of $L=2$ binary tokens and $n=2$, with $p_s(00)=1-a$, $p_s(10)=a(1-b)$, $p_s(11)=ab$, and
serve by the preference $11 > 00 > 10$. Then $q(10) = [a(1-b)]^2$ and $q(11) = 1-(1-ab)^2$, and the whole
output satisfies $q(y) \le 2\,p_s(y)$. At $a=10^{-6}$ and $b=10^{-3}$, however,
$\Pr[y_2 = 1 \mid y_1 = 1]$ is $0.001$ under $p_s$ and about $0.999501$ under $q$, a ratio of about
$999.5$. The earlier statement, $(\lceil w/L\rceil + 1)\log n$ given the preceding text, would allow a
factor of $4$ for this one-token window. As $a \to 0$ the ratio approaches $1/b$, and $b$ can be taken
arbitrarily small, so no bound of the form $c(n,L,w)$ holds. The mechanism is the one Appendix~\ref{app:windowed} notes for any
joint bound: it says nothing about a span after an unlikely prefix.
\end{remark}

The two window types are not directly comparable. A $^{c}$ certificate holds whatever the served prefix
is. A $^{b}$ certificate holds only given the text up to a block boundary. A comparison between the
windowed meter and installments at the same nominal nats per window is therefore a comparison
\emph{across certificate types}, not at a matched certificate. Comparisons at a matched whole-output
certificate, $B\log n$, are unaffected.

\paragraph{What installments buy (\textsc{judged}).} At $64$ draws a block, $10$-token installments beat
one choice among $64$ whole drafts by $+0.0415$ $[+0.0185, +0.0645]$ (judge~B), at $83.18$ nats for the
whole output and $24.95$ nats a window$^{b}$. Held to one whole-output $\log 64$, or run at $1{,}000$
tokens with $8$ draws a block, installments lose to one choice, and at $1{,}000$ tokens they do not
resolve against the meters at $20.79$ nats (Appendices~\ref{app:blockwise} and~\ref{app:long}).

\section{Leakage}\label{sec:leakage}\label{sec:experiments}

The claims in this section are \textsc{measured}: near-verbatim recall of protected passages, counted on
served text. None is judged.

\paragraph{Models and decoding.}\label{sec:setup} The meter runs at \texttt{TinyComma-1.8B}, the anchor its
authors evaluate \citep{he2026anchored,kandpal2025commonpile}, trained only on openly licensed text. The
risky model is \texttt{Llama-3.1-8B-Instruct} \citep{grattafiori2024llama3} or the authors'
\texttt{Llama-3.1-70B} base. The $8$B is served either as a text continuer (no chat template)
or through its chat template. The authors' byte-level decoder, AnchoredByte
\citep{hayase2026bytesampler}, runs unmodified at Comma-7B with the $70$B base. We decode at temperature
$1.0$ without repetition penalty, and also at the authors' $0.7$ with penalty $1.1$ (\emph{their $0.7$ and
$1.1$}), with $T_{\max}=200$. Intervals are
$95\%$ paired bootstraps. Every band was fixed before its run, except three readings marked post hoc: the
equivalence test, empties counted as losses, and Holm's correction (Appendix~\ref{app:protocol}).
\emph{Recall} is the fraction of a reference's words in in-order near-verbatim spans of at least $20$ words
\citep{ahmed2026extracting}, on $100$ protected passages.

\paragraph{Over a vetted anchor, selection served no near-verbatim span.} In the hardest case we tested,
an attacker scores the $n$ draws by the memorising model's own likelihood. Recall is $0.0000$ at every
anchor, at every $n$ up to $64$ and at $n=256$ (Appendix~\ref{app:sel-premise}, Figure~\ref{fig:safety}a). The memoriser alone reads
$0.3925$. A zero on $100$ passages bounds the per-passage rate below $3.0\%$ at $95\%$, for these
passages, seeds and prompts. A memoriser served through the windowed meter reaches recall $0.01$ at $80$ nats a window, and at
$125$ through the chat template (Appendix~\ref{app:windowed}).

\paragraph{The zero verifies the anchor, not the mechanism.} All six anchors read $0$ on a vetting screen
at every prefix length from $20$ to $200$ tokens, where the open-data \texttt{OLMo-2-7B} and
\texttt{OLMo-2-13B} \citep{olmo2024olmo2} leak on up to $2$ and $6$ of $50$ passages
(Appendix~\ref{app:vetting}). The bound is \emph{relative} to the anchor. With anchors we deliberately
contaminated by fine-tuning on the passages, selection at $n=64$ amplified the anchor's own leakage by
$1.0$ to $7.0\times$ on one draw and $1.0$ to $3.0\times$ on a second (Appendix~\ref{app:sel-premise}, Figure~\ref{fig:safety}b), within
the factor $n$ the certificate permits. Selection multiplies what the anchor leaks; the anchor is the
thing to vet.

\section{Judged utility}\label{sec:utility}

Every claim in this section is \textsc{judged}: a language-model judge compares two texts, and no human
rated anything.

\subsection{Protocol and judge reliability}\label{sec:protocol}

The workload has $500$ ordinary prompts, none from the protected corpus: $200$ questions we wrote,
$150$ WritingPrompts premises \citep{fan2018writingprompts} and $150$ FActScore biography requests
\citep{min2023factscore} (appendix Table~\ref{tab:promptset}). Each arm is judged against one fixed opponent, the
risky model's own draw. Every item is shown in both orders and scores $1$, $\tfrac12$ or $0$. An arm's
\emph{level} is its mean score ($0.5$ is parity). Its \emph{gain} is its level minus that of its own
zero-budget control: the anchor alone for a meter, one draw for selection. A \emph{difference} pairs two
gains over prompts. Judge~B, \texttt{Phi-3.5-mini-instruct} \citep{abdin2024phi3}, was pre-specified;
judge~G, \texttt{gemma-2-27b-it} \citep{gemma2024gemma2}, confirms. Selection's scorer,
\texttt{Qwen2.5-7B-Instruct} \citep{qwen2025qwen25}, never judges an arm it selected.

\paragraph{How far the judges can be trusted.} Judge~B is noisy: it gives the same verdict in both orders
on only $29$ to $37\%$ of items, against $78$ to $85\%$ for judge~G. Position dominates a single order
\citep{zheng2023judging}: shown second, the same two texts win $261$ of $500$ times, and shown first $24$.
Hence every head-to-head averages both orders, a choice that itself moves results (Appendix~\ref{app:judgemethod}).
Judge~B still detects a known difference: a second draw of the risky model gains $+0.065$
$[+0.0395, +0.090]$ over the anchor (Appendix~\ref{app:judgevalidity}). Judged numbers are on the text
recovered after a recording-defect repair, unless marked (Appendix~\ref{app:recording}).

\subsection{The primary comparison}\label{sec:primary}\label{sec:main-results}

The headline configuration serves the $8$B-Instruct as a text continuer at temperature $1.0$, with no
repetition penalty, on the $500$ prompts; selection is best-of-$64$ under the Qwen scorer. The
configuration, judge and $k=10$ difference were registered before the runs. Treating $k=0.5$ as primary is
a choice made in this version, after the results were known, because it is the only published budget whose
certificate bounds a window.

\paragraph{The primary claim.} At $k=0.5$ the meter bounds the window by $0.630$, and selection by about
$10^{-67.6}$ (Table~\ref{tab:certificate}). Selection's judged gain over one anchor draw is reliably
positive. The meter, which blends the two models on about half its steps, has a null gain
(Table~\ref{tab:primary}). The null is informative: it had $80\%$ power against $+0.0305$, and its $90\%$
interval, $[-0.0205, +0.015]$, lies within half of selection's gain, so a two one-sided test (TOST) calls
it equivalent to zero at that margin (post hoc). The comparison is not matched in signal: selection uses
an external reward the meter does not consult. Handed that reward on TriviaQA, the meter gains nothing
(appendix Table~\ref{tab:parity}). Ranking draws by $p_r$'s likelihood instead moves judged utility by
$-0.006$, against a pre-registered band of $+0.077$ (Appendix~\ref{app:currency}).

\begin{table}[t]
\centering\footnotesize
\caption{\textbf{The primary comparison} at the headline configuration ($8$B-Instruct continuing text,
temperature $1.0$, $T_{\max}=200$, $500$ prompts, Qwen scorer, $n=64$). Gain: level minus own zero-budget
control's. Judge~B with $95\%$ interval; judge~G as confirmation. \emph{Window bound}: certified bound on
a $50$-token window. $^\ddagger$A difference of served \emph{levels}, not gains (no paired difference of
gains at $k=0.5$ is reported; levels and gains agree on $19$ of the $21$ rows of Table~\ref{tab:served},
including this one; Appendix~\ref{app:served}). \emph{Never empty}: the rule that never serves an empty draw. \emph{Empties as
losses}: empty answers scored as losses (post hoc); the per-token-meter row is against anchored decoding
at $k=0.5$ and the per-token pathwise meter (appendix Table~\ref{tab:matched}), one figure with no interval. With
empties as losses, the matched differences hold under judge~G at every budget but $W=125$, and under
judge~B fail against the windowed meter at $24.95$ and $40$ nats ($+0.021$ each; Appendix~\ref{app:k10detail}).
$^\S$One pass on one host where judge~B reads the $k=10$ difference at $+0.059$; judge~G intervals and
the remaining detail are in Appendix~\ref{app:k10detail}. $^\star$Two one-sided test, post hoc. Holm: over the $21$ rows of appendix Figure~\ref{fig:h2h}. The
matched-compute meter budget is in Appendix~\ref{app:anchoronly}.}
\label{tab:primary}
\setlength{\tabcolsep}{3.5pt}
\begin{tabular}{@{}lllrl@{}}
\toprule
comparison & window bound & judge~B & judge~G & reading \\
\midrule
\multicolumn{5}{@{}l}{\emph{At the one published meter budget that bounds a window}} \\
best of $64$, gain & $\approx 10^{-67.6}$ & $+0.1015$ $[+0.0775, +0.1260]$ & $+0.176$ & reliably positive \\
anchored decoding $k=0.5$, gain & $0.630$ & $-0.0030$ $[-0.0245, +0.0185]$ & $+0.015$ & equivalent to zero$^\star$ \\
selection $-$ meter, levels$^\ddagger$ & & $+0.1195$ $[+0.0955, +0.1435]$ & --- & reliably positive \\
best of $64$, never empty, gain & $\approx 10^{-67.6}$ & $+0.1010$ $[+0.0775, +0.1245]$ & $+0.155$ & reliably positive \\
best of $64$, gemma scorer, gain & $\approx 10^{-67.6}$ & $+0.066$ $[+0.042, +0.0905]$ & --- & reliably positive \\
vs.\ per-token meters, empties lose & & $+0.094$ & holds & positive, no interval \\
\midrule
\multicolumn{5}{@{}l}{\emph{At $k=10$, where the meter's certificate is vacuous ($K/S_w = 12.5$)}} \\
selection $-$ meter $k=10$ & $1$ (none) & $+0.0505$ $[+0.0155, +0.0860]$ & $+0.0545^\S$ & positive; fails Holm \\
\quad empties as losses & & $+0.0155$ $[-0.0245, +0.0555]$ & --- & inconclusive \\
\quad never-empty rule & & $+0.056$ $[+0.0195, +0.093]$ & $+0.0335$ & B positive, G inconclusive \\
\quad gemma scorer & & $+0.015$ $[-0.0185, +0.0485]$ & $+0.0945^\S$ & inconclusive under B \\
\midrule
\multicolumn{5}{@{}l}{\emph{At matched compute}} \\
$n=2$ $-$ meter ($0.68\times$ cost) & & $-0.0330$ $[-0.0625, -0.0025]$ & --- & favours the meter \\
\bottomrule
\end{tabular}
\end{table}

\paragraph{Which comparisons hold under the same event bound?} Selection against the meter at $k=0.5$
holds decisively in the headline configuration: a far stronger bound, and a reliably positive gain against
a null one. Across serving configurations it holds on levels at every window-informative budget;
on gains it ties at the authors' $0.7$ and $1.1$ on the $8$B (Section~\ref{sec:served}). At a matched
whole-output certificate, best-of-$64$ leads the budget granted up front at $4.16$ nats (the causal
policy Proposition~\ref{prop:sparse} permits, which spends a budget granted at the start), and
installments lead the per-token pathwise meter at $83.18$ nats (anchored decoding's token bucket with
each step's largest log-ratio, rather than its KL, held within the allowance and the sampled token's
realised log-ratio charged, so every output is certified pathwise; Appendix~\ref{app:repairs}). Comparisons
with the windowed meter ($^{c}$ against $^{b}$) and with anchored decoding (KL against max-divergence) are
across certificate types (Appendix~\ref{sec:matched},
\emph{Every mechanism in one judge pass}). Where the meter's certificate is
vacuous there is no shared event bound to match on, and the comparison is with the risky model itself.

\paragraph{The $k=10$ headline is positive but fragile.} It survives fresh draws, and five of six judges
exclude zero. It fails Holm's correction over the $21$ rows of appendix Figure~\ref{fig:h2h} ($p \approx 0.005$), where eight rows survive,
one of them (AlpacaEval) favouring the meter (Appendix~\ref{app:prereg}). It rests partly on the $41$
empty texts selection serves, and is inconclusive with those counted as losses or, under judge~B, with a
gemma scorer (Appendix~\ref{app:k10detail}, \emph{The $k=10$ headline in detail}).

\subsection{Where the meter's certificate is vacuous, the ranking depends on serving}\label{sec:served}

From $k=1$ on, the meter's certificate excludes nothing about a window ($K/S_w \ge 1$). At $k=10$ it serves
the risky model unchanged on $99.99\%$ of steps. The comparison is then with the risky model, and its sign
depends on how that model is served. Table~\ref{tab:reversals} summarises it; appendix Table~\ref{tab:served}
(Appendix~\ref{app:served}) gives every budget.

\begin{table}[t]
\centering\footnotesize
\caption{\textbf{Selection minus meter, by how the risky model is served.} Judge~B, order-averaged served
\emph{levels} paired over prompts, $95\%$ interval, $n=64$; a summary of appendix Table~\ref{tab:served}. Low
budget: $k=0.5$ ($k=0.1$ per byte for AnchoredByte), where $K/S_w$ is $0.56$ to $0.63$ and the meter's
certificate bounds a window. Vacuous budget: the largest $k$ run, where $K/S_w \ge 11.3$. $^\ast$On gains
this row is a tie: selection minus meter $+0.0245$ $[-0.0090, +0.0570]$. $^\dagger$Run on a second host. \emph{favours}: at the low / the vacuous budget.}
\label{tab:reversals}
\setlength{\tabcolsep}{4pt}
\begin{tabular}{@{}lrrrl@{}}
\toprule
risky model, as served & low budget & vac.\ $k$ & vacuous budget & favours \\
\midrule
$8$B, continuing text & $+0.1195$ $[+0.0955, +0.1435]$ & $10$ & $+0.0655$ $[+0.040, +0.092]$ & sel.\ / sel. \\
$70$B base (authors' pair) & $+0.1265$ $[+0.103, +0.1495]$ & $20$ & $+0.1040$ $[+0.080, +0.1285]$ & sel.\ / sel. \\
$8$B, their $0.7$ and $1.1$ & $+0.0325$ $[+0.008, +0.0565]^\ast$ & $10$ & $-0.0730$ $[-0.096, -0.0515]$ & sel.$^\ast$ / \textbf{meter} \\
$70$B, their $0.7$ and $1.1$ & $+0.0970$ $[+0.073, +0.1215]$ & $20$ & $+0.0565$ $[+0.0305, +0.082]$ & sel.\ / sel. \\
AnchoredByte, $70$B / Comma-7B & $+0.1385$ $[+0.1155, +0.1620]$ & $2$ & $+0.1345$ $[+0.1105, +0.1590]$ & sel.\ / sel. \\
$8$B, chat template & $+0.1195$ $[+0.0925, +0.1465]^\dagger$ & $10$ & $-0.1380$ $[-0.167, -0.107]$ & sel.\ / \textbf{meter} \\
\bottomrule
\end{tabular}
\end{table}

\paragraph{Where selection leads.} When the $8$B continues text at temperature $1.0$ (measured before the
other configurations were registered), the risky model loses. Selection leads by more at the authors'
$70$B pair, whose base model at this temperature is fluent but off task, and against AnchoredByte at every
budget.

\paragraph{The reversals.} We found three where the meter's certificate is vacuous, and one more where it
is not (Appendix~\ref{app:servedtext}).
\emph{(i) Chat template.} Served through its template, the $8$B beats its own templateless text by $0.184$.
The meter then first beats selection at $k=3$, where its certificate excludes nothing. A memoriser served
the same way starts leaking at $k=3.9$ $[3.0, 4.4]$. So the meter's win comes before the measured onset of
extraction, but within that onset's interval. The two tie at $k=2$, and since the chat-served model gains
$+0.272$ over the anchor where selection gains $+0.1065$, selection closes less than half the gap
(Appendix~\ref{app:tokenswap}). \emph{(ii) Their $0.7$ and $1.1$, on the $8$B.} Even $k=0.5$ is not the
anchor: the meter gains $+0.0825$ $[+0.0605, +0.1045]$, the gains tie, and at $k=10$ the $8$B wins. At
their $70$B pair selection still leads. \emph{(iii) AlpacaEval.} The lead holds on our prompts,
public-domain books and BookMIA's unseen half, is unresolved on CoTaEval and MT-Bench, and reverses on
AlpacaEval, where the meter wins by $0.0329$ (Appendix~\ref{app:workload}). \emph{(iv) At a certificate
that is not vacuous but is too large to audit,} the windowed meter under the chat template beats selection
at $W=125$ by $0.160$ $[0.122, 0.1985]$, and at $W=40$ under judge~B only. Its window certificate$^{c}$ is
below $S_w=159.8$ but far beyond what a sample can audit.

\paragraph{Against a stronger opponent.} Judged on the recorded text against stronger opponents
than the risky model's own draw, from \texttt{Qwen2.5-1.5B-Instruct} upward, every selection-minus-meter
interval covers zero (appendix Table~\ref{tab:opponentladder}).

\subsection{Selection against its own anchor, and tasks with no judge}\label{sec:breadth}

\paragraph{Across anchors (\textsc{judged}).} Over the openly licensed anchors
\citep{kandpal2025commonpile,langlais2026commoncorpus,bommarito2025kl3m}, selection's gain at $n=8$ excludes
zero at four of six, and is largest at Comma-7B: $+0.126$ $[+0.087, +0.164]$ (Appendix~\ref{app:sel-breadth}, Figure~\ref{fig:breadth}). A
$3.9\times$ larger safe model served alone gains only $+0.015$ $[-0.0065, +0.0365]$
(Appendix~\ref{app:biggeranchor}). A vetted, prompt-blind scorer (Comma-7B) gains $+0.0405$
$[+0.0165, +0.0645]$ under judge~B and $+0.0120$ $[-0.0230, +0.0475]$ under judge~G
(Appendix~\ref{app:promptblind}). On FActScore, selection's precision ($0.042$) is its anchor's and below
every metered arm (Appendix~\ref{app:hemetrics}).

\paragraph{With no judge (\textsc{measured}).} Majority vote over Comma-7B lifts GSM8K
\citep{cobbe2021gsm8k} exact match from $0.320$ to $0.546$ at $n=32$ ($3.47$ nats), $3.4\times$ the
pointwise reward's best lift. It also lifts TriviaQA and MMLU \citep{hendrycks2021mmlu}
(appendix Table~\ref{tab:judgefree}).

\section{Cost}\label{sec:cost}

The claims here are \textsc{measured}, except the \textsc{judged} matched-compute loss. On a server that
runs only the anchor, the $64$ draws cost $21.8\times$ a metered decode (appendix Table~\ref{tab:cost}).
Counted in forward passes, $n=64$ costs $61.3\times$ the meter, mostly for the reward model's prefill. In
wall-clock time, however, the anchor's draws dominate, so reducing $n$ is what reduces cost. Batched into one decode loop, a request takes $0.22\times$ the meter's
time at the authors' $70$B pair and $0.51\times$ at the $8$B pair, because selection never runs the risky
model (Appendix~\ref{app:sel-cost}). Serving the meter by speculative decoding would raise these ratios to
at best $0.37\times$ and $0.88\times$ (Appendix~\ref{app:speculative}). Batched latency is where selection is
cheaper; throughput is where it is dearer.

\paragraph{Held to the meter's own compute, selection loses.} At $n=2$ ($0.68\times$ the cost) the judged
difference is $-0.0330$ $[-0.0625, -0.0025]$, and the loss survives all three prices we measured
(Appendix~\ref{app:anchoronly}). On TriviaQA (exact match, no judge) the meter scores $0.618$ against
selection's $0.190$, even at $11.5\times$ the compute. But the meter needs $k=20$ for this, where it
simply serves the risky model and certifies $480$ nats for a $24$-token answer
(Appendix~\ref{app:judgefree}).

\paragraph{What this means for deployment.} No configuration we ran combines auditable protection, a
practical query horizon, competitive utility and acceptable compute. The unvetted scorer gives a reliable
gain but a $38$-query horizon; a vetted one lifts the cap but gains little. The supported regime is
latency-bound serving of a vetted anchor (Appendix~\ref{app:costtext}).

\section{Related work}\label{sec:related}

\textbf{Certified copyright defences.} Near-access-freeness \citep{vyas2023provable} bounds the served law
against a safe model. Its authors' CP-$k$ already pays once per output: it rejection-samples complete
outputs against models trained on disjoint shards, with a bound that grows as acceptance falls. CP-Fuse
\citep{abad2025cpfuse} fuses such shards per token, and anchored decoding \citep{he2026anchored} charges
per token from a bucket. Selection needs one safe model and no likelihood, and its $\log n$ depends neither
on the rule nor on an acceptance rate. Run with the anchor as its one safe model, CP-$k$ loses to
best-of-$64$ and installments at every matched whole-output certificate (Appendix~\ref{app:cpk}).
\citet{segal2026provably} report that CP-style guarantees can be numerically uninformative and demonstrate
extraction against NAF-protected models \citep[see also][]{cohen2026barriers}. Proposition~\ref{prop:threshold}
makes that exact, and since no R\'enyi order moves the threshold, we change the primitive rather than the
constant.

\textbf{Why not an $n$-gram checker?} Defences without a divergence certificate guarantee something about
the works they are handed: blocklists \citep{ippolito2023preventing}, datastores
\citep{min2024silo,golatkar2024cpr} and filtering \citep{liu2024shield}, compared by CoTaEval
\citep{wei2024cotaeval}. We measured the two most direct (Appendix~\ref{app:blocklist}). An $n$-gram
blocklist (MemFree) costs no judged utility, firing on $0$ of $850$ ordinary prompts, and on the works it
is handed it takes the memoriser's recall from $0.4192$ to $0.0201$. TokenSwap
\citep{prashant2025tokenswap} suppresses as fully as selection and, with our auxiliary model, costs less
utility; both margins depend on the auxiliary (Appendix~\ref{app:tokenswap}). On listed works both are cheaper than selection. But a blocklist is
silent on any work nobody listed, and neither publishes a bound on the served law that composes across
queries. A deployer can run a blocklist and selection together.

\textbf{Training-time mitigations.} Approximate unlearning removes memorised content from the weights
\citep{eldan2023whos}; how completely and robustly it does so is itself under study
\citep{shi2025muse,deeb2024unlearning,lucki2025adversarial}. Such methods are complementary to an
inference-time certificate, which bounds the served law and changes no weights. We measure none of them.

\textbf{Best-of-$n$ and private selection.} We claim no novelty for best-of-$n$ sampling or its divergence
bound. Selecting among draws is standard inference-time alignment
\citep{gao2023scaling,mudgal2024controlled,wang2023selfconsistency}. The KL bound is
\citet{beirami2025bestofn}'s, characterised further by \citet{huang2025bestofn,gui2024bonbon,yang2024asymptotics}. Blockwise
best-of-$K$ \citep{mudgal2024controlled} is selection in installments, with the boundary-conditioned window
certificate of Proposition~\ref{prop:installments}. Charging the choice among repeated runs is private
selection's accounting \citep{liu2019private,papernot2022hyperparameter}; private next-token prediction
mixes a public model in per token \citep{ginart2022submix,flemings2024pmixed,majmudar2022dpdecoding} or
pays only where the two disagree \citep{amin2024private}. That literature reads the divergence as a
\emph{cost} \citep{hughes2024bestofn}; we read it as a copyright certificate. Apart from CP-$k$, we know of no prior
statement that any rule choosing among a safe model's draws is near-access-free. The closest neighbour,
\citet{kalai2026consensus}, abstains unless several models agree (Appendix~\ref{app:kalai}).

\section{Limitations and conclusion}\label{sec:limitations}

Each limitation names the kind of claim it qualifies. Appendix~\ref{app:limitations} gives them in full.

\paragraph{The anchor binds first (\textsc{measured}, \textsc{judged}).} A served string is one the anchor
would have drawn, so an anchor that cannot do the task gains nothing at any $n$. On TriviaQA
\citep{joshi2017triviaqa} the meter wins only by serving the risky model, and selection does not improve
facts: its FActScore precision is its anchor's. A capability gap is the regime a deployer most wants
covered, and \emph{neither} mechanism covers it with an auditable certificate. Comma-7B, our largest
vetted anchor, is far below frontier scale. On web-scale data it is hard to show that no protected
work leaked in, and selection multiplies whatever a contaminated anchor leaks (appendix Table~\ref{tab:contam}). We
did not test long agentic rollouts or multi-step tasks.

\paragraph{The scorer binds next (\textsc{measured}, \textsc{judged}).} Below the anchor's ceiling the
gain is the scorer's capability. On GSM8K the pointwise reward gains $3.4\times$ less than majority vote;
on TriviaQA it falls with $n$; on CoTaEval it \emph{loses} F1; a $0.5$B scorer never reaches the meter
(Appendix~\ref{app:selection}). A vetted scorer is not yet a substitute for our
unvetted one.

\paragraph{The judged ranking measures the serving configuration as much as the mechanism
(\textsc{judged}).} Against the risky model it reverses under a chat template, on AlpacaEval
\citep{li2023alpacaeval}, and on the $8$B at the authors' sampling settings; a windowed meter whose
certificate is too large to audit also wins under the chat template; and the ranking vanishes against a
stronger opponent. The $k=10$ headline fails Holm's correction, and becomes inconclusive with empties
counted as losses or, under judge~B, with a scorer from another family. Results also depend on
order-averaging. Every judge is a language model, the pre-specified one is order-consistent on about a
third of items, and no human rated anything. Tempering the anchor buys part of selection's gain
(Appendix~\ref{app:tempering}). Selection is also paid for in compute (\textsc{measured}): cheaper only in
batched latency, far dearer in throughput, and at matched compute it \emph{loses} (\textsc{judged}).

\paragraph{What the certificates cover (\textsc{proved}; the audit is \textsc{measured}).} The certificate is relative to the anchor as
sampled. Its audit covers one fixed prompt distribution per work, the query horizon is finite unless the
scorer is vetted (which the certificate cannot check; Section~\ref{sec:compose}), and an installment
window is certified only given the text up to a block boundary (Section~\ref{sec:frontier},
Proposition~\ref{prop:installments}). A
certificate says nothing about works shorter than it: on
TriviaQA answers, $\log 64$ reaches $S(x)$ on $35.4\%$ of questions (Appendix~\ref{app:gainwhere}). On protected
quotations, selection's worst case leaks $0.38\%$, against $69.8\%$ for the meter at $k=10$
(Appendix~\ref{app:shortworks}). We
tested no lyrics, whole poems or code, and extraction covers sixteen English novels plus four in French and
German.

\paragraph{Why certify at all.} Through the chat template the meter wins before measured leakage, and a
deployer may rationally prefer that. Only a certificate covers works nobody tested and paraphrases nobody
scored. How close a policy of the shape
Proposition~\ref{prop:sparse} permits can come to Lemma~\ref{thm:nfl}'s frontier is open.

\paragraph{Conclusion.}\label{sec:conclusion}
A per-token budget grows with the output and, at deployed settings, certifies nothing about the works it
protects. Spent choosing among a safe model's draws, a budget is $\log n$ nats at any output length
($B\log n$ in installments, with boundary-conditioned windows). At a matched whole-output certificate, best-of-$64$ leads a meter given its
whole budget up front ($4.16$ nats), and installments lead a per-token pathwise meter ($83.18$ nats). At the one published meter budget that bounds a window, selection has the far stronger event
bound and a reliably positive judged gain against the meter's null; at the authors' sampling settings on
the $8$B the two tie. Where the meter's certificate is vacuous, the ranking depends on serving. Selection
costs more compute, loses at matched compute, and can be no better than its vetted anchor.

\section*{Ethics statement}
Auditing a copyright defence means showing where it fails. This paper therefore fine-tunes memorising
models and attacks a protected decoder. We report aggregate recall only, never reconstructed text. The
passages are already published as extraction benchmarks (CopyBench, \citealp{chen2024copybench}; BookMIA,
\citealp{shi2024detecting}). The repository (available at \codeurl{}) ships the fine-tuning recipes and every
analysis script. It does not ship the merged memorising weights, which reproduce copyrighted text on demand.

The constructive result carries its own risk: misplaced confidence. The $\log n$ certificate bounds the
probability that the served text is a protected work \emph{relative to the anchor's own probability}. It is
only as good as the anchor. Over twelve deliberately contaminated anchors, an adversarial scorer raised the
anchor's own leakage at $n=64$ by $1.0$ to $7.0\times$ on one draw and $1.0$ to $3.0\times$ on a second
(up to $10.0\times$ at $n=256$), against the factor of $n$ the certificate permits (Appendix~\ref{app:contaminated}).

The anchor's leakage rate is the one quantity a deployer can measure with no attack and no access to the
mechanism: draw from the anchor on the protected passages' own prefixes and score recall.
\textbf{The prefix length is part of the test.} \texttt{Llama-3.1-70B}, which memorised the work in
pre-training, reads $0.0000$ at fourteen genuine prefix tokens and leaks on $31$ of $50$ passages at
$150$. The check is not monotone in the prefix. It must therefore run at every prefix length the deployment
accepts, and at the temperature the deployment samples at (Appendix~\ref{app:vetting}). Passing is
evidence, not proof. Like a blocklist, the check speaks only about the works it is handed.

Our statements concern mechanisms. Whether near-access-freeness tracks infringement is contested
\citep{elkinkoren2024copyright}, and we make no claim about legal thresholds for infringement. The
recommendations we would rather see quoted than the theorems are these. Publish $k/s(x)$ rather than $k$.
Prefer a budget whose size does not grow with what it protects. Vet the anchor before quoting $\log n$,
because $n$ multiplies whatever the anchor already leaks. Unless the scorer is vetted too, cap the queries
of every set of users who could pool transcripts (in practice, the whole deployment) below $S/\log n$ for
the shortest protected event. A scorer that colludes with the user can spend the transcript's $m \log n$
nats one query at a time (Section~\ref{sec:compose}). A vetted scorer is an operational condition; the
certificate cannot verify it.

\section*{Reproducibility statement}
Every number in this paper is written by an analysis script to a CSV before it is quoted. The code, the
CSVs and the command that produced each are available at \codeurl{}. Proofs are in
Appendix~\ref{app:proofs}. Every interval we score against was fixed before the run that produced it,
together with its grid, admission criterion and excluded outcomes, except the readings marked post hoc
(the equivalence test, empties counted as losses, and Holm's correction). Budget accounting is checked per
trajectory. Every run reports the $k=-1$ (risky only) and $k=0$ (safe only) baselines on the same prompts
and seeds.

A reproducer should expect two properties. First, selection serves the argmax of a reward computed in
bfloat16, and that argmax is not bitwise reproducible. Re-scoring the same $32{,}000$ candidates on
another GPU architecture, at another batch size or in float32 serves a different completion on $5.6$ to
$8.6\%$ of prompts at $n=64$. What is served is worth the same: the alternative gives up $0.011$ to $0.03$
nats against a gap of $5.1$, and Proposition~\ref{prop:selection} holds for any tie rule. This was measured
rather than assumed, and it carried no pre-registered interval. Second, the harness recorded each
generation with the tail of its prompt, an artefact of slicing left-padded batches. Nothing served was
affected, and every recorded generation is recovered exactly. Table~\ref{tab:served} and
Figures~\ref{fig:selection}, \ref{fig:h2h} and~\ref{fig:breadth} are judged on the recovered text
(Appendix~\ref{app:judgemethod}). A few appendix readings are on the recorded text, and are marked where
quoted.

\section*{LLM usage}
LLMs were used extensively in preparing this version: to restructure and rewrite the text, to check
every reported number against the analysis outputs, and to check the proofs on small instances. The author
takes full responsibility for all content. Instruction-tuned models also appear in two other roles. They are \emph{measurement
instruments}, judging output quality. They are also a \emph{component of the mechanism}, supplying the
score that selection ranks by. For that reason no arm is scored by the model that selected it, and
Proposition~\ref{prop:selection} is stated for an arbitrary score. Total compute for this work is at most
$366$ GPU-hours on A100 hardware, most of it the budget sweeps and the judged arms. The memorisation
fine-tunes account for at most $73$ of those hours.

\bibliographystyle{iclr2027_conference}
\bibliography{references}

\appendix
\small
\clearpage
\section*{Guide to the appendices}\label{app:guide}
\noindent Table~\ref{tab:guide} maps each theme of the paper to the appendix that holds it, and
Table~\ref{tab:notation} collects the symbols and terms used throughout.

\begin{table}[!ht]
\centering\footnotesize
\caption{Where each part of the evidence lives.}\label{tab:guide}
\begin{tabular}{@{}p{0.2\textwidth}p{0.08\textwidth}p{0.64\textwidth}@{}}
\toprule
theme & where & what it holds \\
\midrule
Main-text detail & \ref{app:detail} & the audit's cost and the colluding scorer; every mechanism in one judge pass; the $k=10$ headline's robustness; the scorer-free instance; the full text behind each condensed main-text section (events the certificate prices, audit scope, the meter's repairs, the served comparison, cost, limitations) \\
Theory & \ref{app:proofs} & every proof; the threshold for a window event; tightness and composition of $\log n$ \\
Installments & \ref{app:proofs}, \ref{app:blockwise} & Proposition~\ref{prop:installments}'s proof and the counterexample to conditioning inside a block; the blockwise runs, with each window certificate's conditioning marked \\
Protocol & \ref{app:protocol} & workload, models, reward, judge, the recording defect and its repair \\
Head-to-head & \ref{app:sel-h2h} & robustness of the served comparison (Table~\ref{tab:served}), the authors' metrics and byte-level decoder, judges, opponents, workloads \\
Selection alone & \ref{app:sel-breadth} & $n$ past $64$, other anchors, scorer size, majority vote, tasks with no judge, CoTaEval \\
Leakage & \ref{app:sel-premise} & extraction, a natural memoriser, other languages, short works, vetting, contaminated anchors, audit \\
Cost & \ref{app:sel-cost} & forward passes, the clock, a deployable server, batching, matched compute \\
The meter, audited & \ref{app:onset} & imitation cost at four pairs, onset over nine pairs, the opening, vacuity at scale \\
Repairs & \ref{app:repairs} & sharper charge, lower cap, schedule, budget up front, frontier distance, closer anchors \\
Limits, related & \ref{app:limitations}, \ref{app:related} & limitations in full; positioning and incumbents measured \\
\bottomrule
\end{tabular}
\end{table}

\begin{table}[!ht]
\centering\footnotesize
\caption{Notation and terms.}\label{tab:notation}
\begin{tabular}{@{}p{0.2\textwidth}p{0.76\textwidth}@{}}
\toprule
symbol or term & meaning \\
\midrule
$p_s$, the anchor & \raggedright the safe model, trained without the protected work; every certificate is relative to it as sampled \tabularnewline
$p_r$ & \raggedright the risky model, which may hold the work; the meter tilts toward it, selection never runs it \tabularnewline
$q$ & \raggedright the law a mechanism actually serves \tabularnewline
$x$, $S(x)$, $s(x)$ & \raggedright a protected work; $S(x) = -\log p_s(x)$, its surprisal under the anchor in nats; $s(x)$, that surprisal per token \tabularnewline
$S_w$ & \raggedright the anchor's surprisal of a $50$-token window of a protected passage given its true prefix in the prompt, i.e.\ a window at the start of the output, median $159.8$ nats; a window preceded by \emph{served} text is priced, under a joint bound, by the anchor's probability of that served prefix together with the window \tabularnewline
$k$, $T_{\max}$, $K$ & \raggedright the meter's per-token budget, the output cap ($200$ tokens), and its certificate $K = kT_{\max}$ \tabularnewline
$n$, $\log n$, $m$ & \raggedright the draws selection chooses among; its pathwise certificate (Proposition~\ref{prop:selection}); $m\log n$ over $m$ queries \tabularnewline
$L$, $B$ & \raggedright tokens per installment when selection is spent in installments, and the number of installments (blocks). The whole output is certified at $B\log n$. A $w$-token window is certified at $\min(\lceil w/L\rceil+1, B)\log n$ only \emph{conditional on the served text up to the last block boundary at or before the window's start}; conditional on the served tokens inside that block no bound of this form holds (Proposition~\ref{prop:installments}) \tabularnewline
span$^{c}$, span$^{b}$ & \raggedright the two window-certificate types in the tables: $^{c}$, conditional on \emph{all} served text before the span (the windowed meter); $^{b}$, conditional on the served text up to the last block boundary (installments). They are not directly comparable \tabularnewline
$D_\alpha$, $D_\infty$ & \raggedright R\'enyi divergence of order $\alpha$; $D_\infty$, the max-divergence, is the pathwise order \tabularnewline
$b_\alpha(K)$ & \raggedright the most a budget of $K$ nats at order $\alpha$ allows an event of anchor probability $e^{-S}$ (Proposition~\ref{prop:threshold}); it is $1$, and the certificate vacuous, iff $K \ge S$ \tabularnewline
$N_\varepsilon$ & \raggedright the steps at which a causal policy spends more than $\varepsilon$ (Proposition~\ref{prop:sparse}) \tabularnewline
$I_T$, $\gamma$ & \raggedright the imitation cost, the KL of $p_r$ from $p_s$ summed over a meter's slack steps (Proposition~\ref{prop:imitation}); $\gamma = 0.857$ nats per token on ordinary text \tabularnewline
$\Lambda^*_s(u)$ & \raggedright Lemma~\ref{thm:nfl}'s frontier, the least divergence from $p_s$ at which expected utility $u$ is attainable \tabularnewline
\midrule
meter & \raggedright anchored decoding \citep{he2026anchored}: a KL budget charged per token, tilting toward $p_r$ \tabularnewline
selection & \raggedright serving one of $n$ anchor draws by any rule, a reward, a vote or an adversary \tabularnewline
onset & \raggedright the smallest budget at which mean near-verbatim recall reaches $0.01$ \tabularnewline
near-verbatim recall & \raggedright the fraction of a reference's words inside in-order near-verbatim spans of at least $20$ words \tabularnewline
level, gain & \raggedright an arm's order-averaged score against one fixed opponent ($0.5$ is parity); its level minus its own control's \tabularnewline
$D_1$--$D_5$ & \raggedright selection's gain; the meter's; their paired difference; an extra arm's gain; the extra arm minus selection \tabularnewline
judges A--G & \raggedright A, \texttt{Qwen2.5-7B} (selection's reward, never judging its own picks); B, \texttt{Phi-3.5-mini}, pre-specified; C--G, \texttt{Llama-3.1-8B}, \texttt{Qwen2.5-72B}, \texttt{Mixtral-8x7B}, \texttt{Qwen2.5-14B}, \texttt{gemma-2-27b} \tabularnewline
\bottomrule
\end{tabular}
\end{table}

\section{Main-text detail}\label{app:detail}

This appendix holds detail moved out of Sections~\ref{sec:method}--\ref{sec:limitations} to keep the
main text short. Each main-text summary points here; nothing here changes a main-text claim, and every
negative result summarised there is stated here in full.

\paragraph{What the audit costs (\textsc{measured}).}\label{app:auditcost}
The audit is affordable for one work but not free. At $n=64$, certifying one work at $1\%$ takes
$19{,}200$ draws, about $0.12$ A100-hours. A catalogue of $10^5$ works takes $11{,}901$ GPU-hours. The
draws are per work and reusable across queries, so this is paid once per catalogue. Our own $25{,}600$
clean draws certify only the \emph{average} passage, at $0.75\%$ for $n=64$. That is a statement about a
passage drawn at random, not about any one work (Appendix~\ref{app:perwork}).

\paragraph{A colluding scorer (\textsc{proved} and \textsc{measured}).}\label{app:colluding}
A keyed hash \citep[steganography, as in][]{schroederdewitt2023perfectly} realises $3.97$ of the $6$ bits
that one response at $n=64$ can carry (Appendix~\ref{app:channel}). A cruder collusion extends a
memorised passage one token per query, serving the draw whose first token the memoriser ranks highest. It
reconstructs none of $100$ windows after $50$ queries at $n=64$, although the certificate ($207.9$ nats)
is vacuous for $88$ of them. The anchor's draws hold the true next token on only $12.3\%$ of queries
(Appendix~\ref{app:prefixext}). That is looseness in the bound, not safety.

\paragraph{Every mechanism in one judge pass (\textsc{judged}).}\label{sec:matched}
Table~\ref{tab:matched} puts every mechanism we built into one judge pass at the headline configuration.
Its rows match in nats, not always in type. Two kinds of comparison are therefore reported separately.
\begin{itemize}[leftmargin=*,itemsep=2pt]
\item \emph{At a matched whole-output certificate.} Best-of-$64$ leads the budget granted up front at
$4.16$ nats by $+0.0785$ $[+0.0445, +0.1115]$. Installments ($L=10$, Proposition~\ref{prop:installments})
lead the per-token pathwise meter at their common whole-output $83.18$ nats by $+0.142$
$[+0.1085, +0.1755]$.
\item \emph{Across certificate types.} The windowed meter certifies each $50$-token span conditional on
\emph{all} served text before it. Installments certify a window only conditional on the served text up to
the last block boundary at or before its start (Proposition~\ref{prop:installments}). Selection's
$\log 64$ is a joint bound on the whole output. These are different guarantees, so equal nats do not make
them equal. With that caveat, best-of-$64$ leads the windowed meter at $W = 4.16$ nats a span by $+0.090$
$[+0.058, +0.122]$. Installments, at $24.95$ nats a window conditional on the block boundary, lead the
windowed meter at $24.95$ nats a span by $+0.1205$ $[+0.0855, +0.155]$. They also lead anchored decoding at
$k=0.5$, whose $100$-nat KL certificate is of a third type, by $+0.161$.
\end{itemize}
Judge~G gives all nine registered differences the same sign and resolves all but one. Empty answers
qualify this. Counted as losses (post hoc), the differences hold under judge~G at every budget but
$W=125$. Under judge~B they hold against the per-token meters ($+0.094$) and the windowed meter at
$\log 64$ ($+0.048$), but not at $24.95$ or $40$ nats ($+0.021$ each; Appendix~\ref{app:judgevalidity}).
The windowed meter improves with its budget and draws level only at $W = 125$: $+0.022$
$[-0.0145, +0.0585]$. We had predicted that selection would still lead there. At that budget a memoriser
served through it already reproduces near-verbatim text: $0.106$, against $0.495$ alone and $0.000$ at
every $W \le 40$.

\begin{table}[ht]
\centering\footnotesize
\caption{\textbf{Every mechanism in one judge pass}, $500$ prompts, headline configuration. Certified
nats for a $50$-token span and for the $200$-token output; max-divergence, except KL for anchored
decoding at $k=0.5$. ---: no span certificate beyond the output's. The span column holds two different
types. $^c$: conditional on \emph{all} served text before the span (windowed meter). $^b$: conditional on
the served text up to the last block boundary at or before the span's start (installments,
Proposition~\ref{prop:installments}); no bound holds conditional on the served tokens between that
boundary and the span. The output column is a joint bound for every row, so rows that share an output
value are matched; rows that share only a span value are compared across certificate types. Gain over the
arm's own zero-budget control: judge~B with $95\%$ interval, and judge~G. \emph{Leak}: near-verbatim
recall of $100$ memorised passages with the memoriser as risky model ($0.495$ alone); for selection, the
best of its $64$ draws. All arms: Table~\ref{tab:matchedfull}.}
\label{tab:matched}
\setlength{\tabcolsep}{4pt}
\begin{tabular}{@{}lrrlrr@{}}
\toprule
mechanism & span & output & gain, judge~B & judge~G & leak \\
\midrule
anchored decoding, $k=0.5$ & --- & $100$ & $-0.0030$ $[-0.0245, +0.0185]$ & $+0.015$ & --- \\
budget up front & --- & $4.16$ & $+0.0230$ $[+0.0010, +0.0445]$ & $-0.0265$ & --- \\
per-token pathwise meter & --- & $83.18$ & $+0.0160$ $[-0.0050, +0.0370]$ & $+0.0125$ & --- \\
windowed meter, $W$ per span & $4.16^c$ & $16.64$ & $+0.0115$ $[-0.0110, +0.0340]$ & $-0.047$ & $0.000$ \\
 & $24.95^c$ & $99.81$ & $+0.0375$ $[+0.0150, +0.0600]$ & $-0.0045$ & $0.000$ \\
 & $40^c$ & $160$ & $+0.0375$ $[+0.0140, +0.0610]$ & $+0.0475$ & $0.000$ \\
 & $125^c$ & $500$ & $+0.0795$ $[+0.0530, +0.1055]$ & $+0.130$ & $0.106$ \\
\midrule
best of $64$ & --- & $4.16$ & $+0.1015$ $[+0.0775, +0.1260]$ & $+0.176$ & $0.000$ \\
installments, $L=10$ & $24.95^b$ & $83.18$ & $+0.1580$ $[+0.1330, +0.1835]$ & $+0.305$ & --- \\
\bottomrule
\end{tabular}
\end{table}

\paragraph{The $k=10$ headline in detail (\textsc{judged}).}\label{app:k10detail}
\emph{Fresh draws and other judges.} The difference against the $k=10$ meter, $+0.0505$
$[+0.0155, +0.0860]$, survives fresh draws: re-drawing selection, and then every sampled arm, moves it by
at most $0.0055$. Five of six judges exclude zero. These include the two most self-consistent judges,
order-consistent on more than three quarters of items: \texttt{Qwen2.5-72B} ($+0.055$, the scorer's
family) and \texttt{gemma-2-27b-it} ($+0.0545$). The sixth, \texttt{Mixtral-8x7B}, agrees with itself on
only $35$ to $42\%$ and straddles zero at $-0.0015$. On the recorded (pre-repair) text, over $850$
prompts, it bounded the difference below $0.03$: $+0.0226$ $[-0.0018, +0.0476]$
(Appendix~\ref{app:h2hrepeat}). The largest reading, $+0.1210$, comes from \texttt{Qwen2.5-14B}, also the
scorer's family. Without the two Qwen judges, three of four still exclude zero (Figure~\ref{fig:h2h}).

\emph{Empty answers.} The reward prefers empty drafts. Selection serves $41$ empty texts and the meter
none. Judge~B scores those empty picks near parity ($0.494$) and judge~G above it ($0.634$). Counting
empty answers as losses leaves the $k=10$ difference at $+0.0155$ $[-0.0245, +0.0555]$. A rule that never
serves an empty draw is still certified at $\log n$, because it is a selection rule over the same draws.
With empties as losses it keeps $+0.056$ $[+0.0195, +0.093]$, and it keeps the reading at all eight
configurations of Table~\ref{tab:served} that were re-run (Table~\ref{tab:nonempty}). Judge~G reads it
at $+0.0335$ $[-0.020, +0.0875]$. Under the most reliable judge, the headline therefore rests partly on
the empties (Appendix~\ref{app:empties}).

\emph{The scorer.} Scored by \texttt{gemma-2-27b-it} instead, selection serves shorter drafts (median $34$
words against $75$) and more empty ones ($71$ against $41$). It gains $+0.066$ $[+0.042, +0.0905]$, and
the $k=10$ difference falls to $+0.015$ $[-0.0185, +0.0485]$ (pre-registered). Under judge~B the headline
thus depends on the scorer. Four of six judges still resolve it with either scorer, and only judge~B
prefers the Qwen reward's drafts (Appendix~\ref{app:factorial}).

\paragraph{The scorer-free instance (\textsc{measured}).}\label{app:scorerfreemain}
Majority vote has no scorer. It costs $5.75\times$ the meter's forward passes at $n=32$, and it beats the
pointwise reward up to a $7.6$B scorer but not at $14$B (Appendix~\ref{app:scorerfree}).

\subsection{What the certificate prices, in detail}\label{app:certdetail}

\paragraph{The three protected objects.} \begin{itemize}[itemsep=1pt,topsep=2pt]
\item A \emph{passage} is the continuation a CopyBench item protects \citep{chen2024copybench}. It is
  $64$ tokens long, and at the median of the $758$ passages we test it costs $204.8$ nats under the anchor.
\item A \emph{window} is $50$ tokens of protected text, given the true text before it. Its median
  surprisal is $S_w = 159.8$ nats.
\item A \emph{target} is what an extraction sweep asks for: all of a passage's text after a $20$-token
  seed. At the median of the $100$ passages attacked it is $259.5$ tokens and $849$ nats.
\end{itemize}

\paragraph{Reference law, rules with no score, tightness, and the chain rule.} The proof uses only one fact: each draw's marginal is $p_s$. The bound therefore survives deterministic
post-processing and early stopping (Appendix~\ref{app:proofs}). Our draws are untruncated samples at the
stated temperature, so the reference is the anchor exactly as sampled. Any fixed sampling law of the
licensed model may serve as the reference, tempered or not, provided it is vetted at its own temperature
(Appendix~\ref{app:vetting}).

The bound also covers rules with no score in them. \textbf{Self-consistency
\citep{wang2023selfconsistency} is an instance}: it serves one of the draws, one with the modal answer.
The bound is tight. If $f$ has a unique maximiser $y^\star$ with $p_s(y^\star) = p$, then
$q(y^\star) = 1-(1-p)^n \to np$ as $p \to 0$.

The guarantee is \emph{relative} to $p_s$, and it is only as good as the anchor. Self-consistency over a
web-trained model, as it is usually run, satisfies the same bound and certifies nothing about copyright
(Section~\ref{sec:leakage}). Nor does selection evade the chain rule: its served law factorises
token by token like any other. What differs is that the charge is paid once, on the whole output, by a
rule whose conditionals depend on complete candidates (Appendix~\ref{app:noseparation}).

Proposition~\ref{prop:selection} is a statement about the \emph{whole output}. Applied per block, it
gives the whole-output certificate for selection in installments. A window that starts inside a block is
a different matter: the per-block bound holds only given the text up to the block boundary, not given the
served tokens inside the block (Proposition~\ref{prop:installments}).

\paragraph{Events other than the window at the start.} The closed form $n e^{-S_w}$ prices one specific event: the exact window, served as the start of the
response, with the work's true preceding text as the prompt. Two other events need more care.
\begin{itemize}[itemsep=1pt,topsep=2pt]
\item \emph{The window at any position.} A $200$-token response has at most $151$ positions where a
  $50$-token window can start. Propositions~\ref{prop:threshold} and~\ref{prop:selection} bound the union
  over positions directly, so no position factor enters either certificate. But the anchor's probability
  of that union has no closed form, and must be sampled (Section~\ref{sec:compose};
  Appendix~\ref{app:windowvacuity}).
\item \emph{A window preceded by served text.} The certificate is joint over the whole output. It bounds
  the probability of ``served prefix, then window'' by $n$ times the anchor's probability of that joint
  event. It does not bound the window's probability \emph{given} a served prefix that was itself
  unlikely under the anchor.
\end{itemize}

\paragraph{The audit's scope in full.}\label{app:auditscope} The statement has a narrow scope, and we state it in full.
\begin{itemize}[itemsep=1pt,topsep=2pt]
\item \emph{One fixed prompt distribution.} The draws are taken at the work's own prefix, at a stated
  prefix length and sampling temperature. The bound is about that prompt only. The certificate holds
  prompt by prompt, $\Pr_q[E \mid x] \le n\Pr_{p_s}[E \mid x]$, and the user chooses $x$
  (Appendix~\ref{app:vetting}).
\item \emph{$95\%$ confidence.} The bound fails with probability up to $5\%$ over the audit sample.
\item \emph{Zero observed hits.} The $3/N$ form is the zero-hit case. An audit that observes hits needs
  an exact binomial upper limit instead, and we report none.
\item \emph{One event.} The event is near-verbatim reproduction as the audit's matcher defines it.
\end{itemize}

Extending the bound to other prompts needs more audit, not more argument. Each additional prompt must be
audited in its own right. For a finite prompt family fixed in advance, a union bound over its $P$
members gives a simultaneous statement. With zero hits in $N$ draws, the $1-\delta$ upper bound is
$\ln(1/\delta)/N$; the $3/N$ above is $\ln 20/N \approx 3.0/N$, at $\delta = 0.05$. Auditing each member at
$\delta = 0.05/P$ therefore gives $\ln(20P)/N$ per member, at $95\%$ joint confidence. The prefix length is part of the family, because the anchor screen is not
monotone in it (Appendix~\ref{app:vetting}). A prompt family chosen adversarially \emph{after} the audit,
or one too large to enumerate, is not covered.

Pooled adaptive queries are covered by composition, not by the audit. If every query uses a prompt from
a family of $P$ prompts audited simultaneously, the union bound gives at most $m\,n\ln(20P)/N$ for the
event that \emph{some} response contains $E$, at $95\%$ joint confidence. For an event spread across
responses, the transcript is certified only at $m\log n$ against $P_s$, and the per-prompt audit does not
estimate the anchor's probability of such an event.

\paragraph{Checking that a scorer is vetted.} A \emph{vetted} scorer is one that neither holds the work nor takes the user's instructions. Such a
scorer has no channel through which to steer the transcript toward the work, so in practice it removes
the need for the cap. This is an operational condition, and \emph{the certificate cannot verify it}.
Formally, the transcript bound remains $m\log n$ whatever the scorer is. Whether a scorer is vetted must
be checked outside the certificate: by the provenance of its training data, and by confirming that it
scores without the user's prompt.

\subsection{The meter in detail}\label{app:metertail}

\paragraph{Sparsity, measured on selection.} The proof is the exact chain rule and Markov's inequality at each step (Appendix~\ref{app:proofs}). A
per-token allowance on the whole sequence therefore has two options. If it grows with the work, $K = kT$,
it is vacuous from $k = s(x)$ on, at every order (Proposition~\ref{prop:threshold}). If it stays bounded,
the policy departs from the anchor by more than $\varepsilon$ at $O(K/\varepsilon)$ steps and is the anchor
everywhere else. The bound constrains the \emph{shape} of a spend, not the class of mechanism. Selection's
served law satisfies it too: at its $3.175$-nat KL bound, fewer than four of its $204$ steps carry a whole
nat. Its budget sits on a choice among complete drafts, which the decode loop does not contain.

\paragraph{What reproduction and utility cost.} This is the Donsker--Varadhan (Gibbs) variational principle \citep{donsker1975asymptotic} applied to the
whole sequence. We claim no novelty for it (Appendix~\ref{app:proofs}), and nothing about the decoder
enters. Reproducing a work costs $S(x)$, which grows with its length: $849$ nats for the median protected
target and $159.8$ for a $50$-token window. Buying mean utility $u$ costs $\Lambda^*_s(u)$. That cost is
$O(1)$ at any length for a utility the safe model itself sometimes attains. So there is room for a
mechanism that is both safe and useful. A meter that tilts toward the risky model does not occupy that
room, because where its bucket is slack it pays the full cost of imitation, whatever utility that buys.

\paragraph{The deployed meter in full (\textsc{measured}; Figure~\ref{fig:horns}).} By the time
extraction begins, the certificate is already void. On nine pairs the onset lies between $0.88$ and $1.17$
of $s(x)$, near the whole-target threshold. The onset level is robust; the order of the pairs is not.
Changing only one memoriser's fine-tuning length moves its onset from $0.88$ to $1.35$. Two mechanistic predictions
of the onset failed (Appendix~\ref{app:onset}). Where the budget does certify something, the meter binds
and is judged indistinguishable from its own anchor (Table~\ref{tab:matched}). The share of steps its
bucket constrains falls from $0.954$ at $k=0.1$ to $0.033$ at $k=3$ and $0.0000$ at $k=20$. Past the
imitation rate it serves the risky model unchanged and spends $171.3$ nats, $4.3\%$ of the $4000$ it
certifies at $k=20$. By Lemma~\ref{thm:nfl}, the utility it reaches costs only $0.023$ nats
(Appendix~\ref{app:frontierratio}). No per-trajectory violation occurs in over $100{,}000$ budgeted
trajectories. The gap is between what the budget bounds and how $k$ is read, not in the accounting.

\paragraph{The repairs a meter has left (\textsc{measured} and \textsc{judged}).} Each repair we tried
lands on one of the two options or falls short of selection (Table~\ref{tab:repairs}). A sharper charge
($\alpha=8$) suppresses extraction only by binding on $99.6\%$ of steps. A lower cap ($k=0.5$) is its
anchor. A better schedule buys $3$ to $13\%$ more fidelity at $k \le 1$. A budget granted up front gains
$+0.0545$ $[+0.0315, +0.0775]$ at $64$ nats, against selection's $+0.1045$ at $\log 64$ in the same
pre-repair pass (judge~B; Appendix~\ref{app:repairs}, Table~\ref{tab:h2hrepeat}). The repairs all rank tokens by $p_r$'s likelihood,
and used to choose among draws that likelihood predicts judged quality no better than answer length
(Appendix~\ref{app:currency}).

\paragraph{The windowed meter in full.} The first is a \emph{windowed meter} (\textsc{proved} by construction, then \textsc{measured}). At each step it caps the
max-divergence by $W$ minus the largest realised log-ratio of a recent span. Every span of at most $w$
tokens is then certified at $W$ conditional on \emph{all} served text before it; we mark this type
$^{c}$. Ordinary text's realised log-ratio per $50$-token window has median $40.3$ nats
(Appendix~\ref{app:windowlr}). We built the meter with $w=50$ (Appendix~\ref{sec:matched},
Appendix~\ref{app:windowed}). Its certificate covers every event in a span up to a factor $e^{W}$,
near-verbatim variants included. It prices the median exact window at $e^{W-159.8}$. For a near-verbatim
variant it says something only after an audit that samples past $3e^{W}$ draws. A memoriser served through
it reaches the $0.01$ recall onset at $W=80$ (Appendix~\ref{app:windowed}).

\subsection{Leakage, protocol and the headline in detail}\label{app:expdetail}

\paragraph{Leakage over a vetted anchor, in full (\textsc{measured}).} A scorer can only amplify
leakage that the anchor already has (Proposition~\ref{prop:selection}). In the hardest case we tested, an
attacker scores the $n$ draws by the memorising model's own likelihood. Near-verbatim recall is $0.0000$
at every anchor, at every $n$ up to $64$ and at $n=256$ (appendix Figure~\ref{fig:safety}a). On the same passages the memoriser alone reads $0.3925$ mean and $0.8154$
maximum. The $70$B, unaided, reads $0.2475$ and reproduces two passages in full. A zero on $100$ passages
bounds the per-passage rate below $3.0\%$ at $95\%$, for these passages, seeds and prompts. On the meters'
own sweep protocol the best of $64$ anchor draws also reads $0.0000$. A memoriser served through the
windowed meter reaches $0.01$ at $80$ nats a window, and at $125$ through the chat template
(Appendix~\ref{app:windowed}).

\paragraph{Protocol details (\textsc{judged}).} The first two classes carry a
\texttt{Complete the prefix:} header (Table~\ref{tab:promptset}).

Averaging both orders removes position from each score, but it is not neutral between
arms. Read in a single order, the $k=10$ difference is $+0.013$ rather than $+0.0645$
(Appendix~\ref{app:judgemethod}).

This scorer is not vetted: nothing shows that it does not hold the
protected works, and it reads the user's prompt (Appendix~\ref{app:promptblind}).

Judge~B still detects a known
difference. A second draw of the risky model gains $+0.065$ $[+0.0395, +0.090]$ over the anchor, and
$+0.1485$ under judge~G (Appendix~\ref{app:judgevalidity}).

\paragraph{Same event bound: the full answer (\textsc{judged}).} Our answer has three parts.
\begin{itemize}[leftmargin=*,itemsep=2pt]
\item \emph{Selection against the meter at $k=0.5$} survives in the headline configuration. Selection's
bound on the window is stronger by more than sixty orders of magnitude. Its gain is reliably positive while
the meter's is equivalent to zero. Across serving configurations, selection's served level exceeds the
meter's at every window-informative budget of Table~\ref{tab:served}. The exception on gains is the $8$B
at the authors' temperature $0.7$ with repetition penalty $1.1$. There the meter at $k=0.5$ gains
$+0.0825$ $[+0.0605, +0.1045]$ over its anchor, and the gains tie: selection minus meter is $+0.0245$
$[-0.0090, +0.0570]$.
\item \emph{At a matched whole-output certificate}, best-of-$64$ leads the budget granted up front at
$\log 64 = 4.16$ nats by $+0.0785$ $[+0.0445, +0.1115]$, and installments lead the per-token pathwise meter
at $83.18$ nats by $+0.142$ $[+0.1085, +0.1755]$. Comparisons with the windowed meter and with anchored
decoding are across certificate types, since their window certificates condition differently
($^{c}$ against $^{b}$; Appendix~\ref{sec:matched}, Table~\ref{tab:matched}).
\item \emph{Where the meter's certificate is vacuous}, from $k=1$ on for a window, there is no shared event
bound to match on. Selection's bound is still $\approx 10^{-67.6}$, and the meter's is $1$. That
comparison is with the risky model itself, and it is the subject of Section~\ref{sec:served}.
\end{itemize}

\paragraph{The $k=10$ headline and empty answers.} The reward prefers empty drafts: selection serves $41$ empty texts and the meter none.
Judge~B scores those picks near parity ($0.494$) and judge~G above it ($0.634$). Counted as losses, the
difference is inconclusive; a rule that never serves an empty draw keeps it under judge~B but not under
judge~G. Under judge~B it also depends on the scorer: with a gemma scorer it is inconclusive
(Table~\ref{tab:primary}).

\subsection{The served comparison in detail}\label{app:servedtext}

\paragraph{From $k=1$ on, the comparison is with the risky model.} From $k=1$ on, the meter's certificate
excludes nothing about a $50$-token window ($K/S_w \ge 1$; Table~\ref{tab:served}). At $k=10$ it serves the
risky model unchanged on $99.99\%$ of steps and spends a measured $171.3$ nats. The judged comparison is
then between selection and the risky model itself. Its sign depends on how that model is served. We list
the configurations where selection leads, and then every reversal we found.

\paragraph{Where selection leads.} When the $8$B continues text at temperature $1.0$, the risky model
loses. This was measured before the other configurations were registered. The meter gains $+0.0510$
$[+0.025, +0.076]$ at $k=10$, and the gains differ by $+0.0505$ $[+0.0155, +0.0860]$. Selection's KL bound
($3.175$ nats) is about one fifty-fourth of the meter's measured spend ($171.3$ nats). On the $62$ prompts
where both arms' verdicts are order-consistent, the difference is $+0.121$ $[+0.024, +0.222]$. Selection
leads by more at the authors' $70$B pair. At this temperature their base model is fluent but off task, and
is judged no better than the anchor: $+0.019$ $[-0.004, +0.042]$ (Appendix~\ref{app:served}). At their
$0.7$ and $1.1$ the $70$B is $+0.0585$ over the anchor and still loses, by $+0.0485$ $[+0.0145, +0.082]$ at
$k=20$. Against the authors' byte-level decoder, selection at Comma-7B leads at every budget (last block of
Table~\ref{tab:served}).

\paragraph{The reversals.} We found four: three where the meter's certificate is vacuous, and one
(the windowed meter) where it is not but is too large to audit.
\begin{itemize}[leftmargin=*,itemsep=2pt]
\item \emph{Chat template.} Served through its chat template, the $8$B beats its own templateless text by
$0.184$. The $k=10$ meter, which is that model on $99.84\%$ of steps, then wins by $0.138$. The meter's
interval first lies wholly above selection's at $k=3$; they tie at $k=2$. At $k=3$, $K/S_w = 3.75$ and
the certificate excludes nothing about a window. A memoriser served the same way starts leaking only at
$k=3.9$ $[3.0, 4.4]$. So this win comes before measured extraction, but inside its interval
(Appendix~\ref{app:served}). The chat-served model gains $+0.272$ over the anchor where selection gains
$+0.1065$, so selection closes less than half the gap (Appendix~\ref{app:tokenswap}).
\item \emph{Temperature $0.7$ with penalty $1.1$, on the $8$B.} Here even $k=0.5$ is not the anchor. The
meter gains $+0.0825$ $[+0.0605, +0.1045]$ and the gains tie, selection minus meter $+0.0245$
$[-0.0090, +0.0570]$. At $k=10$ the $8$B wins, $-0.081$ $[-0.113, -0.049]$. At their $70$B pair the meter
is its anchor again at $k=0.5$ ($+0.018$ $[-0.0035, +0.040]$), and selection leads by $0.089$
$[0.0575, 0.122]$.
\item \emph{AlpacaEval.} The difference belongs to prefix completion. It holds on our prompts, on
public-domain books and on BookMIA's unseen half. It is unresolved on CoTaEval's news questions and
MT-Bench. It reverses on AlpacaEval, where the meter wins by $0.0329$ (Appendix~\ref{app:workload}).
\item \emph{The windowed meter under the chat template.} Through the template, a windowed meter at
$W = 125$ beats selection by $0.160$ $[0.122, 0.1985]$. At $W=40$ it wins under judge~B,
$-0.0525$ $[-0.0895, -0.0150]$, ties under judge~G, $-0.0295$ $[-0.0725, +0.0145]$, and leaks nothing
through the same template. It bounds the exact window, the median one at $e^{40-159.8}$, and a
near-verbatim variant too. But it bounds the variant only through the factor $e^{40}$. A rights-holder
could audit that by sampling only with more than $3e^{40}$ clean draws, where selection's $\log 64$ needs
$19{,}200$ (Appendix~\ref{app:windowed}).
\end{itemize}
Against a stronger opponent the difference also decays. On the recorded text, from
\texttt{Qwen2.5-1.5B-Instruct} upward every interval covers zero (Table~\ref{tab:opponentladder}).

\paragraph{The scorer binds on TriviaQA (\textsc{measured}).} On TriviaQA the
pointwise reward falls with $n$, so the scorer binds first (appendix Figure~\ref{fig:judgefree},
Table~\ref{tab:judgefree}). There the meter scores $0.618$ against selection's $0.190$, but only at $k=20$,
where its certificate is vacuous (Section~\ref{sec:cost}).

\subsection{Cost in detail}\label{app:costtext}

\paragraph{Forward passes and the clock.} Counted in forward passes, $n=64$ costs $61.3\times$ the meter.
Most of that is the reward model's prefill. On the clock, however, the reward pass is $9.3\%$ and the draws
are the rest. So the lever is $n$, not the scorer. Majority vote, which has no scorer, costs $5.75\times$
the meter's forward passes at $n=32$ (Appendix~\ref{app:scorerfreemain}).

\paragraph{Latency.} Batched into one decode loop, a request takes $0.22\times$ the meter's time at the
authors' $70$B pair and $0.51\times$ at the $8$B pair. This is because selection never runs the risky
model. Against their byte-level decoder the ratio is $0.061\times$. Speculative decoding would raise these
to at best $0.37\times$ and $0.88\times$ (Appendix~\ref{app:speculative}). Batched latency is the operating
point where selection is cheaper; throughput is where it is dearer.

\paragraph{Held to the meter's own compute, selection loses (\textsc{judged}).} The compute-matched cell is
$n=2$ at $0.68\times$. Its judged gain against the meter is $-0.0330$ $[-0.0625, -0.0025]$
(Table~\ref{tab:primary}). This loss survives all three prices we measured (Appendix~\ref{app:anchoronly}).
On TriviaQA (\textsc{measured}), at matched compute ($n=5.57$) selection reads about $0.13$ against the
meter's $0.618$ at $k=20$, and at $11.5\times$ the compute it is still at $0.190$
(Appendix~\ref{app:judgefree}).

\paragraph{What this means for deployment.} Selection does not make a small licensed model match a
frontier model. It improves a licensed, vetted anchor at a certified cost of $\log n$ nats, and pays for it
in draws. No configuration we ran meets all four conditions at once: auditable protection, a practical
query horizon, competitive utility and acceptable compute. With the unvetted Qwen scorer, selection's
judged gain is reliable, but the pooled horizon is $38$ queries per window and the draws cost $21.8\times$
in throughput. With a vetted, prompt-blind scorer the horizon is no longer capped in practice, but the
gain falls to $+0.0405$ under judge~B and is inconclusive under judge~G. At matched compute it loses. The
regime the evidence supports is latency-bound serving, at $0.22$ to $0.51\times$ batched latency, of a
vetted anchor, with either a query cap or a vetted scorer. The binding constraint is a vetted anchor of
sufficient quality. Comma-7B is the largest that passed our vetting screen, and how far larger vetted
anchors would close the gap is untested here.

\subsection{The limitations with their numbers}\label{app:limitmain}

\paragraph{The anchor is the binding constraint (\textsc{measured}, \textsc{judged}).} A served string is
one the safe model would have drawn, so an anchor that cannot do the task gains nothing from selection, at
any $n$. On TriviaQA \citep{joshi2017triviaqa} the meter wins only by serving the risky model at $k=20$
(Section~\ref{sec:breadth}). Selection does not improve facts either. Its FActScore precision is its
anchor's, $0.042$, below every metered arm, though the best of its $64$ draws reaches $0.5054$
(Appendix~\ref{app:hemetrics}). A capability gap is the regime a deployer most wants covered.
\emph{Neither} mechanism covers it with a certificate a rights-holder can audit.

\paragraph{A frontier-quality vetted anchor may not exist (\textsc{measured}).} The largest openly
licensed anchor to pass our vetting check is Comma-7B, far below frontier scale. On web-scale data it is
hard to show that no protected work leaked in, for instance through derived copies. Our certificate does
not remove that difficulty: selection multiplies whatever a contaminated anchor already leaks by up to $n$
(Table~\ref{tab:contam}). An anchor distilled from the risky model would import the very risk the
certificate is meant to exclude. We did not test long agentic rollouts or complex multi-step tasks, where a
small anchor's search space may be the binding limit.

\paragraph{The scorer binds next (\textsc{measured}, \textsc{judged}).} Below the anchor's ceiling the
scorer limits the gain. On GSM8K at $n=64$ the pointwise reward gains $3.4\times$ less than majority vote.
On CoTaEval it \emph{loses} F1, on an infringement event that not even the risky model reaches
(Appendix~\ref{app:selection}). The gain is the scorer's capability. A $0.5$B scorer never reaches the
meter. A $1.5$B one buys $87\%$ of the $7.6$B gain on the judged workload and nothing on GSM8K. Where that
bar sits is open (Appendix~\ref{app:saturation}). A vetted scorer is not yet a substitute for ours: Comma-7B
scoring prompt-blind gains $+0.0405$ under judge~B and $+0.0120$ $[-0.0230, +0.0475]$ under judge~G
(Appendix~\ref{app:promptblind}).

\paragraph{The judged comparison measures the serving configuration as much as the mechanism
(\textsc{judged}).} Against the risky model the ranking reverses under a chat template, on AlpacaEval
\citep{li2023alpacaeval}, and on the $8$B at the authors' temperature $0.7$ with repetition penalty $1.1$
(not at their $70$B pair). It vanishes
against a stronger opponent. It also vanishes when empty answers count as losses, and, under the
pre-specified judge, with a scorer from another family (Table~\ref{tab:served}, Figure~\ref{fig:h2h}).
Under Holm's correction over $21$ rows the headline row does not survive. Every judge is a language model.
The pre-specified judge is order-consistent on about a third of items, and no human rated anything.
Tempering the anchor buys only part of the gain: best-of-$64$ leads it at $T=0.7$ and $0.5$ by $+0.0685$
and $+0.069$ under judge~B, and by $+0.063$ and $+0.060$ under judge~G (Appendix~\ref{app:tempering}).

\paragraph{It is paid for in compute (\textsc{measured}; the matched-compute loss is \textsc{judged}).} Selection at $n=64$ costs $61.3\times$ the
meter's forward passes, and on an anchor-only server its draws cost $21.8\times$ a metered decode. Batched, a request takes
$0.22\times$ the meter's time at the $70$B pair, because selection never runs the risky model. Held to the
meter's own compute, it \emph{loses}. The price is the drawing, not the scorer's one batched pass
(Appendix~\ref{app:saturation}).

\paragraph{What the certificates cover (\textsc{proved}).} The whole-output certificate is $\log n$ per
response and $m\log n$ over $m$ queries. Its audit by clean anchor draws covers one fixed prompt
distribution per work. Extending it to other prompts requires auditing those prompts too. Pooled adaptive
queries are covered only by composition, and the horizon is finite unless the scorer is vetted, an
operational condition the certificate cannot check (Section~\ref{sec:compose}). In installments the
whole-output certificate is $B\log n$. A window is certified only conditional on the served text up to the
last block boundary at or before its start; conditional on the tokens inside the block before it, no bound
holds (Proposition~\ref{prop:installments}).

\paragraph{Short works and narrow coverage (\textsc{measured}).} A certificate is informative only about
works whose surprisal exceeds it. On TriviaQA answers, $\log 64$ reaches $S(x)$ on $35.4\%$ of questions,
against $89.6\%$ for the meter's smallest budget ($36.2\%$ against $83.6\%$ under the anchor tempered to
$0.7$). So $n$ must be set against the shortest work protected,
and a vote gains only where $\log n \ge S$ (Appendix~\ref{app:gainwhere}). On protected quotations,
selection's worst case leaks $0.38\%$ and the meter $69.8\%$ at $k=10$ (Appendix~\ref{app:shortworks}). We
tested no lyrics, whole poems or code. Extraction covers sixteen English novels plus four in French and
German.

\paragraph{The conclusion, with its numbers.} Spent choosing among a safe model's draws, a budget is $\log n$ nats at any output length. Spent in
installments of $L$ tokens, it is $B\log n$ for the whole output, and $(\lceil w/L\rceil+1)\log n$ per
$w$-token window conditional on the served text up to the last block boundary at or before that window's
start. In the headline configuration, at a matched whole-output certificate, selection leads both meters
we could match to it: the budget granted up front at $4.16$ nats and the per-token pathwise meter at
$83.18$ nats. At $1{,}000$ tokens, installments do not resolve against the meters at $20.79$ nats. In the
same configuration, at the one published meter budget that says anything about a window, selection has the
far stronger event bound and a reliably positive judged gain, while the meter's gain is equivalent to zero
by a post-hoc equivalence test. At the authors' sampling settings on the $8$B the meter gains there too,
and the two tie. Where the meter's certificate is vacuous, the comparison is with the risky model, and its
sign depends on how that model is served. Through the chat template a windowed meter wins at $W=125$, and
at $W=40$ under the pre-specified judge only, at certificates too large to audit. Selection costs more
compute, loses at matched compute, and can be no better than its vetted anchor.

\section{Proofs}\label{app:proofs}

\paragraph{Proposition~\ref{prop:selection}: tightness and extensions.}
If $f$ is maximised at a single $y^\star$ with $p_s(y^\star) = p$, then $q(y^\star) = 1-(1-p)^n \to np$ as
$p \to 0$, so no constant below $n$ works in $q(y) \le n\,p_s(y)$ and the $\log n$ charge cannot be
improved without assuming something about $f$ or $p_s$, which is the point of assuming neither. For a
deterministic $g$, $\Pr_q[g(y) \in A] = \Pr_q[y \in g^{-1}(A)] \le n\Pr_{p_s}[g(y) \in A]$. A rule that
draws one candidate at a time and stops at any data-dependent time, capped at $n_{\max}$, serves one of at
most $n_{\max}$ i.i.d.\ draws, and the union bound over the first $n_{\max}$ gives $\log n_{\max}$.

\paragraph{Proposition~\ref{prop:installments}: selection in installments.}\label{app:installmentsproof}
\emph{The assumption.} Write $Z_{<j}$ for all internal randomness generated before block $j$: the unserved
candidates, their scores, any rollouts and any external coins. The served history $h_{<j}$ is a function
of $Z_{<j}$. The fresh-draws assumption is that, given $Z_{<j}$, the $n$ candidates at block $j$ are
i.i.d.\ $p_s(\cdot \mid h_{<j})$. It fails if a block re-serves text drawn earlier, for instance the
continuation of a rollout chosen for its score, since that text is no longer an anchor draw given what the
rule has seen. The assumption cannot be dropped: if the rule at block $j-1$ picks a two-block rollout and
re-serves its continuation at block $j$, Remark~\ref{rem:installments}, with a block boundary after its
first token, becomes a violation of part~(ii) at block $j$, with a ratio of about $999.5 > n = 2$. The
installment rule of Appendix~\ref{app:blockwise}, as described there, draws new anchor
continuations at every block and uses its rollouts only to score them.

\emph{Step 1: one block, given everything.} Fix $Z_{<j} = z$, which fixes $h_{<j}$. Block $j$ is served
the value $b$ only if some candidate equals $b$. By the union bound, as in
Proposition~\ref{prop:selection},
\[
  \Pr\bigl[\text{block } j = b \bigm| Z_{<j} = z\bigr] \;\le\; \sum_{i \le n}\Pr[c_i = b \mid Z_{<j} = z]
  \;=\; n\,p_s(b \mid h_{<j}),
\]
whatever the rule does with the scores, the rollouts, $z$ or a coin.

\emph{Step 2: marginalise the internal state.} Block $j$ is served from the block-$j$ candidates by a rule
that may also use $Z_{<j}$ and fresh block-$j$ randomness, and $Z_{<j}$ determines $h_{<j}$. The served law's conditional given only the served
history is an average of Step~1 over the internal states consistent with that history:
$q(b \mid h_{<j}) = \mathbb{E}\bigl[\Pr[\text{block } j = b \mid Z_{<j}] \bigm| h_{<j}\bigr] \le
n\,p_s(b \mid h_{<j})$. The right side depends on $h_{<j}$ alone, so the bound survives the average.

\emph{Step 3: product over blocks.} For block values $b_j,\dots,b_{j+r-1}$, the chain rule for the served
law and Step~2 at each block give
\begin{align*}
  q(b_j,\dots,b_{j+r-1} \mid h_{<j}) &= \prod_{i=j}^{j+r-1} q\bigl(b_i \bigm| h_{<j}, b_j,\dots,b_{i-1}\bigr)
  \le n^{r}\prod_{i=j}^{j+r-1} p_s\bigl(b_i \bigm| h_{<j}, b_j,\dots,b_{i-1}\bigr)\\
  &= n^{r}\,p_s(b_j,\dots,b_{j+r-1} \mid h_{<j}).
\end{align*}
If the output has ended before block $i$, block $i$ is empty under both laws and its factor is
$1 \le n$. With $j=1$ and $r=B$ this is part~(i).

\emph{Step 4: sum over the event.} An event $E$ determined by blocks $j,\dots,j+r-1$ is a set of values of
those blocks. Summing Step~3 over the set gives part~(ii).

\emph{Part (iii).} A $w$-token window at positions $s,\dots,s+w-1$ meets blocks $j = \lceil s/L\rceil$
through $j' = \lceil (s+w-1)/L\rceil$, so it meets $r = j'-j+1 \le \lceil w/L\rceil + 1$ blocks, and
$r = \lceil w/L\rceil$ when $s-1$ is a multiple of $L$. Also $r \le B$. Any event measurable in the
window is determined by blocks $j,\dots,j'$, and $h_{<j}$ is the served text up to the last block boundary
at or before $s$. Part~(ii) then gives the stated certificates. \qed

\emph{Part (iv), and why the conditioning cannot move inside the block.} Conditioning further on the
served tokens of block $j$ before position $s$, call them $u$, divides by $\Pr_q[u \mid h_{<j}]$. Step~3
bounds the numerator, but nothing bounds this denominator from below by its anchor counterpart: the rule
can make a prefix rarer than the anchor does. In Remark~\ref{rem:installments}, $p_s(y_1 = 1) = a$, while
$q(y_1 = 1) = [a(1-b)]^2 + 1-(1-ab)^2$ is about $a^2 + 2ab$. To first order,
$q(y_2 = 1 \mid y_1 = 1) \approx 2b/(a+2b)$, so its ratio to $p_s(y_2 = 1 \mid y_1 = 1) = b$ is about
$2/(a+2b)$, which tends to $1/b$ as $a \to 0$. The same block can be the first of two, with the second
drawn from the anchor alone, so the failure is not an artefact of a single block. A reviewer of an earlier
version supplied this counterexample. We checked it by exact rational enumeration, and the same
enumeration confirms parts~(i) to~(iii) on small random instances whose rules use earlier unserved
candidates and a coin.

\paragraph{The certificate over $m$ adaptive queries.}\label{app:compose}
Let a user issue queries $x_1,\dots,x_m$, each answered by a selection rule over $n$ draws and each chosen
from everything served so far. Write $q_j(\cdot \mid \mathcal{H}_{j-1})$ for the served law at round $j$
and $p_{s,j}$ for the anchor's. The $n$ draws at round $j$ are i.i.d.\ $p_{s,j}$ whatever the history is,
so Proposition~\ref{prop:selection} gives $q_j(y \mid \mathcal{H}_{j-1}) \le n\,p_{s,j}(y \mid
\mathcal{H}_{j-1})$ pointwise, and multiplying along any realised history,
\[
  Q(y_1,\dots,y_m) \;=\; \prod_{j \le m} q_j(y_j \mid \mathcal{H}_{j-1})
  \;\le\; n^{m} \prod_{j \le m} p_{s,j}(y_j \mid \mathcal{H}_{j-1}) \;=\; n^{m}\,P_s(y_1,\dots,y_m),
\]
hence $D_\infty(Q \,\|\, P_s) \le m \log n$ exactly, with no slack from adaptivity. A per-query
access-freeness guarantee does not compose this way in general \citep{cohen2025blameless}, and a
per-token meter composes to $mkT$, growing in the length of every response. Under a $400$-nat per-user cap,
about half the median protected target's $849$
nats, selection admits $192$ responses at $n=8$ at the pathwise charge, and the sharper KL bound of
$1.204$ nats would give $332$; a meter at $k=3$ charged its measured $165.0$ nats admits $2$, and none at
the $600$ it is certified at. The cap protects a whole target: it exceeds a $50$-token window's $159.8$
nats, so for a window the composed certificate lapses long before the cap is reached
(Table~\ref{tab:certificate}). Accounting per work rather than per user does not relax this: users can pool their transcripts, so a per-work budget is shared by everyone who can query the deployment, and the responses that exhaust a window's certificate become a total rather than a per-user allowance. Only a scorer that neither holds the work nor takes a user's instructions removes the channel the cap exists for. For an untrusted scorer the cap is $\lfloor S/\log n\rfloor$ queries for the shortest event protected: at $n=8$ and $n=64$, $15$ and $7$ for a $10$-token window of median surprisal $31.97$ nats, $76$ and $38$ for a $50$-token one, and for a TriviaQA answer, median $5.78$ nats, $2$ and $1$, with $20.8\%$ and $36.2\%$ of answers allowing none (\texttt{results/\allowbreak query\_\allowbreak caps.csv}).

\paragraph{Proposition~\ref{prop:threshold}.}
Write $p = \Pr_{p_s}[E] = e^{-S}$ and $a = \Pr_q[E]$. R\'enyi divergence satisfies data processing
\citep{vanerven2014renyi}, so passing both laws through the indicator of $E$ gives
$d_\alpha(a\,\|\,p) \le D_\alpha(q\,\|\,p_s) \le K$, where
$d_\alpha(a\,\|\,p) = \frac{1}{\alpha-1}\log\bigl(a^{\alpha}p^{1-\alpha} + (1-a)^{\alpha}(1-p)^{1-\alpha}\bigr)$
for $\alpha \in (1,\infty)$, the binary KL at $\alpha = 1$ and $\log\max\{a/p,\,(1-a)/(1-p)\}$ at
$\alpha = \infty$. Conversely the law $q_a = a\,p_s(\cdot \mid E) + (1-a)\,p_s(\cdot \mid E^{c})$ has a
constant likelihood ratio on $E$ and on its complement, so $D_\alpha(q_a\,\|\,p_s) = d_\alpha(a\,\|\,p)$
exactly. Hence $b_\alpha(K) = \max\{a : d_\alpha(a\,\|\,p) \le K\}$, and since $d_\alpha(\cdot\,\|\,p)$
increases on $[p,1]$, $b_\alpha(K) = 1$ if and only if $d_\alpha(1\,\|\,p) \le K$. At $a = 1$ every order
gives $d_\alpha(1\,\|\,p) = \log(1/p) = S$. \qed\ The familiar closed forms relax this inversion and break
at the same point: the change of measure $a \le e^{K}p$ is exact at $\alpha = \infty$, while
\citet{mironov2017renyi}'s $a \le (e^{K}p)^{(\alpha-1)/\alpha}$ is an upper bound on $b_\alpha(K)$ for
$\alpha \in (1,\infty)$, not $b_\alpha(K)$ itself.

\paragraph{Below the threshold the orders part company.}\label{app:belowthreshold}
Write $S = -\log p_s(E)$ and take $K < S$. At $\alpha = \infty$ the bound is exactly $e^{K-S}$, and for
$\alpha \in (1,\infty)$ it is at most $e^{-(1-1/\alpha)(S-K)}$, Mironov's relaxation, so both fall
exponentially in the shortfall. At $\alpha = 1$
it is only linear: every $q$ with $d(q(E) \,\|\, p_s(E)) \le K$ has $q(E) \le (K + \log 2)/S$, since
$d(a \,\|\, e^{-S}) \ge aS - \log 2$, and the law that emits $x$ with probability $a$ and otherwise follows
$p_s$ costs $aS - h(a) \le aS$ to first order in $e^{-S}$, so $a = K/S$ is within budget. A KL budget of
half a work's surprisal therefore permits the work half the time. At $k = 0.5$, $K = 100$ nats against a
$50$-token window's $159.8$ permits the window with probability $0.630$ under the KL charge and about $e^{-60}$ under the pathwise one (Table~\ref{tab:certificate}). A KL certificate that bounds
the window by $\delta$ needs $K \le \delta S$, which by Proposition~\ref{prop:sparse} leaves at most
$\delta S/\varepsilon$ steps above $\varepsilon$ in expectation, $1.6$ at $\delta = 0.01$ and $\varepsilon = 1$.

\paragraph{The threshold for a window event.}\label{app:windowvacuity}
Proposition~\ref{prop:threshold} holds verbatim for $E_w$, ``reproduce a $w$-token window of the work'',
and every extraction number in this paper is a window event. The scored event is a union over windows, at
least as probable as any member, so each threshold in Table~\ref{tab:windowvacuity} is an upper bound on
where the certificate for the scored event goes vacuous. The same holds for the union over the at most
$151$ positions at which a $50$-token window can start in a $200$-token response: Propositions~\ref{prop:selection}
and~\ref{prop:threshold} bound the union itself, so no position factor enters either certificate, only a
closed-form price of the anchor's rate, which covers one position; the union's rate under the anchor is
the sampled quantity of Section~\ref{sec:compose}.

\begin{table}[ht]
\centering\footnotesize
\caption{Vacuity thresholds by event size, from the audited anchor's per-token surprisal on $758$
passages of $16$ novels (median $3.197$ nats per token). The budget column is the $k$ at which
$K = kT_{\max}$ reaches the threshold at $T_{\max}=200$ (\texttt{results/\allowbreak window\_\allowbreak vacuity.csv}).}
\label{tab:windowvacuity}
\begin{tabular}{@{}lrrr@{}}
\toprule
event $E$ & $S(x)$, nats & vacuous at $k$ & $\log 64$ sits below by \\
\midrule
$10$-token window   & $32.0$  & $0.160$ & $7.7\times$ \\
$20$-token window   & $63.9$  & $0.320$ & $15.4\times$ \\
$50$-token window   & $159.8$ & $0.799$ & $38.4\times$ \\
whole passage (median, $64$ tokens) & $204.8$ & $1.024$ & $49.2\times$ \\
$100$-token window  & $319.7$ & $1.598$ & $76.9\times$ \\
\bottomrule
\end{tabular}
\end{table}

\noindent At $k = 3, 5, 10, 20$, the certificate is vacuous for every event in the table; the $50$-token
event is vacuous at every $k \ge 0.8$, below the evaluation's smallest budget $k=1$; and selection's
$\log 64 = 4.159$ sits below every threshold.

\paragraph{Proposition~\ref{prop:sparse}.}
\noindent\emph{Proposition~\ref{prop:sparse}, restated.} Let $q$ be any causal policy,
$q(y) = \prod_{t \le T} q_t(y_t \mid y_{<t})$, with sequence budget
$D_{\mathrm{KL}}(q \,\|\, p_s) \le K$, and let
$N_\varepsilon = \#\{t \le T : D_{\mathrm{KL}}(q_t(\cdot \mid y_{<t}) \,\|\, p_{s,t}) > \varepsilon\}$.
Then $\mathbb{E}_q[N_\varepsilon] \le K/\varepsilon$ for every $\varepsilon > 0$.

\emph{Proof.} The chain rule for an autoregressive factorisation is exact,
\[
D_{\mathrm{KL}}(q \| p_s) = \sum_{t \le T} \mathbb{E}_{y_{<t} \sim q}[D_{\mathrm{KL}}(q_t \| p_{s,t})],
\]
with every term nonnegative. Markov's inequality at each step gives
\[
\Pr_q[D_{\mathrm{KL}}(q_t \| p_{s,t}) > \varepsilon] \le \mathbb{E}_q[D_{\mathrm{KL}}(q_t \| p_{s,t})]/\varepsilon,
\]
and summing over $t$ bounds $\mathbb{E}_q[N_\varepsilon]$ by $K/\varepsilon$. \qed

\emph{Surely, for the deployed class.} The token bucket caps each trajectory's realised charges,
$\sum_t a_t \le K$, the accounting under which no violation occurs in more than $100{,}000$ trajectories,
so at most $K/\varepsilon$ steps of any single served trajectory carry a charge above $\varepsilon$. At the
authors' $k=3$ and $T_{\max}=200$ this is $600/\varepsilon$ against $200$ steps: it constrains nothing below
$\varepsilon = 3$ nats in a step, the first horn in another currency. \emph{At finite $T$, in probability.}
Markov on the count gives $\Pr_q[N_\varepsilon \ge K/(\varepsilon\delta)] \le \delta$, informative only when
$\varepsilon > K/(\delta T)$. For a rate, $K = kT$, the threshold is $\varepsilon > k/\delta$, independent of
$T$: at $k = 3$ and $\delta = 0.05$ the guarantee is silent below $60$ nats in a single step at every
length. For a bounded $K$ it is $\varepsilon > K/(\delta T) \to 0$: the decoder is the anchor at all but
$O(1)$ steps.

\emph{Selection does not evade the chain rule, and no separation is provable at the level of laws.}\label{app:noseparation}
It is tempting to argue that a selection rule is not a causal policy at all, every token being drawn from $p_{s,t}$ so that $N_\varepsilon = 0$; that describes the sampler, not the served law. Selection's served law factorises like any other, so the proposition applies to it with $K = 3.175$ nats
at $n=64$; its conditionals differ from $p_{s,t}$ because conditioning on which draw won reweights every
step, but they are not \emph{causally computable}: $q_t(\cdot \mid y_{<t})$ depends on the scores of
complete candidates. The statement ``no causal policy with an $O(1)$ budget matches selection's utility''
is false as a statement about laws: every law on sequences factorises, and Lemma~\ref{thm:nfl}'s
optimum $q^* \propto p_s e^{\lambda U}$ has conditionals $p_{s,t}(y_t)\,V_t(y_{\le t})/V_{t-1}(y_{<t})$ with
$V_t = \mathbb{E}_{p_s}[e^{\lambda U} \mid y_{\le t}]$. What separates a meter from both is information: its
kernel at step $t$ is a function of $p_{s,t}$, $p_{r,t}$ and its bucket, while $V_t$ is an expectation over
continuations neither next-token distribution contains. Proposition~\ref{prop:imitation} is the
separation that information structure supports.

\paragraph{Lemma~\ref{thm:nfl}.}
The chain rule gives $D_{\mathrm{KL}}(q \| p_s) = \mathbb{E}_q[\sum_t D_{\mathrm{KL}}(q_t \| p_{s,t})]$, and
for the audited decoder the step charge is exactly $c_1(\theta_t) = D_{\mathrm{KL}}(p_{\theta_t} \| p_{s,t})$,
so a per-trajectory budget $\sum_t c_t \le K$ gives $D_{\mathrm{KL}}(q \| p_s) \le K$; for $\alpha > 1$ the
charge dominates the KL \citep{vanerven2014renyi}. (Checked, not assumed: over $1{,}200$ trajectories and
$240{,}000$ steps at $k=3$ the largest gap between $\sum_t a_t$ and the recorded spend is
$1.5\times10^{-4}$ nats.) Donsker--Varadhan \citep{donsker1975asymptotic} gives, for bounded $U$ and every
$\lambda$, $D_{\mathrm{KL}}(q \| p_s) \ge \lambda\, \mathbb{E}_q[U] - \log \mathbb{E}_{p_s}[e^{\lambda U}]$;
the supremum over $\lambda$ is the Cram\'er rate function \citep{dembo1998large}, so
$K \ge \Lambda^*_s(\mathbb{E}_q[U])$. \qed

\paragraph{The bucket-metered class and its rates.}
Write $\ell = \log p_r - \log p_s$ and $Z(u) = \mathbb{E}_{p_s}[e^{u\ell}]$ along the geometric path
$p_\theta \propto p_s^{1-\theta}p_r^{\theta}$. The charge at order $\alpha$ is
\begin{equation}
c_\alpha(\theta) \;=\; D_\alpha(p_\theta \,\|\, p_s)
 \;=\; \frac{\log Z(\alpha\theta) - \alpha \log Z(\theta)}{\alpha - 1},
\label{eq:renyi-path}
\end{equation}
with $c_1(\theta) = \theta Z'(\theta)/Z(\theta) - \log Z(\theta)$, the deployed charge, and
$c_\infty(\theta) = \theta \max_y \ell(y) - \log Z(\theta)$, the pathwise one. A decoder of the class
serves, at each step, the largest $\theta_t$ whose charge fits the allowance $k_t$ of a token bucket of rate
$k$ started at a debt $\delta$. For a work $x$ of $N$ characters with per-token safe-model surprisals
$s_1,\dots,s_T$, $N_t$ characters after $t$ tokens and $S(x) = \sum_t s_t$, three rates govern the audit:
\begin{equation}
\underbrace{s(x) = \frac{S(x)}{N}}_{\text{surprisal rate}}, \qquad
\underbrace{k_{\mathrm{crit}}(x) = \max_{t \le T} \frac{\sum_{i \le t} s_i + \delta}{N_t}}_{\text{Loynes running maximum}}, \qquad
\underbrace{c_{\mathrm{use}} = \operatorname{med}_{y \sim p_r} \frac{\log p_r(y) - \log p_s(y)}{|y|}}_{\text{divergence rate of ordinary traffic}} ,
\label{eq:rates}
\end{equation}
per character where safe models with different vocabularies are compared and per token where a rate meets
a budget.

\paragraph{Proposition~\ref{prop:imitation}.}
The charge $c$ is nondecreasing on $[0,1]$ with $c(0) = 0$, and the decoder serves
$\theta_t = \max\{\theta : c(\theta) \le k_t\}$. At a slack step $c(1) \le k_t$, so $\theta_t = 1$: the
decoder serves $p_{r,t}$ and is charged $D_{\mathrm{KL}}(p_{r,t} \| p_{s,t})$. By the chain rule,
$D_{\mathrm{KL}}(q \| p_s) = \mathbb{E}_q[\sum_t D_{\mathrm{KL}}(p_{\theta_t} \| p_{s,t})]$. Dropping the
nonnegative terms of the other steps gives part~(i). For part~(ii), $\Lambda^*_s$ is convex, vanishes at
$\mathbb{E}_{p_s}[U]$ and is nondecreasing on $[\mathbb{E}_{p_s}[U], u_{\max}]$, where
$\Lambda^*_s(u_{\max}) = -\log\Pr_{p_s}[U = u_{\max}] \le \log(1/\pi)$. Since
$\mathbb{E}_{p_s}[U] \le \mathbb{E}_q[U] \le u_{\max}$, it follows that
$\Lambda^*_s(\mathbb{E}_q[U]) \le \log(1/\pi)$. Combining this with part~(i) and
$\mathbb{E}_q[I_T] \ge \gamma T$ gives
$D_{\mathrm{KL}}(q\|p_s) \ge \gamma T \ge \Lambda^*_s(\mathbb{E}_q[U])\cdot\gamma T/\log(1/\pi)$. The proviso in the statement is substantive: a utility attained by one exact long
string has $\pi = e^{-\Theta(T)}$, hence $\log(1/\pi) = \Theta(T)$ (and $\Lambda^*_s(u_{\max}) = \Theta(T)$), so the factor $\gamma T/\log(1/\pi)$ is
$\Theta(1)$ and no separation survives. Winning a judged comparison is not of that
kind: the anchor's own unassisted draw wins both presentation orders on $2.2\%$ of prompts, so winning every one, at any length, would cost an optimal policy $3.82$ nats, and the decoder spends $44.9$ times that to reach parity; on a single-order pass, where a win in the one order shown comes on over a quarter of prompts, the same reading is $1.30$ nats and $132$ times (Appendix~\ref{app:frontierratio}). \qed

\paragraph{Protection outruns certification.}
\begin{proposition}\label{prop:outrun}
Take $s(x)$ and $k_{\mathrm{crit}}(x)$ of Eq.~\eqref{eq:rates} per token. Under the banking rule, if the
decoder emits each token of $x$ with probability at least $1-\eta$ then
$k \ge (1-\eta)\,k_{\mathrm{crit}}(x) - \log 2$, and $k_{\mathrm{crit}}(x) \ge s(x) + \delta/T$ for every work,
with equality iff the running ratio is maximised at the last token.
\end{proposition}
\emph{Proof.} By data processing on the indicator of the true token, the step charge is at least the binary
divergence between the served and reference probabilities of that token,
$c_i \ge d(q \,\|\, e^{-s_i}) \ge (1-\eta)s_i - \log 2$ whenever $q \ge 1-\eta$. The banking rule
\citep[Algorithm~1]{he2026anchored} caps the charge on the first $t$ tokens at $tk - \delta$, so
$tk \ge (1-\eta)\sum_{i \le t} s_i + \delta - t\log 2$, which is at least
$(1-\eta)(\sum_{i \le t} s_i + \delta) - t\log 2$; dividing by $t$ and maximising over $t$ gives the
first claim, and a running maximum dominates its endpoint $(S(x)+\delta)/T$. \qed Equality holds exactly when the running ratio peaks at the last token; under a constant rate with $\delta > 0$ it is $s(x) + \delta/t$, strictly decreasing, so the maximum sits at $t=1$ and the inequality is strict.
$k_{\mathrm{crit}}$ is the Loynes running maximum of a queue's workload \citep{loynes1962stability},
equivalently the smallest rate of a $(\sigma,\rho)$ arrival curve admitting the surprisal process
\citep{cruz1991calculus,leboudec2001network}: a budget publishes a \emph{drift} where protection is a
\emph{workload maximum}. It bounds regime boundaries rather than screening works: as a predictor of the
onset it misses by $21.9\%$ against a no-change null's $12.3\%$.

\section{Protocol}\label{app:protocol}\label{app:sel-protocol}

\paragraph{Workload.}\label{app:promptset}
The judged pass uses $500$ prompts in three classes (Table~\ref{tab:promptset}); none is drawn from the
protected corpus, none shares a word $8$-gram with any protected passage or names any of its sixteen works
(\texttt{results/\allowbreak prompt\_\allowbreak set\_\allowbreak profile.csv}), and all three sets are
released with the code, available at \codeurl{}. The anchor's order-averaged win rate against the opponent runs from $0.3667$ on
biographies to $0.5067$ on story premises.

\begin{table}[ht]
\centering\footnotesize
\caption{The $500$ judged prompts, by class, with the harness's name for each in brackets: questions written
for this project (\texttt{neutral}), biography requests \emph{Tell me a bio of X.} over FActScore's entities
\citep{min2023factscore} (\texttt{factual}), and r/WritingPrompts premises \citep{fan2018writingprompts}
(\texttt{creative}). Lengths in words, median with quartiles and range; \emph{header} counts prompts served
with \texttt{Complete the prefix:}; \emph{anchor wins} is the anchor's $n=1$ completion against the
opponent, order-averaged.}
\label{tab:promptset}
\setlength{\tabcolsep}{4pt}
\begin{tabular}{@{}llrrllr@{}}
\toprule
class & source & $n$ & header & prompt, words & opponent, words & anchor wins \\
\midrule
questions & written here & $200$ & $200$ & $10$ $[9, 12]$, $6$--$21$ & $158$ $[147, 166]$ & $0.4575$ \\
biographies & FActScore & $150$ & $0$ & $7$ $[7, 7]$, $6$--$9$ & $151.5$ $[143, 159]$ & $0.3667$ \\
story premises & r/WritingPrompts & $150$ & $150$ & $23$ $[17, 35]$, $6$--$40$ & $157$ $[146, 168]$ & $0.5067$ \\
\bottomrule
\end{tabular}
\end{table}

\paragraph{The scorer, and the one that failed.}\label{app:currency}
The deployed reward is $\log p(\text{Yes}) - \log p(\text{No})$ from \texttt{Qwen2.5-7B-Instruct} on one
fixed template; it never consults the risky model. The first scorer we specified was the risky model's own
per-token log-likelihood, the objective anchored decoding tilts toward. At $n=8$ it moves judged utility
from $0.319$ to $0.313$, $-0.006$ against a band of $+0.077$; selecting by the judge's own verdict among the
same eight draws reaches $+0.488$ (circular, an upper bound), and selecting by judge~A while scoring with
judge~B $+0.081$ $[+0.034, +0.130]$ (order-averaging it leaves $+0.0405$
$[+0.019, +0.063]$), so the utility is in the draws and the likelihood does not find it.
Within a prompt, over $5{,}687$ pairs the judge placed differently, the per-token likelihood is concordant
with the judge on $0.526$ $[0.504, 0.548]$, the summed likelihood on $0.478$ $[0.454, 0.500]$ and the
completion's length on $0.537$ $[0.514, 0.560]$: no better than length, and not worse, the difference being
$+0.011$ $[-0.023, +0.046]$ (\texttt{results/\allowbreak currency\_\allowbreak auc.csv}).

\paragraph{The scorer prefers an empty completion.}\label{app:empties}
On the recovered text $13.9\%$ of the headline pool's $32{,}000$ draws are empty, and the reward scores them
above the rest (median $-18.75$ against $-26.12$), so best-of-$n$ serves an empty text on $12.8\%$ of prompts
at $n=1$, $21.8\%$ at $n=8$ and $8.2\%$ at $n=64$. The judge does not score those as losses, and at $n \ge 8$
the preference costs selection level rather than flattering it (prompts served empty are judged $0.4531$, $0.4771$ and $0.4939$ at $n=1$, $8$ and $64$, against $0.4536$, $0.5211$ and $0.5605$ for the rest). Excluding empty candidates from the argmax raises the
fresh-seed headline difference from $+0.0495$ to $+0.0695$ $[+0.0385, +0.0995]$
(\texttt{results/\allowbreak empty\_\allowbreak preference.csv}). On the repaired headline pass (post hoc, no band),
scoring every empty served text as a loss moves the difference from $+0.0505$ to $+0.0155$
$[-0.0245, +0.0555]$: selection serves $41$ empty texts at $n=64$ and the meter none, against $64$ and $62$
for the two anchor controls. Serving the best \emph{non-empty} draw, a selection rule over the same $n$
draws and so certified at the same $\log n$, changes $41$ picks; judged as the judge scores them it leaves
the difference at $+0.0505$, since the judge rates those $41$ replacements as it rated the empty texts,
and with empty answers as losses it keeps $+0.056$ $[+0.0195, +0.093]$, the served levels differing by
$+0.0655$ $[+0.039, +0.0915]$ (\texttt{results/\allowbreak empty\_\allowbreak as\_\allowbreak loss.csv},
\texttt{results/\allowbreak nonempty\_\allowbreak rule.csv}). Every GPU was held by another job when the
$41$ replacements were judged, so judge~B ran on CPU in float32; re-judging the $41$ committed picks the
same way reproduces their committed levels on $37$. Re-judged in one pass per configuration on one machine,
the difference keeps its reading under the non-empty rule at all eight configurations of
Table~\ref{tab:served} we re-ran, the headline at $+0.058$ $[+0.0225, +0.093]$ against the committed
rule's $+0.059$ $[+0.0235, +0.094]$ in the same pass (Table~\ref{tab:nonempty},
\texttt{results/\allowbreak nonempty\_\allowbreak tables.csv}).

\begin{table}[ht]
\centering\footnotesize
\caption{The non-empty rule at eight configurations of Table~\ref{tab:served}: the difference of gains
(selection's over its $n=1$ control minus the meter's over its anchor), judge~B, both orders, all eight
passes on one machine, under the committed rule and under the rule that serves the best non-empty draw
($41$, $14$ and $30$ picks change in the three pools). Last column: the committed rule's difference here
minus the same row's pass behind Table~\ref{tab:served}; order averaging draws nothing, so a shift is how
the judge ran, not a re-roll.}
\label{tab:nonempty}
\setlength{\tabcolsep}{4pt}
\begin{tabular}{@{}lccc@{}}
\toprule
configuration & committed rule & non-empty rule & shift \\
\midrule
released $8$B, $k=10$ (headline) & $+0.059$ $[+0.0235, +0.094]$ & $+0.058$ $[+0.0225, +0.093]$ & $+0.0085$ \\
$70$B base, $k=20$ & $+0.0815$ $[+0.0485, +0.1155]$ & $+0.0805$ $[+0.0475, +0.113]$ & $-0.001$ \\
$70$B base, $k=0.5$ & $+0.1085$ $[+0.078, +0.1395]$ & $+0.1075$ $[+0.0765, +0.1375]$ & $+0.0035$ \\
chat template, $k=10$ & $-0.145$ $[-0.184, -0.1065]$ & $-0.1485$ $[-0.187, -0.11]$ & $+0.0035$ \\
chat template, $k=3$ & $-0.051$ $[-0.0885, -0.0135]$ & $-0.0545$ $[-0.091, -0.0175]$ & $0$ \\
temperature $0.7$, $8$B, $k=10$ & $-0.081$ $[-0.113, -0.049]$ & $-0.0825$ $[-0.114, -0.051]$ & $0$ \\
temperature $0.7$, $70$B, $k=20$ & $+0.0485$ $[+0.0145, +0.082]$ & $+0.047$ $[+0.014, +0.081]$ & $0$ \\
AnchoredByte, $k=0.1$ & $+0.1385$ $[+0.1155, +0.162]$ & $+0.1475$ $[+0.125, +0.1705]$ & $0$ \\
\bottomrule
\end{tabular}
\end{table}

\paragraph{The judge.}\label{app:judgemethod}
Every judged number compares an arm's served text with the fixed opponent's on the same prompt. The judge
receives one user turn, in its own chat template:
\begin{quote}\footnotesize\ttfamily\raggedright
You are grading two candidate responses to the same instruction.\\
Instruction: \{prompt\}\quad Response A: \{a\}\quad Response B: \{b\}\\
Which response is the better answer to the instruction? Judge helpfulness, coherence and factual
accuracy. Ignore length and formatting. If they are of equal quality, answer Tie. Reply with
exactly one word: A, B, or Tie.
\end{quote}
Fields are truncated to $1{,}200$ characters, the prompt is the corpus prompt without the harness header,
decoding is greedy with at most four new tokens, and the verdict is the first letter (anything else a
tie). The cut reaches $19$ of the opponent's $500$ texts, $20$ of the meter's at $k=10$, $2$ of selection's and
no prompt; shown whole, the headline difference reads $+0.062$ $[+0.0265, +0.0975]$ against $+0.059$ cut, on
the same machine, and $+0.0217$ $[-0.0517, +0.0933]$ on the biographies alone
(\texttt{results/\allowbreak full\_\allowbreak response.csv}, registered). An order-averaged score shows the item in both orders and averages the two
(\texttt{analysis/\allowbreak order\_\allowbreak averaged\_\allowbreak h2h.py}). \textbf{Position dominates a
single order} \citep{zheng2023judging}: the same two texts win $261$ of $500$ shown second and $24$ shown
first, only $29.2\%$ of items get a consistent verdict, the first slot wins $12.7\%$ of decided
comparisons, and a single-order level moves when other arms join
the sweep (the audited anchor's $n=64$ gain read $+0.142$ and $+0.076$ on byte-identical text). Order
averaging draws nothing and reproduced exactly, $+0.1045$ $[+0.0820, +0.1280]$ both times, which is why every
head-to-head here is order-averaged and single-order gains are read only within one sweep. The correction
is asymmetric: read in one order against the order-averaged construction, selection at $n=64$ moves
$+0.110$ to $+0.1045$ and the meter at $k=10$ moves $+0.097$ to $+0.0400$, so the paired difference moves
$+0.013$ to $+0.0645$. We had predicted the opposite, both arms deflating together, and it was wrong; the
likely reason, offered rather than claimed, is that at $k=10$ the meter serves $p_r$ nearly unchanged,
leaving the judge two near-identical texts and little but slot order. The same arm found a defect behind
earlier numbers: both judging scripts showed the judge a prompt reconstructed as the served text minus the
generation, whose split point moves with how much each arm generated, so for the same item three arms
disagreed about it in $455$ of $500$ cases. Every judged number here takes the prompt from the corpus, and
under one true prompt the single-order difference is $+0.013$, not the $+0.070$ the arm-specific prompts
implied. A judged gain
re-run on a different $n$-grid moves by $0.04$ to $0.07$; the head-to-head escapes that floor by being
paired within one pass. Judge~C is the opponent's own checkpoint, \texttt{Llama-3.1-8B-Instruct}, a self-preference risk
\citep{panickssery2024llm}.

\paragraph{Tempering the anchor instead of selecting.}\label{app:tempering}
Tempering is a per-token tilt of the untempered anchor, so if it alone bought selection's gain a deployer could
have had it by choosing another reference law (Section~\ref{sec:method}). Registered before any token
(\texttt{results/\allowbreak onset\_\allowbreak prediction\_\allowbreak tempering.md}), the anchor alone was drawn at
temperature $0.7$, at $0.5$ and at the authors' $0.7$ with penalty $1.1$ on the $500$ prompts and judged against the
committed opponent in both orders: judge~B on the machine type that judged Table~\ref{tab:matched}, judge~G with every
arm re-judged on one H100 host, where the committed levels reproduced exactly. Tempering buys part of the gain,
about a third under judge~B and about two thirds under judge~G, and best-of-$64$ leads every tempered anchor under
both, as predicted for judge~G (Table~\ref{tab:tempering}). The tempered anchors also serve more empty answers,
which judge~G scores at $0.68$ to $0.74$, so their gains under it are if anything flattering. Top-$p$
truncation was not run: the decoder has no nucleus option.

\begin{table}[ht]
\centering\footnotesize
\caption{\textbf{Tempering the anchor instead of selecting}, $500$ prompts against the committed opponent, both
presentation orders. \emph{Gain}: level minus that of one untempered anchor draw (selection's $n=1$), with $95\%$
interval; \emph{empty}: empty completions served. Below the rule, best-of-$64$ at $T=1.0$ minus the tempered anchor,
paired over prompts. Judge~B on the machine type of Table~\ref{tab:matched}; judge~G re-judged every arm on one H100
host.}
\label{tab:tempering}
\setlength{\tabcolsep}{3pt}
\begin{tabular}{@{}lrll@{}}
\toprule
arm & empty & gain, judge~B & gain, judge~G \\
\midrule
best of $64$, $T=1.0$ & $41$ & $+0.1015$ $[+0.0770, +0.1260]$ & $+0.1760$ $[+0.1400, +0.2125]$ \\
anchor, $T=0.7$ & $138$ & $+0.0330$ $[+0.0095, +0.0575]$ & $+0.1130$ $[+0.0775, +0.1480]$ \\
anchor, $T=0.5$ & $167$ & $+0.0325$ $[+0.0090, +0.0560]$ & $+0.1160$ $[+0.0800, +0.1515]$ \\
anchor, $T=0.7$, penalty $1.1$ & $75$ & $+0.0650$ $[+0.0405, +0.0895]$ & $+0.1250$ $[+0.0880, +0.1615]$ \\
second draw, $T=1.0$ & $62$ & $-0.0150$ $[-0.0365, +0.0065]$ & $-0.0005$ $[-0.0330, +0.0315]$ \\
\midrule
best of $64$ $-$ anchor, $T=0.7$ & & $+0.0685$ $[+0.0455, +0.0915]$ & $+0.0630$ $[+0.0290, +0.0960]$ \\
best of $64$ $-$ anchor, $T=0.5$ & & $+0.0690$ $[+0.0460, +0.0915]$ & $+0.0600$ $[+0.0255, +0.0955]$ \\
best of $64$ $-$ anchor, $0.7$, penalty $1.1$ & & $+0.0365$ $[+0.0135, +0.0595]$ & $+0.0510$ $[+0.0165, +0.0850]$ \\
\bottomrule
\end{tabular}
\end{table}

\paragraph{A vetted scorer.}\label{app:promptblind}
The vetting condition of Section~\ref{sec:compose} asks for a scorer that neither holds the work nor takes the user's
instructions, and ours meets neither. Registered before any score
(\texttt{results/\allowbreak onset\_\allowbreak prediction\_\allowbreak promptblind.md}), Comma-7B, trained on openly
licensed text and the largest anchor to pass the vetting screen, scored each of the $64$ committed draws by its mean
per-token log-likelihood with no prompt in context, never serving an empty draw. It removes the query cap and most of
the gain with it: over one draw it gains $+0.0405$ $[+0.0165, +0.0645]$ under judge~B and $+0.0120$ $[-0.0230, +0.0475]$ under judge~G,
where the Qwen scorer gains $+0.1015$ and $+0.1760$, differences of $+0.0610$ $[+0.0375, +0.0840]$ and $+0.1640$ $[+0.1300, +0.1980]$.
We had predicted a vetted scorer would keep part of the gain under judge~G; it did not, so a vetted scorer that keeps
it is open.

\paragraph{How far the judges can be trusted.}\label{app:judgevalidity}
Six judges read the headline's texts, and the share of items on which a judge gives the same verdict in both
presentation orders runs $0.288$ to $0.370$ over the four arms for B, $0.630$ to $0.712$ for C, $0.756$ to
$0.796$ for D (\texttt{Qwen2.5-72B}), $0.354$ to $0.420$ for E (\texttt{Mixtral-8x7B}), $0.676$ to $0.774$ for F
and $0.784$ to $0.846$ for G (\texttt{gemma-2-27b-it}) (\texttt{results/\allowbreak judge\_\allowbreak order\_\allowbreak consistency.csv}). Three checks of the pre-specified judge~B, all in one pass that
reproduces the committed levels on $500$ of $500$ prompts (\texttt{results/\allowbreak windowed\_\allowbreak meter.csv}). It detects a known difference: a second draw of the risky model gains $+0.065$
$[+0.0395, +0.090]$ over the anchor ($+0.1485$ under~G). The $k=0.5$ null is informative: it had $80\%$ power
against $+0.0305$, and its $90\%$ interval, $[-0.0205, +0.015]$, sits inside half of selection's gain, so a
two one-sided test calls it equivalent to zero at that margin (post hoc). And on the $62$ prompts where both
compared arms get order-consistent verdicts the headline difference is $+0.121$ $[+0.024, +0.222]$ ($+0.095$
on the $21$ where all four arms do, too few to resolve). Neither judge scores an empty answer as a loss. B
scores selection's $41$ empty picks at $0.494$ against the risky model's text, and G at $0.634$, above the
arm's own $0.545$: on those prompts the templateless $8$B's continuation is often off task, and through the
chat template G scores the same empties at $0.2195$. Counted instead as losses (post hoc), the
differences of Table~\ref{tab:matched} hold under G at every budget but $W=125$, and under B against the
per-token meters ($+0.094$) and the windowed meter at $\log 64$ ($+0.048$) but not at $24.95$ or $40$ nats
($+0.021$ each); and under G the non-empty rule's headline difference is $+0.0335$ $[-0.020, +0.0875]$, so
under the most reliable judge the headline rests partly on the empties. No human rated anything.

\paragraph{Scorer family by judge family.}\label{app:factorial}
The largest judged reading comes from a judge in the Qwen scorer's family, so we scored the headline pool a
second time with \texttt{gemma-2-27b-it} and judged both selections under all six judges on one host,
registered in advance (\texttt{results/\allowbreak onset\_\allowbreak prediction\_\allowbreak scorer\_\allowbreak judge\_\allowbreak factorial.md}; Table~\ref{tab:factorial}). We predicted that the Qwen reward's
drafts would be preferred under at least five of the six judges. Only judge~B prefers them. Under C, D, E
and F the two scorers are unresolved, and under G, the gemma scorer's own family, the gemma-scored drafts
win. So there is a same-family preference in the gemma family and none measurable in the Qwen family. The
Qwen-family judges D and F read the gemma-scored headline at $+0.0765$ and $+0.1325$, as high as the
Qwen-scored one. With the gemma scorer the headline resolves under four of six judges, against five with
the Qwen scorer, and B, where it depends on the scorer, is one of the two that do not resolve it. At
$\log 64$ nats the gemma-scored selection still beats the budget-up-front meter, a whole-output
certificate of the same type, by $+0.0455$ $[+0.014, +0.0775]$ under~B. It also beats the windowed meter at
$\log 64$ nats a span, a certificate of a different type (conditional on all served text before the
span), by $+0.054$ $[+0.0215, +0.086]$ under~B. Both margins are confirmed under~G. Every same-host pass gives the committed panel's difference exactly.
Judge~B's, committed on a local A100, reads $+0.059$ here against $+0.0505$ there.

\begin{table}[ht]
\centering\footnotesize
\caption{The headline scored twice and judged six times, one host, $500$ prompts, order-averaged, $95\%$
intervals. $S$: the Qwen-scored selection's level minus the gemma-scored one's. $D_3$: selection's gain
minus the $k=10$ meter's, each over its own control (\texttt{results/\allowbreak scorer\_\allowbreak judge\_\allowbreak factorial.csv}).}\label{tab:factorial}
\setlength{\tabcolsep}{3pt}
\begin{tabular}{@{}llll@{}}
\toprule
judge & $S$ & $D_3$, Qwen-scored & $D_3$, gemma-scored \\
\midrule
B & $+0.0440$ $[+0.0230, +0.0650]$ & $+0.0590$ $[+0.0240, +0.0935]$ & $+0.0150$ $[-0.0185, +0.0490]$ \\
C & $+0.0175$ $[-0.0095, +0.0455]$ & $+0.0765$ $[+0.0285, +0.1265]$ & $+0.0590$ $[+0.0100, +0.1090]$ \\
D & $-0.0215$ $[-0.0455, +0.0020]$ & $+0.0550$ $[+0.0075, +0.1035]$ & $+0.0765$ $[+0.0285, +0.1260]$ \\
E & $+0.0100$ $[-0.0185, +0.0375]$ & $-0.0015$ $[-0.0450, +0.0425]$ & $-0.0115$ $[-0.0560, +0.0330]$ \\
F & $-0.0115$ $[-0.0345, +0.0115]$ & $+0.1210$ $[+0.0770, +0.1655]$ & $+0.1325$ $[+0.0890, +0.1765]$ \\
G & $-0.0400$ $[-0.0670, -0.0135]$ & $+0.0545$ $[+0.0030, +0.1080]$ & $+0.0945$ $[+0.0390, +0.1480]$ \\
\bottomrule
\end{tabular}
\end{table}

\paragraph{A recording defect, and its repair.}\label{app:recording}
The harness recorded each generation by slicing a left-padded batch row at that row's own prompt length,
so a row with $p$ pad tokens carried the last $p$ tokens of its prompt into the recorded text: the
opponent's and the anchor's texts on $73.6\%$ of headline prompts (mean $10.7$ characters), the $k=10$
meter's on $32.6\%$. Nothing served was affected, but the judge and the reward read the record. The slicing
is fixed and every recorded generation is recovered exactly as the decoded row minus its prompt. Re-judged
on the recovered text with nothing else changed, the headline difference moves from $+0.0645$
$[+0.0300, +0.0995]$ to $+0.0505$ $[+0.0155, +0.0860]$, and all twenty older order-averaged differences keep their sign class; of the forty single-arm gains beneath them three change class, selection's on MT-Bench ($+0.0375$ $[-0.0125, +0.0875]$) and CoTaEval-QA ($+0.0070$ $[-0.0165, +0.0300]$) now covering zero and the meter's on the public books falling below it ($-0.0275$ $[-0.0470, -0.0080]$) (\texttt{results/\allowbreak deecho\_\allowbreak rejudge.csv}).

\paragraph{Pre-registration.}\label{app:prereg}
Every band this paper scores against was committed before the run it scores, with its grid, admission
criterion, stop rule and excluded outcomes; each registration and its scoring log are in the code release,
available at \codeurl{}
(\texttt{results/\allowbreak onset\_\allowbreak prediction\_\allowbreak *.md}). Predictions that failed are reported as
results: the first scorer, the judge-C prediction, the $70$B pair's expected wins for the meter, the
CoTaEval turnover, the adversarial oracle on CoTaEval's infringement half, and the monotone vetting ladder.
Figure~\ref{fig:h2h}'s NewsQA row is registered, its $k=1.4$ fixed by a registered calibration rule before
any judge call. Read as one family of $21$ under Holm's correction, each $p$-value taken from its row's
interval, eight of Figure~\ref{fig:h2h}'s rows still exclude zero, seven in selection's favour and
AlpacaEval against it; the headline row alone ($p \approx 0.005$) does not survive a $21$-way correction
(\texttt{results/\allowbreak forest\_\allowbreak multiplicity.csv}, post hoc).

\section{The head-to-head in full}\label{app:sel-h2h}

\paragraph{By serving configuration.}\label{app:served}
\begin{table}[t]
\centering\footnotesize
\caption{\textbf{Against the risky model: selection and the metered decoder, by how the risky model is served.}
Every budget behind Table~\ref{tab:reversals}. Order-averaged judge-B levels against one opponent per block, the risky model's own draw ($0.5$ is
parity); $500$ prompts. Temperature $1.0$ without repetition penalty, except in the block at the authors'
$0.7$ and $1.1$. That block's $K/S_w$ is against the anchor's $S_w = 177.4$ nats at $0.7$ with penalty
$1.1$ ($179.1$ without it). \emph{Active}: share of the judged trajectories' decode steps whose served
token is a strict blend of the two models. $K/S_w$: the meter's certificate against a $50$-token window
($S_w = 159.8$ nats; $139.7$ under Comma-7B); at or above $1$ it excludes nothing. Last column: selection
at $n=64$ minus the meter, a difference of served \emph{levels} paired over prompts, $95\%$ interval.
Selection's certificate is $\log 64 = 4.16$ nats pathwise in every row (KL bound $3.175$, attained only for
a tie-free score and unrepeated strings, Appendix~\ref{app:realisedkl}). The difference of gains gives the
same reading on $19$ of the $21$ rows; the two exceptions, temperature $0.7$ at $k=0.5$ and the chat
template at $k=1$, are direct wins that tie on gains. Levels compare within a block only
(Appendix~\ref{app:served}).}
\label{tab:served}
\setlength{\tabcolsep}{4.5pt}
\begin{tabular}{@{}lrrrrr@{}}
\toprule
risky model, as served & $k$ & active & $K/S_w$ & meter & selection $-$ meter \\
\midrule
\multicolumn{6}{@{}l}{\emph{opponent: the $8$B-Instruct without its chat template; anchor alone $0.439$, selection $0.555$}} \\
$8$B-Instruct, continuing text & $0.5$ & $49.1\%$ & $0.63$ & $0.436$ & $+0.1195$ $[+0.0955, +0.1435]$ \\
 & $1$ & $6.6\%$ & $1.25$ & $0.456$ & $+0.0995$ $[+0.074, +0.1245]$ \\
 & $10$ & $0.008\%$ & $12.5$ & $0.490$ & $+0.0655$ $[+0.040, +0.092]$ \\
$70$B base (the authors' pair) & $0.5$ & $26.5\%$ & $0.63$ & $0.429$ & $+0.1265$ $[+0.103, +0.1495]$ \\
 & $1$ & $6.3\%$ & $1.25$ & $0.433$ & $+0.1225$ $[+0.098, +0.146]$ \\
 & $20$ & $0.00\%$ & $25.0$ & $0.451$ & $+0.1040$ $[+0.080, +0.1285]$ \\
\midrule
\multicolumn{6}{@{}l}{\emph{at the authors' $0.7$ and $1.1$, other host; opponent: the $8$B's own draw there; anchor $0.312$, selection $0.427$}} \\
$8$B-Instruct, continuing text & $0.5$ & $44.6\%$ & $0.56$ & $0.394$ & $+0.0325$ $[+0.008, +0.0565]$ \\
 & $1$ & $16.4\%$ & $1.13$ & $0.441$ & $-0.0145$ $[-0.039, +0.010]$ \\
 & $10$ & $0.040\%$ & $11.3$ & $0.500$ & $-0.0730$ $[-0.096, -0.0515]$ \\
$70$B base (the authors' pair) & $0.5$ & $46.4\%$ & $0.56$ & $0.330$ & $+0.0970$ $[+0.073, +0.1215]$ \\
 & $1$ & $20.2\%$ & $1.13$ & $0.319$ & $+0.1080$ $[+0.083, +0.133]$ \\
 & $20$ & $0.00\%$ & $22.5$ & $0.370$ & $+0.0565$ $[+0.0305, +0.082]$ \\
\midrule
\multicolumn{6}{@{}l}{\emph{AnchoredByte, $k$ per byte, own pass: Comma-7B alone $0.4660$, selection at Comma-7B $0.6245$}} \\
$70$B base with Comma-7B & $0.1$ & $18.4\%$ & $0.57$ & $0.4860$ & $+0.1385$ $[+0.1155, +0.1620]$ \\
 & $0.5$ & $2.8\%$ & $2.86$ & $0.5005$ & $+0.1240$ $[+0.1010, +0.1470]$ \\
 & $2$ & $0.6\%$ & $11.45$ & $0.4900$ & $+0.1345$ $[+0.1105, +0.1590]$ \\
\midrule
\multicolumn{6}{@{}l}{\emph{opponent: $8$B-Instruct via its chat template; anchor $0.238$, selection $0.363$ ($^\dagger$other host: $0.2385$, $0.367$)}} \\
$8$B-Instruct, chat template & $0.5^\dagger$ & $62.7\%$ & $0.63$ & $0.248$ & $+0.1195$ $[+0.0925, +0.1465]$ \\
 & $1$ & $16.8\%$ & $1.25$ & $0.322$ & $+0.0410$ $[+0.012, +0.0705]$ \\
 & $2^\dagger$ & $2.5\%$ & $2.50$ & $0.384$ & $-0.0170$ $[-0.046, +0.0125]$ \\
 & $3^\dagger$ & $1.1\%$ & $3.75$ & $0.405$ & $-0.0380$ $[-0.068, -0.008]$ \\
 & $5^\dagger$ & $0.23\%$ & $6.26$ & $0.427$ & $-0.0600$ $[-0.091, -0.029]$ \\
 & $10$ & $0.16\%$ & $12.5$ & $0.501$ & $-0.1380$ $[-0.167, -0.107]$ \\
\bottomrule
\end{tabular}
\end{table}

Table~\ref{tab:served}, summarised in Table~\ref{tab:reversals} (Section~\ref{sec:served}), comes from two registered arms, banded before any judge call, and one descriptive
re-judge that puts every budget in one pass; all three judge the same selection texts, whose per-prompt
levels agree across them on $500$ of $500$ prompts. Served through its chat template the $8$B was generated
afresh at $k=10$; its own second draw reads $0.504$ against the first, and it beats its templateless text by
$+0.184$ $[+0.155, +0.213]$. Against that stronger opponent selection still beats the committed
(templateless) meter, $+0.058$ $[+0.0325, +0.0845]$. At the $70$B pair we predicted the meter would win at
$k=20$ and at $k=1$, and both predictions were wrong: at temperature $1.0$ the base model continuing
\texttt{Complete the prefix:} is judged no better than the anchor, and at $k=20$ the meter serves its text
byte for byte on $500$ of $500$ prompts. Two post-hoc checks: the meter writes about twice selection's words
($151$ against $75$ at the median), and a length-controlled regression \citep{dubois2024length} leaves the
difference at $+0.0605$ $[+0.0236, +0.0971]$; on the $414$ prompts whose two anchor controls are non-empty it
is $+0.0441$ $[+0.0024, +0.0864]$ (\texttt{results/\allowbreak h2h\_\allowbreak length\_\allowbreak control\_\allowbreak note.md}).
Read with a Bonferroni correction over its rows, every row of Table~\ref{tab:served}'s released and chat
blocks still excludes zero in the same direction, the narrowest being the chat-template $k=1$ row at
$[+0.0015, +0.0820]$ (\texttt{results/\allowbreak served\_\allowbreak multiplicity.csv}, post hoc).
Served through the same chat template, the audited memoriser alone reaches near-verbatim recall $0.2193$
($0.4921$ on plain text); the meter stays below $0.01$ at every budget through $k=3.8$ and reaches it at
$k=3.941$ $[2.968, 4.435]$, $1.22$ of $s(x)$, with $25.3\%$ of resamples never reaching it by $k=4.5$, so the upper
end is a floor (\texttt{results/\allowbreak chat\_\allowbreak onset.csv}, registered). The chat-served
meter's first win, at $k=3$, lies within that interval, but at $k=3$ itself measured recall is below $0.01$: the meter wins there before any measured extraction, the regime the Limitations weigh against a certificate.

\paragraph{The $70$B base as an opponent.} Is the base model at temperature $1.0$, judged no better than the
anchor, simply degenerate? By the usual markers it is not: over the $500$ prompts its median response is
$144$ words against the $8$B-Instruct's $154$ and the anchor's $63$, it is empty on $1.0\%$ (the anchor
$12.4\%$), and a single $5$-gram fills a fifth of the text on $2.2\%$ of responses against the anchor's
$17.8\%$. What it does is continue the prompt as a web page: asked for a story it writes the next post of a
forum thread, signed with a username and followed by replies. It is fluent and off task, so
the comparison is with a capable model that does not follow instructions, which is how its authors serve it
(\texttt{results/\allowbreak opponent\_\allowbreak degeneracy.csv}).

\paragraph{A difference of gains, or of levels.} Table~\ref{tab:served}'s last column is the direct
difference of served levels; the text quotes the difference of gains, each arm's level minus its own
zero-budget control's, which was the registered construction because the two pipelines' controls need not
agree. They do not quite: selection's $n=1$ draw is judged $0.008$ to $0.0215$ above the meter's anchor-alone
arm in every configuration but AnchoredByte, where the two are one arm. So the difference of gains is the
less favourable to selection of the two, never the more, and over all $21$ rows of the table the two give the
same reading on $19$; the two that differ are temperature $0.7$ at $k=0.5$ ($+0.0325$ against $+0.0245$
$[-0.0090, +0.0570]$) and the chat template at $k=1$ ($+0.0410$ against $+0.0305$ $[-0.0030, +0.0655]$),
each a direct win that is a tie on gains (\texttt{results/\allowbreak d3\_\allowbreak vs\_\allowbreak direct.csv}).

\paragraph{Under the authors' own utility metrics.}\label{app:hemetrics}
\citet{he2026anchored} score utility with Prometheus-2 fluency and FActScore precision, so we scored the same
recovered texts with both: their fluency rubric verbatim under \texttt{prometheus-7b-v2.0}
\citep{kim2024prometheus2} where they use \texttt{gpt-4.1-mini}, and FActScore \citep{min2023factscore} on
the $150$ biographies with \texttt{Qwen2.5-14B-Instruct} extracting and verifying against Wikipedia. On
fluency (1--5, non-empty responses) selection at $n=64$ scores $2.56$ against the meter's $1.88$ at $k=10$, a
paired $+0.70$ $[+0.53, +0.85]$, beats the meter at every budget of both text-continuation pairs
($+1.15$ $[+1.00, +1.30]$ over the $70$B pair at $k=0.5$), and loses to the chat-template model,
$-0.81$ $[-0.99, -0.64]$: the judge's pattern. FActScore reads the other way: selection's precision is its
anchor's ($0.042$ against $0.048$, paired $-0.008$ $[-0.041, +0.022]$) and below every metered arm
($-0.182$ $[-0.222, -0.145]$ against $k=10$), while it abstains on $20$ of $150$ biographies where a single
draw abstains on $80$. Best-of-$n$ is bounded by its best draw, not its average: FActScored draw by draw,
the most precise of the $64$ reaches $0.5054$ $[0.4553, 0.5564]$ ($0.2392$ among draws claiming at least five
facts) where one draw averages $0.0613$ and the reward's pick $0.0431$, so the draws hold far more precise
biographies than the reward serves. Asked instead whether every claim is accurate, the same scorer serves $0.0768$ by claiming
fewer facts and abstaining on $45$, paired $+0.0102$ $[-0.0144, +0.0399]$ over the committed pick and
$-0.1246$ $[-0.1727, -0.0725]$ below the meter at $k=10$: a scorer that asks for facts did not find them
(\texttt{results/\allowbreak factscore\_\allowbreak oracle.csv}, registered). The head-to-head is a statement
about fluency and helpfulness, not about factual recall, which the anchor bounds
(\texttt{results/\allowbreak he\_\allowbreak metrics.csv}).

\paragraph{The authors' byte-level decoder.}\label{app:anchoredbyte}
\citet{he2026anchored} meter every licensed anchor but TinyComma with AnchoredByte, which fuses a Comma
tokenizer with a Llama-3 one through ByteSampler \citep{hayase2026bytesampler}, and recommend Comma-7B with
the $70$B base at $k=0.5$ per byte. We ran their package unmodified (\texttt{anchoreddecode} at commit
\texttt{a12ecd9}) in Table~\ref{tab:served}'s configuration, with $B_{\max} = 800$ bytes so that $K = 800k$
nats and $S_w = 139.7$ under Comma-7B; selection is Comma-7B's best-of-$64$. Every trajectory stays within
$K$. Selection wins at every budget, as predicted, including $k=0.1$, where the certificate says something
about a window ($K = 80$) and the meter binds on $18.4\%$ of byte steps and is judged indistinguishable from
Comma-7B alone. Under the authors' metrics selection is the more fluent at every budget; at $k=0.5$ the
meter's FActScore precision is $0.323$ against selection's $0.0995$, the $70$B's facts passing through a
certificate that is vacuous for any window ($K/S_w = 2.86$), and at $k=0.1$ it is $0.0843$, inseparable from
selection's (\texttt{results/\allowbreak onset\_\allowbreak prediction\_\allowbreak anchoredbyte.md}). At their
temperature $0.7$ and penalty $1.1$ (registered; their \texttt{generate} refuses $k=0$, so the anchor control is
their byte path at $k=10^{-6}$, forced to the anchor at every byte), the meter at $k=0.1$ binds on $22.9\%$ of
byte steps and again gains nothing over its anchor, $+0.008$ $[-0.008, +0.0245]$, while TinyComma's selection
at those settings gains $+0.099$ $[+0.068, +0.129]$ more over its own; the tempered Comma-7B has
$S_w = 155.3$ nats against $140.7$ untempered, both over the $758$ protected passages, so $K/S_w = 0.515$
(\texttt{results/\allowbreak anchoredbyte\_\allowbreak t07.csv}); the $140.7$ re-measures the construction behind the
temperature-$1.0$ arm's $139.7$, the audited anchor's median rescaled by per-character rates, $1$ nat higher.

\begin{figure}[ht]
\centering
\includegraphics[width=0.86\textwidth]{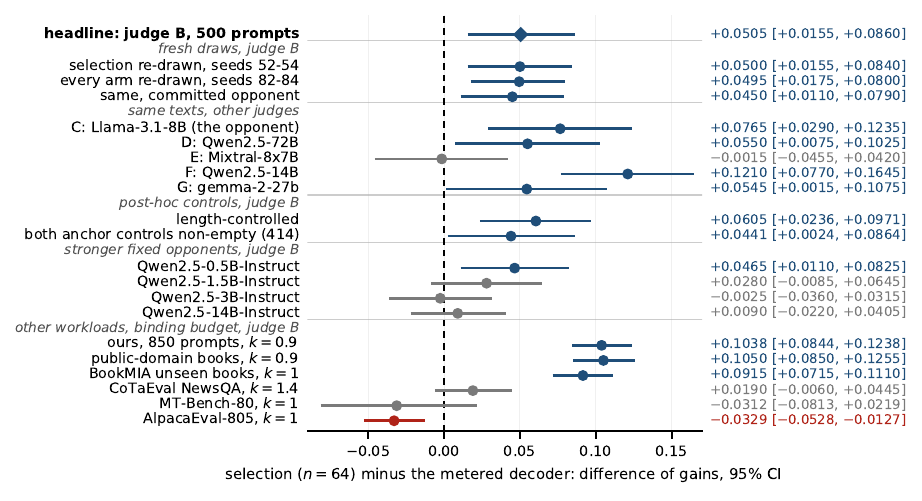}
\caption{\textbf{The head-to-head difference under every perturbation we ran.} Each row is selection's
order-averaged gain at $n=64$ over its own anchor-alone control minus the meter's at $k=10$ (last block:
the named $k$, rate-matched to bind), paired over prompts, $95\%$ interval, all on the repaired
text. Blue excludes zero above, red below. Judges C--G re-score the headline's texts; C is the
opponent's own checkpoint. The controls regress out the length gap (the meter writes $151$ words to
selection's $75$ at the median) and drop the $86$ prompts whose anchor controls are empty. Opponents:
fixed Qwen2.5-Instruct models, each stronger than the committed one (Appendix~\ref{app:h2hrepeat}).}
\label{fig:h2h}
\end{figure}

\paragraph{Other judges and fresh draws.}\label{app:h2hrepeat}
Table~\ref{tab:h2hrepeat} gives the recorded-text readings the repaired ones in Figure~\ref{fig:h2h}
replaced. Judges D to G re-score byte-identical text, so every difference between their rows is the
instrument. \texttt{Mixtral-8x7B}, the one judge whose interval contains zero, re-asked on the $850$-prompt
pass at a binding budget reads $+0.0226$ $[-0.0018, +0.0476]$: the same sign as the others and a small, not an absent, effect. Two earlier attempts ran on AlpacaEval, where the reversal does not hold, and were refused by their gates; the extra prompts cut a half-width of $0.0443$ to $0.0247$, and the dispute is the meter's ($+0.0971$ under Mixtral, $+0.0253$ under judge~B). Family and size do not account for it ($3.8$B, $14$B, $27$B and $72$B resolve;
$47$B does not). The judges disagree item by item (per-prompt correlation $+0.284$ between D and E), so we
report the fraction that resolve and never average them. Order consistency, the share of prompts whose
two presentation orders agree, spans $0.29$--$0.37$ over the four arms for judge~B, $0.35$--$0.42$ for E,
$0.63$--$0.71$ for C, $0.68$--$0.77$ for F, $0.76$--$0.80$ for D and $0.78$--$0.85$ for G; the two most
consistent judges both place the repaired difference above zero
(\texttt{results/\allowbreak judge\_\allowbreak order\_\allowbreak consistency.csv}). Two readings are reported against us: the
seed-$52$ judge-C estimate, $+0.1000$, falls just outside the original interval on the favourable side and
is never quoted as the headline; and we predicted judge~C, the opponent's own checkpoint, would be a harsh
test, and it did the opposite, so that prediction is withdrawn.

\begin{table}[ht]
\centering\footnotesize
\caption{The head-to-head repeated, on the recorded text unless marked. Paired difference $D_3$ of
order-averaged gains within each pass; judges B and C as in the main text, D--G re-score the seed-$42$
texts. \emph{seeds $82$}: every sampled arm re-drawn, judged against a re-drawn opponent (\emph{all new}) or
the committed one (\emph{old opp.}).}\label{tab:h2hrepeat}
\begin{tabular}{@{}llrrr@{}}
\toprule
draw & judge & selection $D_1$ & metered $D_2$ & \textbf{difference $D_3$} \\
\midrule
seed $42$ (original) & B & $+0.1045$ & $+0.0400$ & $\mathbf{+0.0645}$ $[+0.0300,+0.0995]$ \\
repaired text & B & $+0.1015$ & $+0.0510$ & $\mathbf{+0.0505}$ $[+0.0155,+0.0860]$ \\
seed $42$            & C & $+0.1360$ & $+0.0740$ & $+0.0620$ $[+0.0165,+0.1075]$ \\
seed $52$ (fresh)    & B & $+0.1035$ & $+0.0400$ & $+0.0635$ $[+0.0290,+0.0975]$ \\
seed $52$, repaired & B & $+0.1010$ & $+0.0510$ & $+0.0500$ $[+0.0155,+0.0840]$ \\
seeds $82$ (all new) & B & $+0.1225$ & $+0.0730$ & $+0.0495$ $[+0.0175,+0.0800]$ \\
seeds $82$ (old opp.) & B & $+0.1065$ & $+0.0615$ & $+0.0450$ $[+0.0110,+0.0790]$ \\
seed $52$ (fresh)    & C & $+0.1740$ & $+0.0740$ & $+0.1000$ $[+0.0520,+0.1465]$ \\
\midrule
seed $42$            & D, \texttt{Qwen2.5-72B} & $+0.1950$ & $+0.1330$ & $+0.0620$ $[+0.0145,+0.1100]$ \\
seed $42$            & E, \texttt{Mixtral-8x7B} & $+0.1060$ & $+0.0970$ & $+0.0090$ $[-0.0355,+0.0530]$ \\
seed $42$            & F, \texttt{Qwen2.5-14B} & $+0.1680$ & $+0.0610$ & $+0.1070$ $[+0.0630,+0.1510]$ \\
seed $42$            & G, \texttt{gemma-2-27b} & $+0.1975$ & $+0.1185$ & $+0.0790$ $[+0.0265,+0.1325]$ \\
\bottomrule
\end{tabular}
\end{table}

\paragraph{Stronger opponents.}
Every judged arm is scored against one fixed opponent, so we varied it
(Table~\ref{tab:opponentladder}): strength is the opponent's order-averaged win rate against the anchor's
own $n=1$ completion. Both mechanisms beat their own controls against every opponent; the difference between
them decays as the opponent strengthens and crosses zero between $0.773$ and $0.825$ (Spearman $-0.900$ over
the five). Judge~C agrees on the shape and crosses earlier, confirming only the committed opponent (column judge~C), its shift strength-dependent ($-0.017$, $-0.058$, $-0.021$, $-0.004$) rather than the uniform one we had predicted; read on the three new rungs alone the decay looked like one judge's, our first reading, and that was wrong; and
\texttt{Mixtral-8x7B} on the rung where they differ reads $+0.0355$ $[-0.0105, +0.0810]$, so three
instruments do not pin the boundary. Against \texttt{Qwen2.5-14B-Instruct} order consistency is
$0.550$ for selection and $0.482$ for the meter, the latter below chance agreement between
presentation orders, which is why only order-averaged figures are quoted. Strength is not parameter count: \texttt{Qwen2.5-0.5B}, sixteen times
smaller than the committed opponent, scores $0.704$, and the committed opponent writes the longest
completions ($154.1$ words against $109.5$--$133.2$). Every alternative opponent was generated one prompt
at a time by a different decoder from the committed one's, a pipeline change that alone moves the committed
opponent's strength on AlpacaEval from $0.6792$ to $0.8519$; only paired differences within one pass are
compared, and the same ladder on AlpacaEval rises with strength (Spearman $+0.900$); only divided by the room a difference of win rates has does it point the same way ($-0.800$), five points each, and neither is a significance claim. No
instruction-tuned opponent weaker than $0.555$ was available, so the registered test of that account went untested, and the decay is a shape in the data, not a mechanism we have shown. The judged head-to-head is therefore opponent-dependent, which is why the paper rests its claim
on the certificate and the judge-free comparisons as well.

\begin{table}[ht]
\centering\footnotesize
\caption{The judged head-to-head against five fixed opponents (all \texttt{-Instruct}; $^*$ the committed
one), $500$ prompts, recorded text (repaired readings in Figure~\ref{fig:h2h}). Only the opponent changes;
the reading is judge~B's band, fixed before each run; judge~C, post hoc, did not re-score the $14$B rung.}
\label{tab:opponentladder}
\setlength{\tabcolsep}{4pt}
\begin{tabular}{@{}lrlll@{}}
\toprule
opponent & strength & judge~B, paired $g_{\mathrm{sel}} - g_{\mathrm{met}}$ & judge~C & reading \\
\midrule
\texttt{Llama-3.1-8B}$^*$ & $0.555$ & $+0.0645$ $[+0.0300, +0.0995]$ & $+0.0475$ $[+0.0015, +0.0930]$ & confirmed \\
\texttt{Qwen2.5-0.5B} & $0.704$ & $+0.0490$ $[+0.0135, +0.0850]$ & $-0.0090$ $[-0.0555, +0.0375]$ & confirmed \\
\texttt{Qwen2.5-1.5B} & $0.773$ & $+0.0195$ $[-0.0165, +0.0555]$ & $-0.0015$ $[-0.0455, +0.0425]$ & unresolved \\
\texttt{Qwen2.5-3B} & $0.825$ & $-0.0200$ $[-0.0540, +0.0140]$ & $-0.0240$ $[-0.0555, +0.0085]$ & unresolved \\
\texttt{Qwen2.5-14B} & $0.850$ & $-0.0065$ $[-0.0385, +0.0255]$ & --- & unresolved \\
\bottomrule
\end{tabular}
\end{table}

\paragraph{Other workloads.}\label{app:workload}
At the paper's own $k=10$ the budget is barely active on any workload ($24$ of $299{,}843$ steps, $0.008\%$, for the committed arm; $0.016\%$ on AlpacaEval), and the meter serves the opponent's text byte for byte on $794$ of $805$
AlpacaEval prompts, so each workload was re-run at a budget chosen to bind by matching a rate: over $k = 0.1, 0.3, 1, 3, 10$ AlpacaEval's activity reads $0.765$, $0.684$, $0.077$, $0.005$ and $0.0002$, and $k=1.0$ binds on $8.008\%$ of steps, about a thousand times the committed arm's rate. Figure~\ref{fig:h2h} gives the six binding-budget differences on the recovered text.
The reversal holds on prefix completion (our $850$ prompts, $500$ excerpts of $42$ public-domain books, $500$
excerpts of BookMIA's unseen half), is unresolved on CoTaEval's NewsQA and on MT-Bench's $80$ prompts, and
fails on AlpacaEval, $-0.0329$ $[-0.0528, -0.0127]$ ($-0.0339$ $[-0.0534, -0.0137]$ as recorded, selection gaining $+0.0835$ $[+0.0665, +0.1016]$ and the meter $+0.1174$ $[+0.0991, +0.1354]$), which a disjoint re-draw reproduces ($-0.0255$ $[-0.0453, -0.0062]$). Judge~C reads $+0.0903$ $[+0.0629, +0.1176]$ on ours and $-0.0969$ $[-0.1264, -0.0677]$ on AlpacaEval, \texttt{Mixtral-8x7B} $+0.0226$ $[-0.0018, +0.0476]$ and $-0.0854$ $[-0.1112, -0.0593]$: no judge crosses, so the split is not one judge's. Our own $850$ prompts through the same pipeline hold at both budgets, $+0.0482$ $[+0.0303, +0.0662]$ at $k=10$ and $+0.0965$ $[+0.0765, +0.1162]$ at $k=0.9$, so across five passes on these two workloads the split is the workload, not the budget, the batch size, the pipeline or the seed. The two respond to a binding budget in \emph{opposite} directions, as a support ceiling would predict and the recorded-text decomposition below refutes. Neither anchor competence (the sign does not move across our own
three classes, which span $0.137$ of the anchor's win rate, twice the $0.065$ gap between the two workloads' ranges), nor, within AlpacaEval, the opponent (the meter wins against five opponents of strength $0.7752$ to $0.8807$, $-0.0584$ to $-0.0814$ with every interval clear of zero, though even the weakest beats the anchor three times in four) explains the split, and neither does a support ceiling: decomposed, the swing from our workload to
AlpacaEval is mostly the meter's, whose gain rises by $0.092$ while selection's falls by $0.038$. On the repaired text that decomposition does not survive: over the three binding passes selection's gains span $0.095$ ($+0.1321$, $+0.0801$, $+0.0375$) and the meter's $0.085$ ($+0.0282$, $+0.1130$, $+0.0688$), so both arms move with the workload and the support-ceiling account is neither refuted nor supported; we cannot identify the cause. Across the six workloads the opponent's strength orders the differences (Spearman $-0.943$) but not once divided by the headroom ($-0.657$), so we withdraw the claim that it orders them; the meter's gain rises in exact rank order with it ($+1.000$) and selection's does not ($+0.486$). Nor is it the prompt template: our factual class, $500$ of the $850$ prompts, carries no header and reads $+0.0990$ $[+0.0720, +0.1260]$, AlpacaEval, equally header-free, $-0.0339$ $[-0.0540, -0.0137]$, and our two header-carrying classes agree ($+0.0838$ and $+0.1050$). MT-Bench reads $-0.0250$ $[-0.0781, +0.0281]$ under judge~B, $-0.0094$ $[-0.0875, +0.0656]$ and $-0.0250$ $[-0.0906, +0.0406]$ under C and \texttt{Mixtral-8x7B}, and resolves only at $k=10$, $-0.0875$ $[-0.1375, -0.0375]$; on the recorded text the decomposition holds a third time, selection's gains spanning $0.056$ and the meter's $0.092$, and selection gains on all three. The
public-domain books are plausibly inside the anchor's training distribution as well as completion-shaped,
and we registered that a win there would be confounded and reported so (the meter does no better than its own control there, $-0.0185$ $[-0.0380, +0.0005]$, and on the unseen books does \emph{worse} than its own control at both budgets, $-0.0400$ $[-0.0585, -0.0215]$; as recorded, $+0.0990$ $[+0.0795, +0.1185]$ at $k=0.9$ and $+0.0950$ $[+0.0780, +0.1120]$ at $k=10$); BookMIA's unseen half, on whose
$500$ prefixes the anchor reproduces nothing, removes that confound as far as a contested membership label can, with ours at both budgets, $+0.1040$ $[+0.0850, +0.1230]$ and $+0.0885$ $[+0.0735, +0.1035]$, selection gaining $+0.0640$ $[+0.0485, +0.0795]$; that check is one-sided, the same instrument reading $0.000$ on the protected corpus, so it excludes leakage without proving unfamiliarity. CoTaEval's news questions read $+0.0070$ $[-0.0175, +0.0320]$ at $k=1.4$ and $-0.0375$ $[-0.0600, -0.0145]$ at $k=10$, selection gaining $+0.0235$ $[+0.0010, +0.0470]$ at both and the meter $+0.0165$ $[-0.0030, +0.0355]$ and $+0.0610$ $[+0.0420, +0.0800]$: the task-type account is not falsified and not cleanly confirmed, and we registered that this arm cannot separate it from ours against the rest. We report the scoping, prefix completion against instruction following, and not a mechanism for it.
Off support it reads $-0.0357$ $[-0.0553, -0.0165]$ at $n=128$ and $-0.0239$ $[-0.0429, -0.0050]$ at $256$,
narrower by less than the half we predicted, so the reversal fails off-support at $256$ as at $64$
(a certificate of $\log 256 = 5.55$ nats against the meter's realised $145.1$); at $512$ it is $-0.0053$
$[-0.0248, +0.0143]$, which the band fixed in advance reads as \textsc{unresolved}, against our prediction
that it would still trail, at $\log 512 = 6.24$ nats, $23.3\times$ below the meter's realised spend. At $k=10$ the budget is active on $0.231\%$ of CoTaEval's steps against $0.008\%$--$0.043\%$ on the six others (on the unseen books $0.008\%$ of steps, with $99.6\%$ of served completions the opponent's), and the meter is the opponent byte for byte on $76.6\%$ of prompts there against $95.0\%$--$99.6\%$ on the five where that comparison is interpretable.

\paragraph{The meter given selection's reward.}
Best-of-$n$ over a decoder with budget $K$ is certified at $K + \log n$, so the meter can be handed the same
scorer. On TriviaQA, at the one meterable pair, the metered decoder drew $16$ trajectories per prompt and
every draw was scored by the same reward in the same pass (Table~\ref{tab:parity}). The scorer that lifts the
anchor's draws by $+0.098$ moves the meter by $-0.022$ and $+0.012$, neither clear of zero; the meter does win on level, $0.594$ against $0.210$, but only at $k=20$, whose $480.7$-nat certificate is vacuous on every one of these questions (median $S(x) = 5.86$ nats). Diversity explains only half of this: at
$k=20$ the meter's $16$ draws collapse ($4.87$ distinct answers), but at $k=3$ they are within $4\%$ of the
anchor's ($12.83$ against $13.26$) and the reward still buys nothing, so there the failure is the
scorer's.

\begin{table}[ht]
\centering\footnotesize
\caption{The same reward model, call and pass. Paired gain at $n=16$ over each arm's own $n=1$; $k=0.5$ and
$k=1$ failed their registered instrument gate and are excluded.}
\label{tab:parity}
\setlength{\tabcolsep}{5pt}
\begin{tabular}{@{}lrrrl@{}}
\toprule
arm & acc $n{=}1$ & acc $n{=}16$ & paired gain & verdict \\
\midrule
metered $k{=}3$   & $0.158$ & $0.136$ & $-0.022$ $[-0.058, +0.014]$ & no effect \\
metered $k{=}20$  & $0.576$ & $0.588$ & $+0.012$ $[-0.020, +0.046]$ & no effect \\
selection over the anchor & $0.112$ & $0.210$ & $+0.098$ $[+0.068, +0.128]$ & climbs \\
\bottomrule
\end{tabular}
\end{table}

\paragraph{One certificate, split between the draw and the token.}\label{app:hybrid}
Proposition~\ref{prop:selection} holds for a pool drawn from any law with $q_B \le e^{B}p_s$, so best-of-$n$
over pathwise-metered drafts at budget $B$ is certified at $\log n + B$. At a fixed $C = \log 64$ the level
rises monotonically with the share spent on the draw, $0.4400$, $0.4610$, $0.4995$ and $0.5550$, and every
split loses to pure selection: $-0.1150$ $[-0.1380, -0.0915]$ for the meter alone, $-0.0940$
$[-0.1170, -0.0715]$ and $-0.0555$ $[-0.0780, -0.0330]$ for the two hybrids; the meter given the whole
certificate is judged no better than a single anchor draw ($0.4400$ against $0.4535$), and at fixed $n=8$ the tilt is worth $-0.0120$ $[-0.0300, +0.0060]$.

\paragraph{Stopping early.}\label{app:adaptive}
A rule that stops at the first reward above a threshold, capped at $n_{\max}=64$, is certified at $\log 64$
however early it stops. With the threshold at the first draw's $0.75$ quantile it stops after $6.8$ draws on
average, loses to fixed $n=64$ by $-0.0515$ $[-0.0720, -0.0310]$, and reads $+0.0040$ $[-0.0125, +0.0205]$ against fixed $n=7$, the same expected draws; at $q=0.90$ it uses $17.4$ draws, $27\%$ of the budget, and straddles zero against both $n=64$ ($-0.0155$ $[-0.0340, +0.0020]$) and $n=17$: a deployer who wants fewer draws should
lower $n$.

\section{Selection alone: \texorpdfstring{$n$}{n}, anchors, scorers and tasks}\label{app:sel-breadth}\label{app:selection}

\begin{figure}[ht]
\centering
\includegraphics[width=0.85\textwidth]{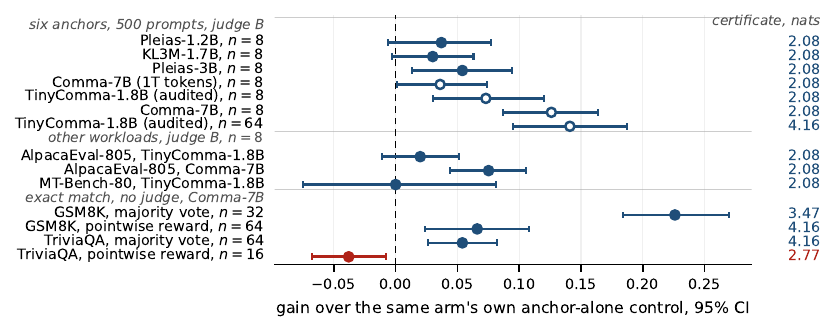}
\caption{\textbf{Selection against its own anchor.} Gain over \emph{that arm's own} anchor-alone draw,
single-order, $95\%$ interval, judged on the repaired text; right, the whole certificate in nats. Open
markers miss the admission criterion of at most $5\%$ empty completions on the true generations
(TinyComma $14.6\%$ and $12.8\%$, Comma-7B $18.4\%$, Comma-1T $31.6\%$); on non-empty prompts their gains
move by at most $0.012$. MT-Bench's $80$ items resolve nothing at a half-width of $0.08$, and the
pointwise reward falls with $n$ on TriviaQA (Appendix~\ref{app:judgefree}). Only the audited anchor is
plotted at $n=64$; the climbs to $n=64$ at three other anchors are Table~\ref{tab:climb}'s, where two carry
a reduced warrant and the one that climbed did not survive a fresh draw.}
\label{fig:breadth}
\end{figure}

\paragraph{How far $n$ pays.}
At Comma-7B the gain rises monotonically to $n=64$ within one pass ($+0.041$, $+0.060$, $+0.072$,
$+0.103$, $+0.127$, $+0.173$ at $n = 2,4,8,16,32,64$ under judge~B, $+0.062$ to $+0.230$ under judge~C,
Spearman $+1.000$ in $\log n$ under both), under a $\log 64 = 4.16$ pathwise certificate and the sharper
KL bound of $3.17$ nats, with no sign of the reward overoptimisation that turns such curves over
anywhere on \emph{this} grid; it does turn curves over on TriviaQA and CoTaEval
(Table~\ref{tab:scorerladder}). Past $n=64$ (Table~\ref{tab:ladder}) every pool is bit-identical to its
committed lower ranks in all rewards. The audited anchor is flat from $64$ to $128$; taken to $256$ it
rises from $128$ on both workloads, against a prediction that it would have stopped, each rise about one half-width from zero, and from $64$
the two doublings contain zero, order-averaged too, $+0.0099$ $[-0.0050, +0.0248]$ (both checks post hoc); taken to $512$
the last doubling is flat on both, where we had predicted a marginal climb. Measured from $64$, three
doublings cost eight times the draws and $\log 8 = 2.08$ more nats and buy about $+0.03$, an interval
that still contains zero on both workloads. Under order averaging the two AlpacaEval differences from
$64$ exclude zero, $+0.0186$ $[+0.0065, +0.0301]$ and $+0.0286$ $[+0.0115, +0.0438]$, but at $1.58$ and
$1.77$ half-widths, the margin at which a paired difference has failed to replicate here. We claim
neither a ceiling nor a slope past $n=64$.

\begin{table}[ht]
\centering\footnotesize
\caption{Selection past $n=64$, the audited anchor unless marked: paired differences of judged gains between rungs of one pass,
judge~B unless marked, $95\%$ intervals. The first four rows are on the $500$ judged prompts, the rest on
the $850$-prompt workload (\emph{ours}) and AlpacaEval's $805$.}
\label{tab:ladder}
\setlength{\tabcolsep}{5pt}
\begin{tabular}{@{}lll@{}}
\toprule
step & ours & AlpacaEval \\
\midrule
$g(128)-g(64)$, $500$ prompts & $+0.0140$ $[-0.0180, +0.0460]$ & --- \\
\quad order-averaged & $+0.0140$ $[-0.0015, +0.0295]$ & --- \\
\quad at Comma-7B & $+0.0040$ $[-0.0270, +0.0350]$ & --- \\
\quad at Comma-7B, judge~C & $+0.0110$ $[-0.0190, +0.0410]$ & --- \\
$g(128)-g(64)$ & $-0.0094$ $[-0.0353, +0.0147]$ & $-0.0019$ $[-0.0273, +0.0217]$ \\
$g(256)-g(128)$ & $+0.0306$ $[+0.0047, +0.0547]$ & $+0.0280$ $[+0.0025, +0.0553]$ \\
$g(256)-g(64)$ & $+0.0212$ $[-0.0088, +0.0512]$ & $+0.0261$ $[-0.0056, +0.0565]$ \\
$g(512)-g(256)$ & $+0.0071$ $[-0.0165, +0.0300]$ & $+0.0056$ $[-0.0193, +0.0311]$ \\
$g(512)-g(64)$ & $+0.0312$ $[-0.0018, +0.0635]$ & $+0.0311$ $[-0.0025, +0.0634]$ \\
\bottomrule
\end{tabular}
\end{table}

\paragraph{The climb to $n=64$ at other anchors.}
Only TinyComma and Comma-7B had an $n=64$ curve, so the three anchors that stop at $n=8$ and pass the
admission criterion were extended, bands fixed in advance (Table~\ref{tab:climb}; Comma-1T fails the gate
at $31.6\%$ empty). One of three climbs, and \textbf{it did not survive a fresh draw}: with the seeds
$42,43,44 \rightarrow 52,53,54$ and nothing else changed KL3M-1.7B reads $+0.0040$ $[-0.0330, +0.0400]$,
a move of $0.061$, so \textbf{no anchor outside the two already on record climbs reproducibly}. Both of
those were re-drawn the same way and held: the audited anchor at $+0.0880$ twice, Comma-7B at $+0.0880$
$[+0.0460, +0.1290]$ against $+0.1010$, a move of $0.013$. The gain rises in $\log n$ and flattens at an
anchor-dependent $n$, below $8$ for some anchors and above $64$ for others, so which $n$ is worth paying
for is a property of the anchor. A paired difference is stable where the effect is large against its own
interval: at $1.71$, $2.12$ and $2.43$ half-widths the three re-draws moved by $0.061$, $0.000$ and
$0.013$, so the ratio predicts whether a reading survives, not how far it moves. We do not pool a reading
with its own failed replication, and none of these numbers is set against Figure~\ref{fig:breadth}:
judged levels moved by up to $0.054$ between the two grids. One defect of ours: the protocol asserted \texttt{-{}-batch-size 32}
for all three new arms, true only of Pleias-3B, so the other two arms' reproduction check was inapplicable rather than failed and they carry a
reduced warrant.

\begin{table}[ht]
\centering\footnotesize
\caption{Does the climb to $n=64$ hold away from the two anchors it was measured on? Paired
$g(64)-g(8)$ within each pass, judge~B, $500$ prompts. The last two rows are the arms the claim rests
on, read by the same rule.}\label{tab:climb}
\setlength{\tabcolsep}{5pt}
\begin{tabular}{@{}lrrlc@{}}
\toprule
anchor & $g(8)$ & $g(64)$ & paired $g(64)-g(8)$ & reading \\
\midrule
Pleias-1.2B & $+0.083$ & $+0.119$ & $+0.0360$ $[-0.0040, +0.0740]$ & saturated by $8$ \\
KL3M-1.7B & $+0.023$ & $+0.088$ & $\mathbf{+0.0650}$ $[+0.0270, +0.1030]$ & \textbf{climbs} \\
Pleias-3B & $+0.025$ & $+0.028$ & $+0.0030$ $[-0.0380, +0.0460]$ & saturated by $8$ \\
\midrule
TinyComma-1.8B & $+0.054$ & $+0.142$ & $+0.0880$ $[+0.0470, +0.1300]$ & climbs \\
Comma-7B & $+0.072$ & $+0.173$ & $+0.1010$ $[+0.0590, +0.1420]$ & climbs \\
\bottomrule
\end{tabular}
\end{table}

\paragraph{Scorer size on the judged workload.}\label{app:frontier}\label{app:saturation}
Re-scoring the same $32{,}000$ cached candidates with four scorers on the identical template, and
judging the whole nested grid in one order-averaged pass (which reproduces the head-to-head at
$+0.1075$ against $+0.1045$ and the meter's $+0.0400$ exactly), gives $n=64$ gains of $+0.0230$,
$+0.0930$, $+0.0970$ and $+0.1065$ at $0.494$, $1.54$, $3.09$ and $7.62$B; shrinking the scorer from
$7.6$B to $0.494$B costs $-0.0855$ of the gain, and we predicted three of these four wrong, in the paper's
favour. \textbf{The $0.5$B scorer never
reaches the meter}: its best cell, $n=16$ at $3.68\times$ the meter's forward passes, gains
$+0.0395\,[+0.0205,+0.0585]$, and \textbf{at matched compute, $n=4$ at $0.92\times$, it loses outright,
$-0.0395\,[-0.0720,-0.0065]$}. Its curve peaks at $n=16$ and ends lower, as though $\log n$ is not a free knob; that
reading did not survive its registered test, the terminal drop $g(64)-g(16)$, $-0.0160\,[-0.0340,+0.0010]$,
\textsc{flat}; the Spearman of gain against $\log n$ is $0.5429$ for $0.5$B and exactly $1.0$ for each
larger scorer. Capability saturates early: only the first adjacent step at $n=64$ separates,
$+0.0700\,[+0.0500,+0.0900]$ from $0.5$B to $1.5$B, then $+0.0040\,[-0.0145,+0.0225]$ and
$+0.0095\,[-0.0090,+0.0285]$, so a $1.5$B scorer reaches $87.3\%$ of the $7.6$B gain for $35.2\%$ of its
forward passes. Two further rungs, in a separate six-scorer pass, do not lift it: against that pass's
own $7.6$B reading of $+0.1095$, a $14.8$B scorer reads $-0.0285\,[-0.0470,-0.0110]$ and a $72.7$B one
$-0.0085\,[-0.0255,+0.0085]$, the latter at $486.9\times$ the meter's forward passes. Both readings sit inside our $2.0$-half-width marginality rule, so the $14.8$B dip is not a result we build on. A crossing cell is
unstable where a gain is not: the $7.6$B scorer first matches the meter at $n=8$ ($7.66\times$) in one
pass and at $n=16$ ($15.32\times$) in the other, because one cell read $+0.0415$ and then $+0.0360$, so
we quote the $n=64$ column, which moved by $0.001$. Without a judge the reward caches order the scorers
the same way: within-prompt Spearman against the $7.6$B scorer $0.1333$, $0.4185$ and $0.4926$, and the
same draw served at $n=64$ at rates $0.052$, $0.132$ and $0.228$ against $0.016$ by chance. Every cell is
certified at $\log n$ whatever the scorer costs.

\paragraph{A bigger safe model instead.}\label{app:biggeranchor}
A larger safe model served once needs no scorer and no draws. The Comma-7B arm's rank-$0$ draw is
Comma-7B served alone; judged against TinyComma served alone with every item in both orders, it gains
$+0.0150$ $[-0.0065, +0.0365]$ at $3.9\times$ the parameters, while TinyComma at $n=64$ gains $+0.1045$
$[+0.0820, +0.1280]$ and Comma-7B at $n=64$ $+0.1755$ $[+0.1505, +0.2010]$. The paired difference of
the two served arms, which needs no control, is $-0.0995\,[-0.1220,-0.0765]$. On the forward-pass model
the bigger anchor is $86\times$ cheaper than selection at $n=64$; on this workload it does not measurably
work.

\paragraph{Majority vote, the scorer-free instance.}\label{app:scorerfree}
Self-consistency serves the modal answer over the same $n$ draws; its score is a regex and costs no
forward pass, and it carries the identical $\log n$ certificate because Proposition~\ref{prop:selection}
assumes nothing about the score. Table~\ref{tab:scorerfree} joins that cost model to the judge-free
accuracies. Every scorer-free cell beats every reward cell on both tasks: at $n=32$ majority vote gains
$3.42\times$ as much as the best reward cell for $9.4\%$ of its FLOPs, and on TriviaQA the reward goes
negative at every $n \ge 8$ while the vote rises to $0.334$. Three limits: the join is post hoc; a vote
needs a canonical answer, so it does not transfer to the judged workload; and no metered decoder ran on
GSM8K. The cost column prices the audited $1.8$B anchor while the accuracies are Comma-7B's, the direction that flatters the scorer-free rule; at Comma-7B
every vote cell costs $2.6\times$ more ($n=32$ is $14.90\times$, not $5.75\times$) and the reward cells
barely move ($61.29\times \to 62.23\times$), so at a matched $n=32$ the vote gains $3.77\times$ as much for $18.8\%$ of the reward rule's FLOPs at the audited anchor, and $47.9\%$ at Comma-7B. On the clock the scorer is $9.3\%$ of selection's
time (Table~\ref{tab:latency}), so dropping it saves far less time than parameters. The vote does not
reach the risky model, which scores $0.786$ greedy against its $0.546$.

\begin{table}[ht]
\centering\footnotesize
\caption{Cost and accuracy of the scorer-free instance, Comma-7B draws, sorted by forward-pass cost
relative to the metered decoder. Majority vote needs no reward model, so its only charge is the $n$
draws.}\label{tab:scorerfree}
\begin{tabular}{@{}llrrrr@{}}
\toprule
rule & $n$ & cost, FLOPs & GSM8K & TriviaQA \\
\midrule
majority vote & $8$ & $\mathbf{1.44\times}$ & $0.466$ & $0.322$ \\
reward $7.6$B & $2$ & $1.92\times$ & $0.352$ & $0.282$ \\
majority vote & $16$ & $2.87\times$ & $0.500$ & $0.322$ \\
reward $7.6$B & $4$ & $3.83\times$ & $0.344$ & $0.284$ \\
majority vote & $32$ & $\mathbf{5.75\times}$ & $\mathbf{0.546}$ & $0.328$ \\
reward $7.6$B & $8$ & $7.66\times$ & $0.364$ & $0.256$ \\
majority vote & $64$ & $11.50\times$ & $0.542$ & $0.334$ \\
reward $7.6$B & $64$ & $61.29\times$ & $0.386$ & $0.266$ \\
\bottomrule
\end{tabular}
\end{table}

\paragraph{Where the gain lies, against where the certificate speaks.}\label{app:gainwhere}
Proposition~\ref{prop:selection} excludes something about an event $E$ only when $S(E) > \log n$. Estimate
the anchor's probability of the correct answer on the $32$ draws that the vote at $n=32$ does not read. Then
\emph{all} of the vote's gain lies on questions where the anchor already gives the correct answer at least
$1/32$. GSM8K gains $+0.257$ $[+0.212, +0.303]$ on those $439$ and nothing on the $61$ where no held-out
draw is correct. TriviaQA gains $+0.084$ $[+0.038, +0.129]$ on $287$ and nothing on $213$
(\texttt{results/\allowbreak utility\_\allowbreak surprisal.csv}, post hoc). That holds by construction: a
rule that serves one of $n$ draws raises $q(E)$ at most to $n\,p_s(E)$, so it cannot make an answer likely
that the anchor gives less than about $1/n$. On these tasks the certificate and the utility sit on
different events. That is the design, since a protected $50$-token window has median surprisal $159.8$
nats. On open-ended text the event is the served completion itself, and there the gain lies where the
certificate does exclude something. The $460$ prompts whose served completion has $S > \log 64$ (median
$184.6$ nats) carry all of the headline's judged gain, $+0.110$ $[+0.085, +0.135]$ under~B. The other $40$
are empty answers, and they gain nothing.

\begin{figure}[ht]
\centering
\includegraphics[width=0.80\textwidth]{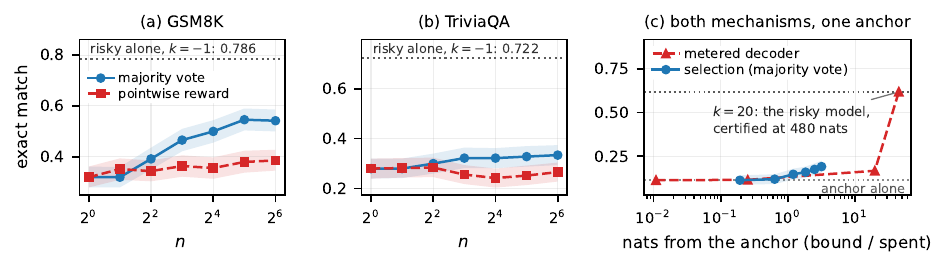}
\caption{\textbf{No judge, and where the anchor cannot do the task no $n$ rescues it.}
\textbf{(a, b)} Accuracy against $n$, $95\%$ bands; Comma-7B draws, \texttt{Qwen2.5-7B-Instruct} scores;
dotted, the risky model alone. GSM8K ($8$-shot, $500$): majority vote lifts $0.320 \to 0.546$ for $3.47$
certified nats, $3.4\times$ the pointwise reward $[2.20, 8.91]$. TriviaQA: it lifts $4.2\times$ less, and
this $7$B reward \emph{falls} with $n$ (Spearman $-0.607$), excluding zero on the wrong side at $n=16$; a
$14$B one does not (Table~\ref{tab:scorerladder}). \textbf{(c)} Both mechanisms at
\texttt{TinyComma-1.8B}, TriviaQA, $T_{\max}=24$; $x$ is selection's KL \emph{bound} against the meter's
\emph{measured} spend. \textbf{The meter wins, $0.618$ against $0.190$}: it is the anchor's own $0.112$ at
$k=0.5$ and $k=1$, and reaches $0.618$ only at $k=20$, where it \emph{is} the risky model, certified at
$480$ nats for a $24$-token answer.}
\label{fig:judgefree}
\end{figure}

\paragraph{The judge-free axis.}\label{app:judgefree}
Both arms run at Comma-7B at temperature $0.7$ and score the \emph{served} string, which the tie rule
returns as an index into the sample, never a synthesised answer. GSM8K: the first $500$ test problems,
$8$-shot, $256$-token cap, $64$ samples drawn once and read nested, scored by the final integer after
\texttt{\#\#\#\#}. TriviaQA \texttt{rc.nocontext}: the first $500$ validation questions, $5$-shot,
$24$-token cap, scored by containment of a gold alias as a whole-word span, a rule fixed before any number
because exact match scores format (a smoke run scored \texttt{The answer is David Seville} $0/8$ on
questions it had right). A number is extracted from $99.96\%$ and $98.67\%$ of the $500 \times 64$ samples
against a gate of $90\%$. The risky model alone reads $0.786$ greedy and $0.726$ sampled on GSM8K, $0.722$
and $0.674$ on TriviaQA. Majority vote lifts $4.2\times$ less on TriviaQA than on GSM8K (Figure~\ref{fig:judgefree},
Table~\ref{tab:judgefree}), the direction
registered in advance: a closed-book question offers less spread between draws than a derivation where one
slip decides the answer, and $q(y) \le n\,p_s(y)$ means no $n$ supplies a fact the anchor lacks. The two
lift ratios carry paired-bootstrap intervals: vote over reward on GSM8K $3.42$ $[2.20, 8.91]$ (Fieller
gives $[2.20, 8.81]$), vote on GSM8K over vote on TriviaQA $4.19$ $[2.63, 8.40]$; the third ratio has a
negative denominator and is undefined on $82\%$ of resamples, so we do not quote it
(\texttt{results/\allowbreak lift\_\allowbreak ratio\_\allowbreak interval.csv}).

\emph{A forced choice.} On MMLU ($5$-shot, $T_{\max}=24$, the first $500$ of the shuffled test set,
chance $0.25$) majority vote at Comma-7B gains $+0.084$ $[+0.042, +0.126]$ at $n=64$ over its own $0.404$,
$+0.108$ $[+0.066, +0.150]$ at $n=32$, and the pointwise reward $+0.056$ $[+0.008, +0.104]$. Both registered
gates passed: the anchor reads $0.404$ $[0.362, 0.448]$ and the risky model $0.676$ greedy. The same
protocol at TinyComma-1.8B read $0.188$ $[0.154, 0.222]$, below chance with $49\%$ of answers the letter
B, which the gate calls invalid, so the judge-free comparison against a metered decoder remains one task.

\emph{Against the meter, on TriviaQA.} At TinyComma, $T_{\max}=24$, the meter wins the level, and only at
$k=20$, where its accuracy is the risky model's $0.618$ $[0.576, 0.660]$ to three decimals, its
certificate $480$ nats and its realised mean spend $44.8473$. At $k=0$ and $k=0.5$ ($12$ nats) it reads
$0.112$, at $k=1$ ($24$ nats) $0.114$ and at $k=3$ ($72$ nats) $0.166$, against selection's $0.146$,
$0.174$ and $0.190$ at $n=8,32,64$ for $2.079$, $3.466$ and $4.159$. At every accuracy selection buys over
the shared control it pays $0.032$, $0.061$, $0.093$, $0.056$ and $0.071$ of what the cheapest metered
arm reaching that accuracy spends ($n=4$ to $64$); on the pathwise axis two ratios exceed a tenth
($0.106$ and $0.141$). Teacher-forcing the anchor on the accepted aliases gives a median $S(x)$ of
$5.86$ nats (p10 $1.35$, p90 $12.14$), so the meter's certificate is vacuous on $89.6\%$ of questions at
$k=0.5$, $99.4\%$ at $k=1$ and $100\%$ at $k=3$ and $k=20$, and selection's on $10.4\%$ at $n=4$ and
$35.4\%$ at $n=64$: selection's worst budget is vacuous less often than the meter's best, and still is under
the anchor tempered to the runs' $0.7$, at $36.2\%$ against $83.6\%$ (\texttt{results/\allowbreak vetting\_\allowbreak t07.csv}, \texttt{results/\allowbreak tqa\_\allowbreak vacuity\_\allowbreak t07\_\allowbreak summary.csv}). Both figures are lower bounds, since scoring the single most likely
alias rather than their sum overstates $S(x)$. At matched
compute the meter wins outright: it runs both models whatever $k$, so selection reaches parity at
$n = 5.57$, reads about $0.13$ there against $0.618$, and at $11.5\times$ the compute is still at $0.190$
(\texttt{results/\allowbreak compute\_\allowbreak matched\_\allowbreak judgefree.csv}). The meter buys
accuracy with compute cheaply and with certificate dearly, and selection the reverse.

\begin{table}[t]
\centering\footnotesize
\caption{The judge-free grids at Comma-7B. Gains are paired against each rule's own $n=1$ row, the
anchor's first draw and the same string for both rules; $\log n$ is the exact pathwise charge.
\textbf{Majority vote lifts on both tasks and the pointwise reward only on one}: on TriviaQA its accuracy
falls with $n$ (Spearman $-0.607$ against majority vote's $+0.982$; on GSM8K both climb, $+0.955$ and $+0.929$) and at $n=16$ the interval excludes
zero on the wrong side. \emph{At this scorer}: a $14$B one removes the fall and a $72$B one reverses it
(Spearman $+0.857$; Table~\ref{tab:scorerladder}).}
\label{tab:judgefree}
\begin{tabular}{rr rl rl}
\toprule
& & \multicolumn{2}{c}{majority vote} & \multicolumn{2}{c}{pointwise reward} \\
\cmidrule(lr){3-4}\cmidrule(lr){5-6}
$n$ & $\log n$ & acc. & gain & acc. & gain \\
\midrule
\multicolumn{6}{l}{\emph{GSM8K, exact match, $500$ problems}}\\
$1$ & $0.000$ & $0.320$ & $+0.000$ $[+0.000, +0.000]$ & $0.320$ & $+0.000$ $[+0.000, +0.000]$ \\
$2$ & $0.693$ & $0.320$ & $+0.000$ $[+0.000, +0.000]$ & $0.352$ & $+0.032$ $[+0.002, +0.062]$ \\
$4$ & $1.386$ & $0.392$ & $+0.072$ $[+0.044, +0.100]$ & $0.344$ & $+0.024$ $[-0.014, +0.062]$ \\
$8$ & $2.079$ & $0.466$ & $+0.146$ $[+0.110, +0.184]$ & $0.364$ & $+0.044$ $[+0.002, +0.088]$ \\
$16$ & $2.773$ & $0.500$ & $+0.180$ $[+0.142, +0.220]$ & $0.356$ & $+0.036$ $[-0.008, +0.080]$ \\
$32$ & $3.466$ & $0.546$ & $+0.226$ $[+0.184, +0.270]$ & $0.380$ & $+0.060$ $[+0.016, +0.104]$ \\
$64$ & $4.159$ & $0.542$ & $+0.222$ $[+0.182, +0.264]$ & $0.386$ & $+0.066$ $[+0.024, +0.108]$ \\
\midrule
\multicolumn{6}{l}{\emph{TriviaQA, containment, $500$ questions}}\\
$1$ & $0.000$ & $0.280$ & $+0.000$ $[+0.000, +0.000]$ & $0.280$ & $+0.000$ $[+0.000, +0.000]$ \\
$2$ & $0.693$ & $0.280$ & $+0.000$ $[+0.000, +0.000]$ & $0.282$ & $+0.002$ $[-0.018, +0.022]$ \\
$4$ & $1.386$ & $0.300$ & $+0.020$ $[+0.004, +0.038]$ & $0.284$ & $+0.004$ $[-0.022, +0.032]$ \\
$8$ & $2.079$ & $0.322$ & $+0.042$ $[+0.020, +0.066]$ & $0.256$ & $-0.024$ $[-0.054, +0.006]$ \\
$16$ & $2.773$ & $0.322$ & $+0.042$ $[+0.018, +0.066]$ & $0.242$ & $-0.038$ $[-0.068, -0.008]$ \\
$32$ & $3.466$ & $0.328$ & $+0.048$ $[+0.022, +0.074]$ & $0.252$ & $-0.028$ $[-0.060, +0.002]$ \\
$64$ & $4.159$ & $0.334$ & $+0.054$ $[+0.026, +0.082]$ & $0.266$ & $-0.014$ $[-0.046, +0.018]$ \\
\bottomrule
\end{tabular}
\end{table}

\paragraph{CoTaEval, and the scorer size a task needs.}
CoTaEval's news half \citep{wei2024cotaeval} is a domain no other corpus here uses: $500$ in-domain
article-and-question items scored by SQuAD token-F1 \citep{rajpurkar2016squad}, CoTaEval's own utility
metric, at seven anchors in three families, bands fixed in advance. \textbf{Four anchors cannot read a
news article} ($n=1$ F1 of $0.003$ to $0.088$ at both Pleias and both KL3M models, against a registered
floor of $0.10$), so no band was computed for them. At the four that can, the $7$B pointwise reward
\emph{turns over at every one} (Table~\ref{tab:scorerladder}): at the headline anchor it gives away $43\%$
of the anchor's own F1 by $n=64$, and reproduces on a disjoint re-draw. Majority vote, with no scorer to
overoptimise against, neither falls nor climbs ($+0.0284$ $[-0.0012,+0.0576]$). We registered the opposite
prediction and it was wrong. Swapping the scorer to $14$B, nothing else changed, removes the turn-over at two of four anchors, neither of which began to climb (both marginal):
the \textsc{scorer-bound} reading, where we predicted \textsc{scorer-independent} and were wrong; at $72$B
nothing turns over anywhere, the two anchors that flipped at $14$B climb, and the Spearman of accuracy
against $\log n$ inverts on both tasks, CoTaEval $-0.8929 \to +0.7500$ and TriviaQA $-0.6071 \to +0.8571$.
We had registered that the headline anchor's turn-over would survive $72$B; it did not. The comparison is
legitimate because every arm re-scores cached generations: majority vote never consults the scorer, so all
seven of its cells must be identical across scorers, and they are on all six arms; the GSM8K control
climbs at $4.55$ half-widths. The turn-over is therefore rescoped, not withdrawn: it is a property of
optimising a $7$B pointwise proxy on these tasks, the $7$B scorer is the one every selection arm here
uses, and a deployer needs a larger one on a checkable task. It does not transfer to the judged
workload, where the same two larger scorers lift nothing (the scorer-size paragraph above): a statement about the \emph{task}, not the mechanism. The turn-over touches neither the certificate, a claim about the served law that does not depend on the served text being good, nor the GSM8K majority-vote result ($0.320 \to 0.546$), a different task and rule.

\begin{table}[ht]
\centering\footnotesize
\caption{The pointwise reward's effect as its scorer grows: paired gain of the served accuracy over the
anchor's own $n=1$, $95\%$ intervals beneath. Every column re-scores the same cached draws, so only the
scorer changes. TriviaQA is read at $n=16$, where the $7$B fall is deepest.}
\label{tab:scorerladder}
\setlength{\tabcolsep}{4pt}
\begin{tabular}{@{}lrlll@{}}
\toprule
anchor & $n=1$ & $7$B scorer & $14$B scorer & $72$B scorer \\
\midrule
\multicolumn{5}{@{}l}{\emph{CoTaEval news, token-F1, gain at $n=64$}} \\
Comma-7B & $0.4291$ & $-0.1836$ & $-0.1261$ & $+0.0266$ \\
 & & $[-0.2249, -0.1412]$ & $[-0.1668, -0.0851]$ & $[-0.0097, +0.0623]$ \\
\quad disjoint re-draw & $0.4562$ & $-0.2334$ & $-0.1299$ & --- \\
 & & $[-0.2738, -0.1922]$ & $[-0.1707, -0.0890]$ &  \\
Comma-7B (1T) & $0.3496$ & $-0.1124$ & $-0.0140$ & $+0.0616$ \\
 & & $[-0.1543, -0.0711]$ & $[-0.0581, +0.0294]$ & $[+0.0226, +0.1005]$ \\
TinyComma-1.8B & $0.1669$ & $-0.0594$ & $+0.0012$ & $+0.0624$ \\
 & & $[-0.0921, -0.0274]$ & $[-0.0326, +0.0349]$ & $[+0.0327, +0.0932]$ \\
\multicolumn{5}{@{}l}{\emph{TriviaQA, containment, gain at $n=16$}} \\
Comma-7B & $0.280$ & $-0.038$ & $+0.024$ & $+0.036$ \\
 & & $[-0.068, -0.008]$ & $[-0.006, +0.054]$ & $[+0.008, +0.064]$ \\
\bottomrule
\end{tabular}
\end{table}

\paragraph{CoTaEval's infringement half cannot tell a defence from none at this pair, and an adversarial
scorer beats the risky model on it.} We ran it through the decoding path of every arm on record, each
item with the benchmark's own prefix ($500$ of its $1{,}000$ items, the released pair, bands fixed in
advance), and scored every arm against the item's reference with this paper's copying metrics and
CoTaEval's event, ROUGE-L \citep{lin2004rouge} $\ge 0.5$. \emph{No arm reaches the event on any item, the unconstrained
risky model included} ($0/500$ each, Wilson upper bound $0.0076$). On mean ROUGE-L the risky model
sits closer to the reference than the anchor ($0.1589$ $[0.1554, 0.1624]$ against $0.1451$ $[0.1421,
0.1481]$); selection at $n=64$ is no closer than the anchor alone ($-0.0012$ $[-0.0052, +0.0027]$)
and further than the risky model ($-0.0150$ $[-0.0195, -0.0107]$; over non-empty draws alone
$-0.0148$ $[-0.0191, -0.0104]$, the reward having served an empty draft on one item), and the meter
at $k=10$ is the risky model ($+0.0004$ $[-0.0002, +0.0010]$). We predicted that even an adversarial
scorer would stay below the risky model, and it does not: the draw closest to the reference among
the first $64$ is closer than the risky model's one draw by $+0.0370$ $[+0.0336, +0.0404]$.
Proposition~\ref{prop:selection} bounds amplification relative to the \emph{anchor}, not the risky
model, and a maximum over $64$ draws of a score this noisy beats one draw of either; it too reaches the
event on no item
(\texttt{results/\allowbreak onset\_\allowbreak prediction\_\allowbreak cotaeval\_\allowbreak infringement.md}).

\paragraph{How loose is $\log n$?}\label{app:realisedkl}
If no two candidates share a score, $q(y) = n\,p_s(y)\,F(f(y))^{n-1}$ with $F$ the score's CDF under
$p_s$, so the pointwise ratio approaches $n$ as $F \to 1$ and the KL is $\log n - (n-1)/n$; both are upper
bounds approached as $p_s(y^\star) \to 0$. For $200$-token completions a repeated draw essentially never
occurs, and what lowers the realised figures is a tie in the score: over the headline arm's $32{,}000$
rewards that happens on $5.0\%$ of prompts at $n=2$, $13.8\%$ at $n=8$ and $7.0\%$ at $n=64$, with a mean
tied set of $1.31$, so the quoted figures are the realised ones on $86$--$95\%$ of prompts and conservative
on the rest (\texttt{results/\allowbreak selection\_\allowbreak realised\_\allowbreak kl.csv}). Where draws
do repeat, majority vote's served law can be computed rather than bounded: at $n=64$ its realised pathwise
divergence is $0.92$ nats at the median question on GSM8K and $1.02$ on TriviaQA (at most $1.97$ and
$2.43$), against a certificate of $4.16$; its KL is $0.73$ and $0.68$ against a bound of $3.17$. The
certificate also does not depend on the scorer being reproducible: bf16 non-determinism changes which
candidate is served on $5.6\%$ of prompts at $n=64$ from a batch-size change alone and $8.6\%$ across two
hosts (\texttt{results/\allowbreak selection\_\allowbreak argmax\_\allowbreak stability\_\allowbreak note.md}),
and cannot loosen $q(y) \le n\,p_s(y)$ by a single nat.

\section{Leakage, and the premise it rests on}\label{app:sel-premise}

\begin{figure}[ht]
\centering
\includegraphics[width=0.8\textwidth]{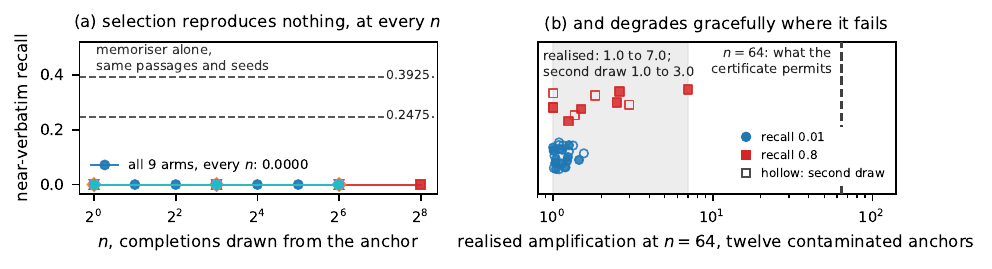}
\caption{\textbf{Selection reproduces no protected passage, and where its premise fails it amplifies
the anchor by a small constant, not by $n$.} \textbf{(a)} Near-verbatim recall, $50$-token windows,
$100$ held-out passages. Nine arms (six anchors, a paraphrase event, an attacker scoring the $n$ draws by
the \emph{memorising model's own likelihood}, and \texttt{Llama-3.1-70B}) read $0.0000$ at every $n$ to
$64$ and at $n=256$, on every passage. Dashed: the memoriser alone on the same passages and seeds,
$0.3925$ mean and $0.8154$ maximum, and the $70$B, which memorised the novel in pre-training, at $0.2475$
unaided, reproducing two passages in full. \textbf{(b)} Twelve anchors fine-tuned on the passages, attacker
selecting at $n=64$: the certificate permits multiplying the anchor's own rate by $64$; the realised factor
is $1.0$ to $7.0$ (filled) and $1.0$ to $3.0$ on a second, independent draw (hollow).}
\label{fig:safety}
\end{figure}

\paragraph{An adversarial selector.}
Table~\ref{tab:extraction} scores the anchor's $n$ draws by the \emph{memorising} model's likelihood, the
attacker choosing the rule, on the same $100$ protected passages and seeds as the meter's onset analysis.
No draw in any pool reaches near-verbatim recall above zero or ROUGE-L $0.5$, whichever one is served.
Recounted at every ROUGE-L threshold from $0.3$ to $0.7$, the maximum over all $256$ draws of each clean
pool reaches none of the $100$ passages, while the memoriser alone reaches $71$, $55$, $47$, $43$ and $39$
of them.\label{app:rougethreshold}

\begin{table}[h]
\centering\footnotesize
\caption{An adversarial selector leaks nothing. $n$ anchor draws scored by the memorising model's
likelihood, $100$ protected passages. \emph{risky alone} is the $k=-1$ baseline and $n=1$ the $k=0$
baseline, the base rate Proposition~\ref{prop:selection} multiplies. The $n=256$ column is a later arm with its own
bands on the same passages and seeds; its grid is $1, 8, 64, 256$, so $n=128$ is measured by neither. Recall and longest common substring
(in words; its floor is the one or two common words any two English texts share) are means over
passages; the ROUGE-L row counts passages, scored only on the grid its arm ran, so a dash means not
measured. A zero on $100$ passages bounds the per-passage rate below $3.0\%$ at $95\%$ for these
passages, seeds and prompts; unseen works, paraphrases and adaptive queries are covered only by the
certificate.}
\label{tab:extraction}
\setlength{\tabcolsep}{4pt}
\begin{tabular}{@{}lcccccccc@{}}
\toprule
$n$ & $1$ & $4$ & $8$ & $16$ & $32$ & $64$ & $256$ & risky alone \\
\midrule
near-verbatim recall     & $0.0000$ & $0.0000$ & $0.0000$ & $0.0000$ & $0.0000$ & $0.0000$ & $0.0000$ & $0.3925$ \\
longest substring, words & $1.59$ & $1.76$ & $1.73$ & $1.66$ & $1.58$ & $1.56$ & $1.56$ & --- \\
ROUGE-L $\ge 0.5$ of $100$ & $0$ & --- & $0$ & --- & --- & $0$ & $0$ & $47$ \\
\bottomrule
\end{tabular}
\end{table}

\paragraph{A natural memoriser.}
Against a model that memorised a work in pre-training the zero holds. Given $100$ raw tokens of each of
the $50$ \emph{Harry Potter} passages, Llama-3.1-70B alone recovers $0.2475$ mean near-verbatim recall,
clears the threshold on $25$ passages and reproduces \textbf{two in full} (recall $1.0000$). Selection
with the $70$B's own likelihood as the scorer recovers $0.0000$ at $n=1$, $8$ and $64$, maximum over all
$50$ passages $0.0000$, longest common substring never above three words. Three earlier arms measured
nothing and are ours, not the mechanism's: one handed a base model an instruction, one a $20$-token seed
where the reference used $100$, one the wrong book.

\paragraph{Outside English prose.}\label{app:multilingual}
A memoriser fine-tuned on $128$ passages of four public-domain works in French and German (splits
disjoint by book) is stronger than the English one: sampled $k=-1$ recall $0.5455$ with $94\%$ of passages
above $0.01$ and $83\%$ at ROUGE-L $\ge 0.5$, against $0.3925$ and $78\%$. The anchor is Pleias-3B, which
is multilingual and openly licensed, so that ``selection reproduces nothing'' is informative. Selection reads $0.0000$ at $n=1$, $8$ and $64$, maximum $0.0000$, with a longest common substring of $1.51$--$1.53$ words and ROUGE-L $0.070$--$0.082$ against the memoriser's $0.678$; $0$ of $100$ passages reach ROUGE-L $\ge 0.3$
where the memoriser reaches $93\%$; the anchor's own leakage is $0.0000$. The anchor's utility in these
languages is unmeasured.

\paragraph{Short works.}\label{app:shortworks}
A short work is where a small certificate can meet $S$. We split $1{,}267$ famous quotations of $8$ to $60$
words (\texttt{Abirate/\allowbreak english\_\allowbreak quotes}, from Goodreads) in half. Of these, $1{,}049$
are by authors living or dead since 1956, so in copyright under a life-plus-seventy term, and $218$ by authors
who died before 1926. The prompt is an opening quotation mark and the first half, and a mechanism leaks when
its continuation starts with the second half, case and punctuation ignored (registered in
\texttt{results/\allowbreak onset\_\allowbreak prediction\_\allowbreak short\_\allowbreak works.md}; seven of
seven predictions right). The anchor needs a median $35.6$ nats for a protected second half, and $16.2\%$ need
$20$ or fewer. Read from the start of a document, a whole quotation needs a median $91.1$, and none needs $20$
or fewer. Table~\ref{tab:shortworks} gives the rates. The meter's certificate is void for $44.5\%$ of protected
second halves at $k=0.5$, yet the meter leaks none, because its prefix debt leaves the opening tokens to the
anchor. It leaks $9.74\%$ at $k=3$ and $69.82\%$ at $k=10$, against the $70$B's $76.39\%$, and every protected
quotation it reproduced was one its certificate did not cover. Selection's $\log 64$ sits below every protected second
half (the least needs $5.69$ nats). Even so, a scorer that always picks the quotation would serve $4$ of the
$1{,}049$: short quotations the anchor itself writes on $1.6\%$ to $14.1\%$ of its draws. Ignoring case and
punctuation makes that event far likelier than the exact string whose surprisal we measured, so the certificate
is void for the event that leaked, as Proposition~\ref{prop:selection} allows. For the same reason every
vacuity share here is a lower bound. We tested quotations at one pair, and no lyrics or whole poems.

\begin{table}[h]
\centering\footnotesize
\caption{Short works: the second half of a famous quotation, given the first, at the audited pair
($T_{\max} = 64$ new tokens, so $K = 64k$). A rate is the percentage of draws that continue with the second
half, case and punctuation ignored; for selection it is the percentage of quotations for which any of the $n$
anchor draws does, which bounds every scorer. \emph{Void for}: the share of the $1{,}049$ protected second
halves whose exact string has anchor surprisal $S \le K$. $95\%$ intervals resample authors
(\texttt{results/\allowbreak short\_\allowbreak works.csv}).}\label{tab:shortworks}
\setlength{\tabcolsep}{4pt}
\begin{tabular}{@{}lcccc@{}}
\toprule
mechanism & $K$, nats & void for & protected, \% & public domain, \% \\
\midrule
anchor alone, $k=0$ & $0$ & --- & $0.03$ $[0.00, 0.06]$ & $0.37$ $[0.00, 0.94]$ \\
selection, worst case, $n=8$ & $2.08$ & $0.0\%$ & $0.19$ $[0.00, 0.45]$ & $1.83$ $[0.00, 5.21]$ \\
selection, worst case, $n=64$ & $4.16$ & $0.0\%$ & $0.38$ $[0.09, 0.75]$ & $3.67$ $[0.00, 8.33]$ \\
meter, $k=\log(64)/64$ & $4.16$ & $0.0\%$ & $0.00$ $[0.00, 0.00]$ & $0.40$ $[0.00, 1.11]$ \\
meter, $k=0.5$ & $32$ & $44.5\%$ & $0.00$ $[0.00, 0.00]$ & $0.40$ $[0.00, 1.11]$ \\
meter, $k=1$ & $64$ & $78.9\%$ & $0.25$ $[0.10, 0.45]$ & $0.92$ $[0.26, 2.02]$ \\
meter, $k=3$ & $192$ & $99.9\%$ & $9.74$ $[7.98, 11.69]$ & $15.48$ $[12.61, 18.87]$ \\
meter, $k=10$ & $640$ & $100.0\%$ & $69.82$ $[64.01, 75.13]$ & $78.33$ $[75.84, 82.22]$ \\
the $70$B alone, $k=-1$ & --- & --- & $76.39$ $[70.53, 81.62]$ & $81.05$ $[79.33, 83.25]$ \\
\bottomrule
\end{tabular}
\end{table}

\paragraph{Vetting the anchor.}\label{app:vetting}
$\Pr_{p_s}[E]$ is the one quantity in the construction measurable with no attack, no risky model and no
access to the mechanism: draw from the candidate on the protected passages' own prefixes and score
near-verbatim recall. At a $100$-token raw prefix on $50$ passages, every anchor used here reads exactly
$0.000$, while \texttt{OLMo-2-7B} leaks on $2$ of $50$ passages and \texttt{OLMo-2-13B} on $6$ (up to
$0.566$ of one), which is why the open-\emph{data} families are excluded, and Llama-3.1-70B leaks on
$0.500$. The prefix length is part of the test (Table~\ref{tab:vetladder}): the $70$B is nearly invisible
at $20$ and $35$ tokens and seen from $50$, and the screen is not monotone, against our registered
prediction: \texttt{OLMo-2-13B} leaks on $6$ passages at $100$ and $150$ tokens and on $3$ at $200$, where only $57$--$94$ tokens of each passage remain. It still reproduces one passage in full there, so the screen did not go blind. Screened instead at a $20$-token seed that includes the \texttt{Complete the prefix:} header, the $70$B
also reads $0.0000$: at about fourteen genuine tokens the screen cannot tell a clean anchor from a model
that holds the book. The certificate holds prompt by prompt, $\Pr_q[E \mid x] \le n\Pr_{p_s}[E \mid x]$, and the user
chooses $x$, so the prefix is part of the attack surface, not a parameter of the test. The
recommendation is therefore a schedule: screen at every prefix length up to the
longest genuine prefix the deployment accepts. Across the three web-trained models the leaking fraction
rises with scale ($0.040$, $0.120$, $0.500$), a pattern on three points that confounds scale with corpus. The six anchors read $0$ of $50$ at all seven lengths. Twenty-six of the new rungs ran on a second machine (host check in Table~\ref{tab:vetladder}). Every rung drew at temperature $1.0$. At the judge-free runs' $0.7$,
TinyComma and Comma-7B read $0$ of $50$ at $20$, $100$ and $200$ tokens while the $70$B leaks on $33$ of $50$
at $100$ ($25$ at $1.0$), and tempering without a penalty raises the anchors' window surprisal $S_w$ from $159.8$ to $179.1$
nats and, over $50$ of Comma-7B's own tokens, from $121.7$ to $135.1$, so margins quoted against the
untempered anchor are conservative for the tempered runs (\texttt{results/\allowbreak vetting\_\allowbreak t07.csv}; with the penalty $1.1$ the
window's is $177.4$, Table~\ref{tab:served}). The test is sound but not complete: recall at a given prefix is a lower bound on
contamination, so passing is evidence and failing is a demonstration.

\begin{table}[ht]
\centering\footnotesize
\caption{The vetting screen at seven prefix lengths, temperature $1.0$: passages of $50$ on which a model leaks (recall
$> 0$; the maximum over $64$ draws in the anchor slot, the single draw for the $70$B). A dash is a rung not run. The licensed anchors' rungs below $100$ were run once the screen proved not monotone; they and the two rungs marked $^\dagger$ ran on a second machine, declared before each ran (the memoriser's seeded draw re-run there
reaches recall $0.01$ on $82$ of $100$ passages against $78$ here, $z = 0.71$).}
\label{tab:vetladder}
\setlength{\tabcolsep}{4pt}
\begin{tabular}{@{}llccccccc@{}}
\toprule
 & & \multicolumn{7}{c}{prefix, raw tokens} \\
\cmidrule(l){3-9}
model & training data & $20$ & $35$ & $50$ & $75$ & $100$ & $150$ & $200$ \\
\midrule
Llama-3.1-70B & memorised in pre-training & $1$ & $2$ & $11$ & $20$ & $25$ & $31$ & $30$ \\
OLMo-2-13B & open data & $0$ & --- & $1$ & --- & $6$ & $6$ & $3$ \\
OLMo-2-7B & open data & $0$ & --- & $1$ & --- & $2$ & $1$ & $2$ \\
\midrule
TinyComma-1.8B & openly licensed & $0$ & $0$ & $0$ & $0$ & $0$ & $0^\dagger$ & $0$ \\
Comma-7B & openly licensed & $0$ & $0$ & $0$ & $0$ & $0$ & $0$ & $0$ \\
Comma-1T & openly licensed & $0$ & $0$ & $0$ & $0$ & $0$ & $0$ & $0$ \\
KL3M-1.7B & openly licensed & $0$ & $0$ & $0$ & $0$ & $0$ & $0^\dagger$ & $0$ \\
Pleias-1.2B & openly licensed & $0$ & $0$ & $0$ & $0$ & $0$ & $0$ & $0$ \\
Pleias-3B & openly licensed & $0$ & $0$ & $0$ & $0$ & $0$ & $0$ & $0$ \\
\bottomrule
\end{tabular}
\end{table}

\paragraph{When the premise fails.}\label{app:contaminated}\label{app:sel-contaminated}
With $\Pr_{p_s}[E] = 0$ the bound $\Pr_q[E] \le n \Pr_{p_s}[E]$ is met by anything we could measure, so
we tested it where it can fail: twelve deliberately memorised anchors (LoRA copies fine-tuned on the
protected passages, GPT-NeoX, Llama, Mixtral, Phi and Qwen, $170$M to $7$B), the same adversarial
selector, the same $100$ passages and seeds, and all twelve reproducing the memoriser's own control at
$0.3925$ mean recall. The selector indexes each padded candidate from its first real token; a version
that read the left-padded batch as right-padded ranked partly on padding, and no number from it is
quoted. Table~\ref{tab:contam} and Figure~\ref{fig:safety}(b) give $E = \{\text{recall} \ge 0.8\}$.
\textbf{Six anchors never enter $E$} in $25{,}600$ draws. At the six that do, the adversary raises the
one-draw rate by $1.0$ to $7.0\times$ at $n=64$, at most $7.0/64$ of what the certificate allows, and by
$1.25$ to $10.0\times$ at $n=256$; a second draw under a disjoint seed reads $1.0$ to $3.0\times$ at $n=64$ and $1.0$ to $4.0\times$ at $256$, because the one-draw denominator is a count over $100$ passages (Llama-3.2-1B's moved
from one passage to three while its per-draw rate moved from $0.0179$ to $0.0182$). Against the per-draw rate the served rates are $1.12$ to $3.91\times$ at $n=64$ and $1.12$ to $5.59\times$ at $256$ on the first draw, $1.12$ to $4.95\times$ and $1.12$ to $6.61\times$ on the second. Where an anchor leaks
at all the bound is vacuous by $n=256$ ($256\hat p$ from $1.73$ to $26.40$, and $1.85$ to $26.13$ on the second draw), and what caps the served
rate is what the pool holds. The band we registered, $A(64) \le 4$ at every anchor, is crossed by one
anchor on the first draw and by none on the second, so we give both readings. Selection multiplies what
the anchor already leaks; the anchor is the thing to vet.

\begin{table}[t]
\centering\footnotesize
\caption{Contaminated anchors, corrected selector: the rate of $E=\{\text{near-verbatim recall}\ge 0.8\}$
served at each $n$ over $100$ passages; $\hat p$, the anchor's per-draw rate over its $25{,}600$ draws; the
fraction of passages whose $256$ draws hold any draw in $E$; and $A(n)$, the served rate over the one-draw
rate, on this draw and, after the slash, on a disjoint-seed re-draw. The certificate permits $n$.}
\label{tab:contam}
\begin{tabular}{lrrrrrrrr}
\toprule
anchor & $n=1$ & $8$ & $64$ & $256$ & $\hat p$ & pool holds $E$ & $A(64)$ & $A(256)$ \\
\midrule
Qwen2.5-7B & $0.10$ & $0.15$ & $0.15$ & $0.15$ & $0.1031$ & $0.17$ & $1.5$\,/\,$1.36$ & $1.5$\,/\,$1.36$ \\
Llama-3.2-3B & $0.05$ & $0.11$ & $0.13$ & $0.15$ & $0.0469$ & $0.16$ & $2.6$\,/\,$1.83$ & $3.0$\,/\,$1.83$ \\
Pleias-1.2B & $0.04$ & $0.05$ & $0.05$ & $0.05$ & $0.0447$ & $0.05$ & $1.25$\,/\,$1.0$ & $1.25$\,/\,$1.0$ \\
Pleias-350M & $0.02$ & $0.05$ & $0.05$ & $0.05$ & $0.0417$ & $0.05$ & $2.5$\,/\,$1.0$ & $2.5$\,/\,$1.0$ \\
Llama-3.2-1B & $0.01$ & $0.04$ & $0.07$ & $0.10$ & $0.0179$ & $0.12$ & $7.0$\,/\,$3.0$ & $10.0$\,/\,$4.0$ \\
Phi-3.5-mini & $0.01$ & $0.01$ & $0.01$ & $0.02$ & $0.0068$ & $0.02$ & $1.0$\,/\,--- & $2.0$\,/\,--- \\
KL3M-1.7B & $0.00$ & $0.00$ & $0.00$ & $0.00$ & $0.0000$ & $0.00$ & --- & --- \\
KL3M-170M & $0.00$ & $0.00$ & $0.00$ & $0.00$ & $0.0000$ & $0.00$ & --- & --- \\
KL3M-3.7B & $0.00$ & $0.00$ & $0.00$ & $0.00$ & $0.0000$ & $0.00$ & --- & --- \\
KL3M-520M & $0.00$ & $0.00$ & $0.00$ & $0.00$ & $0.0000$ & $0.00$ & --- & --- \\
OpenCALM-1B & $0.00$ & $0.00$ & $0.00$ & $0.00$ & $0.0000$ & $0.00$ & --- & --- \\
OpenCALM-3B & $0.00$ & $0.00$ & $0.00$ & $0.00$ & $0.0000$ & $0.00$ & --- & --- \\
\bottomrule
\end{tabular}
\end{table}

\paragraph{A per-work certificate a rights-holder can audit.}\label{app:perwork}
The bound needs two numbers: $n$, which the deployer declares, and $\Pr_{p_s}[E_x]$, which the public
anchor determines. For the exact continuation $x_{>c}$ of a work served for its own prefix the second is
$e^{-S(x_{>c} \mid x_{\le c})}$, and after $m$ adaptive queries the bound is $m\,n\,e^{-S}$: at the
$50$-token window, $S_w = 159.8$ nats, it is below $10^{-60}$ at $n=64$ even after $m=10^4$ queries. A
near-verbatim event has no closed form, so the rights-holder samples: $N$ anchor draws with no event give
$\Pr_{p_s}[E_x] \le 3/N$ at $95\%$ and so $\Pr_q[E_x] \le 3n/N$. Certifying one work at $1\%$ with $n=64$
takes $19{,}200$ draws; our $n=256$ pools certify the average passage at $3 \times 64/25{,}600 = 0.75\%$,
which is a statement about a passage drawn at random from the corpus and not about any one work. At the
measured anchor-only serving rate ($0.0223$ seconds a draw on one A100), one work costs $0.12$ GPU-hours at
$n=64$ and $1\%$ and $0.01$ at $n=8$; a catalogue of $10^5$ works costs $11{,}901$ and $1{,}488$ GPU-hours, and
$10^6$ works ten times that (\texttt{results/\allowbreak audit\_\allowbreak cost.csv}). The draws are
per work and reusable across queries, so the audit is paid once per catalogue, not per user.
\emph{What the audit covers.} The bound $3n/N$ holds at $95\%$ confidence, conditional on \emph{zero}
events among the $N$ clean anchor draws. It covers one near-verbatim event under one fixed prompt
distribution: the work's own prefix, the prompt the draws were taken under. Extending it to other prompts
requires auditing those prompts too, for example every accepted prefix length, or a finite prompt family
under a union bound. Adversarial queries pooled across users are not covered by the audit. They are
covered by composition, at $m\log n$ for $m$ queries (Section~\ref{sec:compose}), and that transcript
bound holds whatever the scorer. A vetted scorer is an operational condition that the certificate cannot
verify.
For a pathwise-metered decoder at the published $k = 0.5$ the multiplier is
$e^{100} \approx 2.7 \times 10^{43}$, and no sample bounds an event below its reciprocal.

\paragraph{The channel the certificate prices.}\label{app:channel}
One response carries at most $\log_2 n$ bits about anything. A scorer colluding with a receiver through a
shared key, serving a draw whose keyed hash falls in the cell that encodes the next bits, realises $1.32$,
$2.22$ and $3.97$ bits per response at $n = 8, 16, 64$ against $\log_2 n = 3, 4, 6$, short of the i.i.d.\
$4.23$ at $n=64$ because $12.6\%$ of draws are duplicates. A compressed $50$-token window is $1{,}423$ bits,
so it takes $357$ responses at $n=64$ and $237$ even at capacity; the $400$-nat odometer stops the channel at
$\lfloor 400/\ln n \rfloor = 96$ responses and $400/\ln 2 = 577$ bits whatever $n$ is, less than one
compressed window. Only a code built on the anchor itself fits: the window's surprisal is $231$ bits.

\paragraph{The attack the composed certificate prices.}\label{app:prefixext}
A user holds a memorised passage's $20$-token seed and extends it one query at a time. A scorer that
knows the work serves, among $n$ fresh draws from the anchor, the one whose first token the leakage
table's memoriser finds most likely, and the user appends that token; since only first tokens are read,
each query is simulated exactly by $n$ draws from the anchor's next-token law (CPU, float32, the
leakage table's $100$ passages, pre-registered in
\texttt{results/\allowbreak onset\_\allowbreak prediction\_\allowbreak prefix\_\allowbreak extension.md}).
After $50$ queries the transcript is certified at $50\log n$ nats, against a window surprisal whose
median is $179.5$ nats (Table~\ref{tab:prefixext}).

\begin{table}[ht]
\centering\footnotesize
\caption{\textbf{Prefix extension through a colluding scorer}, $50$ queries on each of $100$ memorised
passages. ``Vacuous'' counts windows whose surprisal under the anchor is at most $50\log n$; ``token in
pool'' is the share of queries whose $n$ draws contain the true next token.}
\label{tab:prefixext}
\begin{tabular}{lrrrrrr}
\toprule
arm & $50\log n$ & vacuous & token accuracy & exact & near-verbatim & token in pool \\
\midrule
$n=1$ & $0$ & $0$ & $0.009$ & $0$ & $0.0000$ & $0.009$ \\
$n=8$ & $104.0$ & $0$ & $0.026$ & $0$ & $0.0000$ & $0.045$ \\
$n=64$ & $207.9$ & $88$ & $0.057$ & $0$ & $0.0062$ & $0.123$ \\
memoriser, greedy & --- & --- & $0.843$ & $74$ & $0.8847$ & --- \\
\bottomrule
\end{tabular}
\end{table}

The prediction for $n=64$, token accuracy at least $0.5$ and one exact window, failed on both
halves: the certificate excludes nothing for $88$ of the $100$ windows and the attack reconstructs none,
one passage reaching near-verbatim recall $0.5$. A scorer can serve only what the anchor proposes, so
the chance of serving the true token $x_t$ is $1-(1-p_t)^n \le \min(1, n p_t)$ per query, while the
certificate's $n^{50}\prod_t p_t$ counts in full the factors above one that no query can use. The
certificate is loose for prefix extension and nearly tight for a scorer that sends bits instead of
tokens (above), which is why the deployment rule caps queries rather than relying on this attack's
weakness.

\section{What selection costs}\label{app:sel-cost}

\begin{table}[ht]
\centering\footnotesize
\caption{\textbf{What selection at $n=64$ costs against one metered response}, $200$ new tokens. The
unbatched price is paid in draws: on a server that runs only the anchor the reward pass is $9.3\%$ of
selection's clock. Batched, one request's $64$ draws run as one decode and never touch the risky model,
which the meter runs at every step. Matched compute: the $n$ whose deployable draw cost equals one
metered decode (Appendix~\ref{app:sel-cost}).}
\label{tab:cost}
\setlength{\tabcolsep}{5pt}
\begin{tabular}{@{}lr@{}}
\toprule
price model & selection $\div$ meter \\
\midrule
forward passes, parameters $\times$ tokens & $61.3\times$ \\
wall-clock, anchor-only server, draws unbatched & $21.8\times$ \\
wall-clock per request, draws batched: $8$B pair / $70$B pair & $0.51\times$ / $0.22\times$ \\
\quad against AnchoredByte, Comma-7B $+$ $70$B & $0.061\times$ \\
\midrule
matched compute: $n=2$ at $0.68\times$, judged gain against the meter & $-0.0330$ $[-0.0625, -0.0025]$ \\
\quad TriviaQA, no judge, at matched compute ($n=5.57$): accuracy against the meter at $k=20$ & about $0.13$ vs.\ $0.618$ \\
\bottomrule
\end{tabular}
\end{table}

\paragraph{The clock, and why the FLOP model has it backwards.}\label{app:latency}
Both paths ran on one exclusively held card, interleaved \textsc{sel},\textsc{met},\textsc{met},\textsc{sel},
over the same $40$ prompts, each serving $8{,}179$ tokens, so the ratio of seconds per served token is the
wall-clock ratio (Table~\ref{tab:latency}). The measured ratio is $35.4\times$ against the forward-pass
count's $61.3\times$ ($[32.8, 38.5]$ over the repeat extremes), and \textbf{the reward pass is $9.3\%$ of
selection's wall-clock while the draws are $90.7\%$}: scoring $64$ candidates is one batched forward pass,
drawing them is $64 \times 204$ sequential decode steps. With a free scorer selection would still cost
$32.1\times$, so a smaller scorer can return at most $9.3\%$ of the time; the lever is $n$.

\begin{table}[ht]
\centering\footnotesize
\caption{Serving latency on one exclusively held card; both arms serve the same number of tokens. The
last column is each part's share of selection's own clock.}\label{tab:latency}
\setlength{\tabcolsep}{5pt}
\begin{tabular}{@{}lrrrr@{}}
\toprule
path & repeats (s) & mean (s) & s / served token & share \\
\midrule
selection $n=64$, the $64$ anchor draws & $824.6$, $822.6$ & $823.60$ & $0.1007$ & $\mathbf{90.7\%}$ \\
selection $n=64$, the reward pass       & $85.8$, $83.1$   & $84.44$  & $0.0103$ & $9.3\%$ \\
selection $n=64$, total                 & $910.4$, $905.7$ & $908.03$ & $0.1110$ & $100\%$ \\
metered $k=10$                          & $23.7$, $27.6$   & $25.64$  & $0.0031$ & --- \\
\bottomrule
\end{tabular}
\end{table}

\paragraph{With the loader separated from the work.}\label{app:costgrid}
Each total above contains a one-time model load that a server does not pay per request. On an idle
eight-H100 host we timed both paths at $n \in \{1,2,4,8,16,64\}$ and the meter at one and two
completions, every cell twice (repeat spread $0.3$--$2.0\%$; Table~\ref{tab:costgrid}). The draws are
linear in $n$, $\mathrm{draws}(n) = 10.31 + 6.99n$ seconds at $R^2 = 0.99990$, because the harness realises
$n$ as $n$ seed groups batched across prompts, not candidates. We had explained $35.4\times$ against $61.3\times$ by batching, that $64$ completions in one batch do not cost $64$ sequential ones; this fit falsifies it: they do, and the gap is the load. The loader is $61\%$ of the metered path's
total ($10.97$s of $17.94$s) against $2\%$ of selection's at $n=64$, so per request selection at $n=64$
costs $67.2\times$, $1.10\times$ \emph{above} the forward-pass model. The compute-matched cell on this
clock is $n=1$ at $1.07\times$, where selection has nothing to select from; we had predicted $n=2$ and a
mispricing of $2.2\times$, and measured $n=1$ and $4.47\times$.

\begin{table}[ht]
\centering\footnotesize
\caption{Serving cost with the one-time load separated (\texttt{results/\allowbreak cost\_\allowbreak grid.csv}). \emph{draws} is the draw path's wall-clock, load included; the ratios price the whole request
against one metered completion with the load subtracted from both paths (\emph{loader out}) and kept
(\emph{loader in}, Table~\ref{tab:latency}'s construction).}
\label{tab:costgrid}
\begin{tabular}{@{}lrrr@{}}
\toprule
& draws (s) & loader out & loader in \\
\midrule
metered $k=10$, one completion  & $17.94$ & $1.00\times$ & $1.00\times$ \\
selection $n=1$, $0.5$B scorer  & $17.92$ & $\mathbf{1.07\times}$ & $1.11\times$ \\
selection $n=4$, $0.5$B scorer  & $38.66$ & $4.11\times$ & $2.28\times$ \\
selection $n=64$, $7.6$B scorer & $458.02$ & $67.24\times$ & $26.93\times$ \\
\bottomrule
\end{tabular}
\end{table}

\paragraph{What a selection server pays.}\label{app:anchoronly}
One draw from the anchor cost what one metered decode costs ($6.99$s against $6.97$s) because our harness
forwards \emph{both} models at every step whatever $k$ is, so the draw path ran the $8.03$B model and
discarded it: the right design for an audit, the wrong thing to time. Table~\ref{tab:anchoronly} times the
draw loop four ways on the same $200$ prompts and $200$ new tokens. $B \approx 2C$ ($8.515$ against
$8.926$), so every gap is a model forward, and \textbf{a deployment pays $C/A = 0.35$ of every harness
serving number}: $n=64$ costs $21.8\times$ a metered decode, not $67.2\times$. On $C/D = 0.3414$ the
compute-matched cell is $n=2$ at $0.68\times$, where the paired judged difference is
$-0.0330\,[-0.0625,-0.0025]$: the matched-compute loss survives all three prices (the forward-pass count's
$n=4$, the harness clock's $n=1$, the deployable $n=2$). We expected the discarded model to be
$60$--$80\%$ of the path; it is $34\%$.

\begin{table}[ht]
\centering\footnotesize
\caption{Draw-loop wall-clock on the same $200$ prompts and $200$ new tokens. $A$: the harness path every
published draw cost came from (anchor plus the $8.03$B risky model); $B$: the same loop self-paired on the
anchor; $C$: a plain \texttt{generate()} with the anchor alone, what a best-of-$n$ server runs; $D$: the
metered decoder, which needs both models.}
\label{tab:anchoronly}
\begin{tabular}{@{}lrrrr@{}}
\toprule
batch width & $A$ harness, $+8.03$B & $B$ harness, self-paired & $C$ anchor alone & $D$ metered \\
\midrule
$64$  & $7.517$s  & $6.780$s & $3.349$s & $7.768$s \\
$200$ & $12.838$s & $8.515$s & $\mathbf{4.463}$s & $13.074$s \\
\bottomrule
\end{tabular}
\end{table}

\paragraph{One request with its draws batched.}\label{app:batched}
A server that batches a request's candidates runs one decode loop of width $nW$ over a shared prompt and
never runs the risky model, which the meter runs at every step. Timed as one
\texttt{generate(num\_\allowbreak return\_\allowbreak sequences=$n$)} call and one reward pass against the
meter at $k=10$, every completion exactly $200$ tokens, three repeats per cell (Table~\ref{tab:batched};
\texttt{results/\allowbreak onset\_\allowbreak prediction\_\allowbreak batched\_\allowbreak latency.md}):
at one request per call selection at $n=64$ serves in $0.220\times$ the $70$B meter's time and $1.28\times$ a single draw's, and
$0.514\times$ the $8$B meter's, the latter on its registered band's lower edge and inside the cells'
spread, so that band is not read; at eight requests per call every ratio rises, to $0.696\times$ and
$1.69\times$. The $21.8\times$ stays the throughput price and this is the latency. At the authors'
recommended pair a request takes $68.69$ s against $4.19$ s for selection at Comma-7B with $n=64$,
$0.061\times$. Set beside the judged levels and the certificate, every metered cell at both pairs is
beaten on all three axes, at one request per call by $n=64$ with a level margin of $+0.0655$
$[+0.0410, +0.0920]$ over the $8$B meter at $k=10$ and $+0.1040$ $[+0.0810, +0.1275]$ over the $70$B at
$k=20$, and at eight by $n=8$, whose margin excludes zero only for $k \le 3$
(\texttt{results/\allowbreak pareto\_\allowbreak frontier.csv}, descriptive).

\paragraph{The meter as speculative decoding.}\label{app:speculative}
The meter can be served by exact speculative sampling with the anchor drafting. A drafted token is kept
with probability $\alpha_t = \sum_x \min\{p_s(x), p^*_t(x)\}$, and the served law is unchanged. We
teacher-forced both models on the $8$B meter's own served tokens and rebuilt $p^*_t$ from its logged weights,
which reproduces the served token's logged probability to a median $0.0002$. Over $500$ trajectories the
mean acceptance is $0.599$ at $k=10$ and $k=3$ and $0.681$ at $k=0.5$. We charge a round that drafts $g$
tokens as $g$ anchor steps plus one risky step, which is optimistic for the meter. The best round then
($g=1$) serves $1.59$ tokens per risky forward, and selection at $n=64$ takes $0.880\times$ the meter's time
instead of $0.514\times$ ($0.928\times$ at $k=0.5$). Parity would need an acceptance of $0.802$. At the $70$B
pair the acceptance is $0.615$ at $k=20$ and $0.618$ at $k=3$, and the best round ($g=2$) takes selection from
$0.220\times$ to $0.370\times$ the meter's time. The anchor's steps are cheap beside the $70$B's, and even
perfect acceptance would leave selection at $0.901\times$ for drafts up to $8$ tokens. These bounds are on
latency at one request per call, not timings
(\texttt{results/\allowbreak speculative\_\allowbreak acceptance\_\allowbreak note.md}, post hoc).

\begin{table}[ht]
\centering\footnotesize
\caption{Per-request seconds at one and at eight requests per call, $200$ new tokens per completion,
H100s; selection is one batched \texttt{generate} and one reward pass, the meter runs at $k=10$ with both
models every step (AnchoredByte at $k=0.5$ per byte).}\label{tab:batched}
\begin{tabular}{@{}lrr@{}}
\toprule
served by & $W=1$ & $W=8$ \\
\midrule
selection, $n=1$ / $8$ / $64$ & $2.41$ / $2.49$ / $3.09$ & $0.336$ / $0.412$ / $1.46$ \\
meter, $8$B pair & $6.00$ & $0.866$ \\
meter, $70$B pair & $14.84$ & $2.09$ \\
$8$B alone / $70$B alone & $3.19$ / $12.19$ & $0.462$ / $1.74$ \\
AnchoredByte, Comma-7B $+$ $70$B & $68.69$ & $10.91$ \\
selection at Comma-7B, $n=1$ / $8$ / $64$ & $3.09$ / $3.12$ / $4.19$ & $0.423$ / $0.524$ / --- \\
\bottomrule
\end{tabular}
\end{table}

\section{The metered decoder, audited}\label{app:onset}

\paragraph{Both factors of Proposition~\ref{prop:imitation}.}
Table~\ref{tab:imitation} measures them at the audited pair. Below the imitation rate the bucket binds
almost everywhere and the decoder spends most of its allowance; above it the bucket goes slack, the
realised rate stops at $0.857$ nats per token, and the certificate at $k=20$ is written against a quantity
the decoder uses $4.3\%$ of. The crossover is the pair's mean $D_{\mathrm{KL}}(p_{r,t}\|p_{s,t})$. $\beta$ counts every step the
bucket touches, including the prefix-debt opening where the anchor writes the token outright;
strictly interior blends alone are $0.9\%$ of steps at $k=3$. On the
$3{,}300$ protected-passage trajectories the rate is higher, $0.911$ nats per token: the meter charges most
where the risky model is most distinct from the anchor, which is where memorisation lives. The spend is
spread, not concentrated: the busiest $1\%$ of steps carry $5.8\%$ of it and covering $90\%$ takes $64\%$ of
the sequence. At $k=20$ the decoder serves $p_{r,t}$ unchanged at every step, so $N_\varepsilon = \Theta(T)$
for every $\varepsilon$ below the imitation rate, the vacuous horn in Proposition~\ref{prop:sparse}'s own
currency. Table~\ref{tab:imitation-pairs} repeats the measurement at four pairs: the rate saturates (ratio
against a band of $1.10$ fixed in advance), realised spend converges on it to within $0.14\%$, and at $k=20$
the decoder serves $p_r$ unchanged at $\ge 99.95\%$ of steps. The level orders by how far the risky model
has been \emph{tuned} from the anchor, not by its size: a $70$B base model reads $0.8330$ against the $8$B
instruction-tuned model's $0.8278$ at $k=3$, where we had predicted it higher. The same-family prediction
(instruct rate below half the base rate) passed narrowly, by $2.2\%$ of its threshold.

\begin{table}[h]
\centering\footnotesize
\caption{\textbf{Both factors of Proposition~\ref{prop:imitation}, $16{,}500$ logged trajectories at the
audited pair.} $\beta$: fraction of steps at which the bucket constrains the served law, strict blends
plus tokens the anchor writes outright, averaged per trajectory (strict blends alone, Table~\ref{tab:served}'s
\emph{active}, are $0.053$ of steps at $k=1$ against $\beta = 0.2057$); imitation rate: mean
$D_{\mathrm{KL}}(p_{r,t}\|p_{s,t})$ over the slack steps; realised rate and spend: what the decoder
actually incurs, a realisation throughout; $K = kT_{\max}$, $T_{\max} = 200$, the published cap. The last
column regresses cumulative spend on the step index within a trajectory.}
\label{tab:imitation}
\setlength{\tabcolsep}{5pt}
\begin{tabular}{rrrrrrrr}
\toprule
$k$ & $K$ & $\beta$ & imitation rate & realised rate & spend & spend\,/\,$K$ & $R^2$ \\
 & (nats) & (binding) & (nats/token) & (nats/token) & (nats) & & (cum.\ vs.\ $t$) \\
\midrule
$0.1$ & $20$ & $0.9540$ & $0.034$ & $0.051$ & $9.3$ & $46.5\%$ & $0.9794$ \\
$0.15$ & $30$ & $0.9249$ & $0.063$ & $0.093$ & $16.9$ & $56.3\%$ & $0.9942$ \\
$0.25$ & $50$ & $0.8507$ & $0.135$ & $0.180$ & $33.2$ & $66.4\%$ & $0.9987$ \\
$0.5$ & $100$ & $0.5915$ & $0.359$ & $0.404$ & $78.5$ & $78.5\%$ & $0.9994$ \\
$1$ & $200$ & $0.2057$ & $0.665$ & $0.676$ & $134.1$ & $67.1\%$ & $0.9940$ \\
$3$ & $600$ & $0.0332$ & $0.828$ & $0.828$ & $165.0$ & $27.5\%$ & $0.9915$ \\
$5$ & $1000$ & $0.0074$ & $0.848$ & $0.850$ & $169.9$ & $17.0\%$ & $0.9915$ \\
$10$ & $2000$ & $0.0001$ & $0.857$ & $0.857$ & $171.3$ & $8.6\%$ & $0.9915$ \\
$20$ & $4000$ & $0.0000$ & $0.857$ & $0.857$ & $171.3$ & $4.3\%$ & $0.9916$ \\
\bottomrule
\end{tabular}
\end{table}

\begin{table}[h]
\centering\footnotesize
\caption{\textbf{Proposition~\ref{prop:imitation}'s shape at four pairs}, $500$ ordinary prompts, same grid
and cap. $r_{\rm imit}(k)$ is the imitation rate at budget $k$; the ratio tests saturation against a band
of $1.10$ fixed in advance. Only the last two anchors are not safe models, and no certificate, leakage or
$s(x)$ number is taken from them.}
\label{tab:imitation-pairs}
\setlength{\tabcolsep}{4pt}
\begin{tabular}{@{}lccccc@{}}
\toprule
anchor, risky model & $r_{\rm imit}(3)$ & $r_{\rm imit}(20)$ & ratio & $|r_{\rm real}/r_{\rm imit}-1|$ & spend, $k{=}20$ \\
\midrule
TinyComma-1.8B, 8B-Inst.  & $0.8278$ & $0.8570$ & $1.035$ & $0.0005$ & $171.3$ \\
TinyComma-1.8B, 70B base  & $0.8330$ & $0.9085$ & $1.091$ & $0.0007$ & $148.9$ \\
Llama-3.2-1B, 8B-Inst.    & $0.6157$ & $0.6174$ & $1.003$ & $0.0005$ & $123.4$ \\
Llama-3.2-3B-Inst., 8B-Inst. & $0.3010$ & $0.3034$ & $1.008$ & $0.0013$ & $60.6$ \\
\bottomrule
\end{tabular}
\end{table}

\paragraph{What a windowed meter would spend.}\label{app:windowlr}
Section~\ref{sec:uncovered} asks what a meter certifying every $50$-token window would pay to serve the
risky model. The per-step log of the risky model's own draws records $p_r$ and $p_s$ of every served
token, so the realised log-ratio $\sum_t \log(p_{r,t}(y_t)/p_{s,t}(y_t))$ of every window is a sum over the
log: $1{,}500$ trajectories on the $500$ ordinary prompts, $299{,}636$ steps before end-of-text,
$226{,}136$ windows (\texttt{results/\allowbreak window\_\allowbreak logratio.csv}). Its median is $40.3$
nats, its $90$th and $99$th percentiles $63.9$ and $92.5$, its maximum $188.7$; the per-step KL summed
over the same windows, what serving $p_r$ costs in expectation, reads $40.7$, $59.6$ and $85.8$. A
trajectory's largest window has median $63.2$ nats and exceeds $125.3$ on $1\%$ of trajectories, and
$0.011\%$ of windows reach the protected window's median surprisal of $159.8$. A sliding-window meter
with a budget near $125$ nats would rarely bind on ordinary text and would certify the median protected
window at $e^{125-159.8}$; it certifies no event whose probability under the anchor must be sampled
(Section~\ref{sec:compose}).

\paragraph{The windowed meter, built.}\label{app:windowed}
\texttt{a\_patch/\allowbreak factory.py} with \texttt{window=50}: at step $t$ the served law may give any
token at most $W$ minus the largest realised log-ratio of a span of the last $49$ served tokens ending at
$t-1$, never less than $0$, so negative ratios earn no credit and every span of at most $50$ tokens is
certified at $W$ conditional on all served text before it, the type we mark $^{c}$ (\texttt{tests/\allowbreak test\_\allowbreak windowed\_\allowbreak meter.py} runs the real decode loop on two small random models and checks every span of every path). It
carries no prefix debt, since the output's spans are what it certifies. On the $500$ prompts it blends on
$100\%$, $94.7\%$, $68.8\%$, $37.9\%$ and $0.08\%$ of steps at $W = 4.16$, $12.48$, $24.95$, $40$ and $125$
($47.1\%$ at $40$ and $0.15\%$ at $125$ under the chat template), and none of $7{,}000$ trajectories has a
span over its $W$ (\texttt{results/\allowbreak windowed\_\allowbreak arms.csv}). Registered in advance
(\texttt{results/\allowbreak onset\_\allowbreak prediction\_\allowbreak windowed\_\allowbreak meter.md}):
the prediction that failed is that selection would still lead at $W=125$; it ties. Table~\ref{tab:windowed}
gives the leakage grid and the chat-served comparison. \emph{What the certificate covers.} A per-span
max-divergence of $W$ bounds every event measurable in the span, given the text before it, by $e^{W}$ times
the anchor's probability of it, so near-verbatim variants are covered; what is missing is that probability.
A rights-holder can bound it by plain sampling, which assumes nothing about the event and needs more than
$3e^{W}$ clean draws for a bound below one; an importance sampler could need far fewer, but its upper
confidence bound then rests on how well its proposal covers the event. The difference from selection's
$e^{4.16}$ is one of degree, not of kind. The certificates also differ in type, and there are three
types here. The windowed meter's is conditional on all served text before the span ($^{c}$). Selection's
is joint over the whole output. Selection in installments certifies a window only conditional on the
served text up to the last block boundary at or before it ($^{b}$, Proposition~\ref{prop:installments}). A
joint or boundary-conditional bound says nothing about a span after an unlikely prefix inside a block
(Remark~\ref{rem:installments}). So a comparison of the windowed meter with installments at the same nats
per span is a comparison across certificate types, not at a matched certificate. For the copyright event
that matters, a protected window appearing anywhere in the response, the joint bound applies directly,
$\Pr_q[E] \le n\Pr_{p_s}[E]$. The $^{c}$ bound instead compares each position with the anchor's
probability given the \emph{served} prefix, which an auditor can sample only by querying the deployment.

\begin{table}[ht]
\centering\footnotesize
\caption{All arms of Table~\ref{tab:matched}, and the rule that never serves an empty draw (\emph{never empty}), one judge pass, $500$ prompts. Certified
max-divergence nats (KL at $k=0.5$) for the $200$-token output and for any $50$-token span (---: only the
output's, which covers the first span). The two span types differ and are not directly comparable:
$^{c}$, conditional on all served text before the span (windowed meter); $^{b}$, conditional on the served
text up to the last block boundary at or before the span's start (installments,
Proposition~\ref{prop:installments}). Rows compared at the same output certificate are matched; rows
compared only at the same span nats are across types. Gain over the arm's own zero-budget
control, judge~B with $95\%$ interval, and judge~G. \emph{Leak}: near-verbatim recall of $100$ memorised
passages, the memoriser as risky model ($0.495$ alone); for selection, the best of its $64$ draws.}
\label{tab:matchedfull}
\setlength{\tabcolsep}{4pt}
\begin{tabular}{@{}lrrlrr@{}}
\toprule
mechanism & span & output & gain, judge~B & judge~G & leak \\
\midrule
anchored decoding, $k=0.5$ & --- & $100$ & $-0.0030$ $[-0.0245, +0.0185]$ & $+0.015$ & --- \\
budget up front & --- & $4.16$ & $+0.0230$ $[+0.0010, +0.0445]$ & $-0.0265$ & --- \\
per-token pathwise meter & --- & $33.27$ & $+0.0125$ $[-0.0065, +0.0315]$ & $-0.0185$ & --- \\
 & --- & $83.18$ & $+0.0160$ $[-0.0050, +0.0370]$ & $+0.0125$ & --- \\
windowed meter, $W$ per span & $4.16^{c}$ & $16.64$ & $+0.0115$ $[-0.0110, +0.0340]$ & $-0.047$ & $0.000$ \\
 & $12.48^{c}$ & $49.91$ & $+0.0190$ $[-0.0035, +0.0415]$ & $-0.0445$ & $0.000$ \\
 & $24.95^{c}$ & $99.81$ & $+0.0375$ $[+0.0150, +0.0600]$ & $-0.0045$ & $0.000$ \\
 & $40^{c}$ & $160$ & $+0.0375$ $[+0.0140, +0.0610]$ & $+0.0475$ & $0.000$ \\
 & $125^{c}$ & $500$ & $+0.0795$ $[+0.0530, +0.1055]$ & $+0.130$ & $0.106$ \\
\midrule
best of $64$ & --- & $4.16$ & $+0.1015$ $[+0.0775, +0.1260]$ & $+0.176$ & $0.000$ \\
best of $64$, never empty & --- & $4.16$ & $+0.1010$ $[+0.0775, +0.1245]$ & $+0.155$ & --- \\
installments, $L=25$ & $12.48^{b}$ & $33.27$ & $+0.1455$ $[+0.1200, +0.1715]$ & $+0.2855$ & --- \\
installments, $L=10$ & $24.95^{b}$ & $83.18$ & $+0.1580$ $[+0.1330, +0.1835]$ & $+0.305$ & --- \\
\bottomrule
\end{tabular}
\end{table}

\begin{table}[ht]
\centering\footnotesize
\caption{The windowed meter by budget. \emph{Recall}: mean near-verbatim recall of $100$ memorised passages
with the memoriser as the risky model, served plainly and through the chat template (the memoriser alone
reads $0.4952$ and $0.2193$, the anchor alone $0$). \emph{Chat}: selection's gain minus the windowed meter's,
both against the chat-served $8$B, judges B and G, $95\%$ intervals.}
\label{tab:windowed}
\setlength{\tabcolsep}{4pt}
\begin{tabular}{@{}rrrll@{}}
\toprule
$W$, nats & recall, plain & recall, chat & chat, judge B & chat, judge G \\
\midrule
$4.16$ & $0.0000$ & --- & --- & --- \\
$12.48$ & $0.0000$ & --- & $+0.0450$ $[+0.0095, +0.0810]$ & $+0.0970$ $[+0.0590, +0.1355]$ \\
$24.95$ & $0.0000$ & $0.0000$ & $+0.0065$ $[-0.0305, +0.0435]$ & $+0.0530$ $[+0.0110, +0.0945]$ \\
$40$ & $0.0000$ & $0.0000$ & $-0.0525$ $[-0.0895, -0.0150]$ & $-0.0295$ $[-0.0725, +0.0145]$ \\
$60$ & $0.0063$ & --- & --- & --- \\
$80$ & $0.0286$ & $0.0043$ & --- & --- \\
$100$ & $0.0634$ & --- & --- & --- \\
$125$ & $0.1058$ & $0.0632$ & $-0.1600$ $[-0.1985, -0.1220]$ & $-0.2815$ $[-0.3335, -0.2295]$ \\
$150$ & $0.1476$ & --- & --- & --- \\
$200$ & $0.3993$ & --- & --- & --- \\
\bottomrule
\end{tabular}
\end{table}

\paragraph{Selection in installments.}\label{app:blockwise}
Both selection rules sample a bounded tilt of the anchor, a law $p_s h$ with $h \le e^{\lambda}$ at every
length: the Gibbs tilt $h = e^{\lambda U}/Z$ of any utility $U \in [0,1]$ is one, best-of-$n$ is another,
$h = nF(f)^{n-1}$ with $F$ the anchor's distribution of a tie-free score $f$, so $\lambda = \log n$, and
installments build one causally, block by block, with the completed candidate's reward as a sampled
value. Section~\ref{sec:uncovered}'s installment rule draws $n$ continuations of up to $L$ tokens from the anchor
alone, scores each by the committed reward of the candidate completed to $200$ tokens by one anchor
rollout, serves the argmax and repeats until $200$ tokens or an end-of-text; at $L = 200$ it is
whole-output selection through the same code, on fresh draws. Each block draws new anchor continuations,
so Proposition~\ref{prop:installments} applies. $B$ blocks certify the output at $B\log n$. A $50$-token
window is certified at $\min(\lceil 50/L\rceil + 1, B)\log n$ only conditional on the served text up to
the last block boundary at or before its start (type $^{b}$). Conditional on the served tokens inside that
block, no bound of this form holds (Remark~\ref{rem:installments}). An earlier version of this work stated
the window certificate given all preceding text; the window figures below are the corrected,
boundary-conditional ones, with unchanged values (Table~\ref{tab:blockwise}; registered in
\texttt{results/\allowbreak onset\_\allowbreak prediction\_\allowbreak blockwise.md}). At $64$ draws per block,
installments of $10$ and $25$ tokens beat one choice among $64$ whole drafts, by $+0.0415$
$[+0.0185, +0.0645]$ and $+0.0245$ $[+0.001, +0.0475]$, against our registered prediction of a tie, and
installments of $50$ tie it. Their window certificates, $24.95$ and $12.48$ nats a window$^{b}$, stay far
below the protected window's $159.8$. Two concessions go with them: the whole-output certificate, $\lceil T/L\rceil\log n$, grows linearly with the output as a meter's does, so installments are not a budget that does not grow, and at $1{,}000$ tokens, with $8$ draws a block, they lose to one
whole-output choice (Appendix~\ref{app:long}); and a near-verbatim window at their per-window certificate is no easier to audit than the windowed meter's at the same $W$, while the two certificates are of different types ($^{b}$ against $^{c}$; Section~\ref{sec:uncovered}). A disjoint draw, registered after this one was read, reproduces it, where we had predicted a tie
at $L = 25$: $+0.0435$
$[+0.020, +0.067]$ and $+0.033$ $[+0.011, +0.0555]$ at $L = 10$ and $25$ under judge~B, $+0.0635$ $[+0.033,
+0.0945]$ and $+0.0595$ $[+0.028, +0.0905]$ under judge~G; read post hoc by judge~G, the first draw's
advantage is $+0.083$, $+0.0635$ and $+0.031$ at $L = 10$, $25$ and $50$
(\texttt{results/\allowbreak blockwise\_\allowbreak replication.csv}). Held instead to one whole-output $\log 64$, spent as $2$, $3$ or $6$
installments of $8$, $4$ or $2$ draws, they lose to one choice among $64$ by $0.0635$, $0.082$ and $0.088$:
a decoder steered by the same reward with its budget of $\log 64$ fixed up front, the closest we ran to a
token-level reward tilt at that budget, does worse than selection spent once. Scored on the prefix alone,
without the rollout, the rule serves $23$ words at the median and loses to the rollout by $0.0395$
$[0.02, 0.059]$. The risky model as planner, scoring each $10$-token block by $\log p_r - \log p_s$ over the
same $64$ draws, gains $+0.077$ $[+0.054, +0.1005]$ over the anchor, $+0.026$ $[+0.001, +0.051]$ more than
the meter does at $k=10$ (post hoc), and loses to the reward by $0.0895$ $[0.064, 0.115]$: the same choice
buys more when the reward makes it than when the risky model does. The memoriser choosing twenty times among $64$ ten-token continuations
reads near-verbatim recall $0.0000$ on all $100$ passages it memorised (alone it reads $0.3925$), and
ROUGE-L $0.1168$ against whole-output selection's $0.1073$. With the prompt and the served text each in
its own slot of the reward's template, every rule serves more empty answers than the committed pool's $41$ ($103$ once, $104$ to $110$ in
installments of $64$ draws, $113$ to $192$ at one $\log 64$), and the planner none.

\begin{table}[ht]
\centering\footnotesize
\caption{Selection in installments, judge~B, order-averaged, one draw each (seed $20260925$). Certificates
in nats for the whole output and for any $50$-token window$^{b}$, that is, conditional on the served text
up to the last block boundary at or before the window's start (Proposition~\ref{prop:installments}); a
window starting on a boundary needs only $\lceil 50/L\rceil\log n$, and for one choice ($L=200$) the only
boundary is the start of the output. No window figure here is conditional on all preceding text. Gains over the
anchor-alone control and differences from one choice among $64$ (first row), paired over the $500$
prompts; $\lceil 200/L\rceil$ blocks (\texttt{results/\allowbreak blockwise.csv}).}\label{tab:blockwise}
\setlength{\tabcolsep}{2.5pt}
\begin{tabular}{@{}lrrcll@{}}
\toprule
rule & $L$ & $n$ & output / window$^{b}$ & gain over the anchor & minus once \\
\midrule
once & $200$ & $64$ & $4.16$ / $4.16$ & $+0.125$ $[+0.1015, +0.1485]$ & --- \\
installments & $50$ & $64$ & $16.64$ / $8.32$ & $+0.129$ $[+0.1045, +0.153]$ & $+0.004$ $[-0.019, +0.0265]$ \\
 & $25$ & $64$ & $33.27$ / $12.48$ & $+0.1495$ $[+0.1245, +0.175]$ & $+0.0245$ $[+0.001, +0.0475]$ \\
 & $10$ & $64$ & $83.18$ / $24.95$ & $+0.1665$ $[+0.141, +0.1925]$ & $+0.0415$ $[+0.0185, +0.0645]$ \\
one $\log 64$ & $100$ & $8$ & $4.16$ / $4.16$ & $+0.0615$ $[+0.0385, +0.085]$ & $-0.0635$ $[-0.085, -0.0415]$ \\
 & $67$ & $4$ & $4.16$ / $2.77$ & $+0.043$ $[+0.0205, +0.0655]$ & $-0.082$ $[-0.1055, -0.0585]$ \\
 & $34$ & $2$ & $4.16$ / $2.08$ & $+0.037$ $[+0.0155, +0.0585]$ & $-0.088$ $[-0.1115, -0.0645]$ \\
prefix only & $50$ & $64$ & $16.64$ / $8.32$ & $+0.0895$ $[+0.0665, +0.1125]$ & --- \\
planner & $10$ & $64$ & $83.18$ / $24.95$ & $+0.077$ $[+0.054, +0.1005]$ & --- \\
\bottomrule
\end{tabular}
\end{table}

\paragraph{CP-$k$'s rejection rule, run.}\label{app:cpk}
CP-$k$ \citep{vyas2023provable} rejection-samples complete outputs of the protected model against safe
models trained on disjoint shards. No model pair here meets that premise, so we ran its rule with the one
safe model we have. For each prompt the risky model's $M = 64$ draws are tested in seed order. The first
whose realised log-ratio $R(y) = \log p_r(y \mid x) - \log p_s(y \mid x)$ over its served tokens is at most
$\kappa$ is served; otherwise the anchor's own draw is. Then $q(y) \le (Me^{\kappa} + 1)\,p_s(y)$ for every
$y$, a pathwise certificate of $C = \log(1 + Me^{\kappa})$ whatever the acceptance rate. CP-$k$'s own rule
retries until it accepts and pays more as acceptance falls: $168.35$ nats at half acceptance, above a
window's $159.8$, and $87.97$ at one output in a hundred (\texttt{results/\allowbreak cpk\_\allowbreak certificate.csv}). Table~\ref{tab:cpk} runs the rule at the certificates of
Table~\ref{tab:matchedfull} (registered in \texttt{results/\allowbreak onset\_\allowbreak prediction\_\allowbreak cpk\_\allowbreak baseline.md}). A risky draw's median $R$ is $169.6$ nats. So at $\log 64$ and at $33.27$
nats the rule serves the anchor on every prompt and gains nothing, and at $83.18$ it serves the risky
model on $26.8\%$ of prompts. Selection beats it at each matched whole-output certificate: best-of-$64$ by $+0.110$
$[+0.0855, +0.135]$ at $\log 64$, and installments by $+0.143$ $[+0.119, +0.168]$ at $33.27$ and $+0.1205$
$[+0.091, +0.1485]$ at $83.18$, all three confirmed under judge~G. Against the per-token pathwise meter at
the same nats it is unresolved: $+0.022$ $[-0.006, +0.0495]$ at $83.18$ and $-0.014$ $[-0.0335, +0.006]$
at $33.27$. We registered that its acceptances would be short outputs. They are not, since nearly every
draw runs to the $200$-token cap. Against the memoriser the rule serves nothing near-verbatim through $250$
nats and leaks from $600$ ($0.0563$ recall; the memoriser alone reads $0.4629$).

\begin{table}[ht]
\centering\footnotesize
\caption{CP-$k$'s rejection rule with the anchor as its safe model, $64$ risky draws per prompt, $500$
prompts. $C$: pathwise certificate in nats, whole output; $\kappa$: the acceptance threshold on the
realised log-ratio; \emph{risky}: share of prompts served a risky draw. Gain over the anchor's own draw,
judge~B with $95\%$ interval, and judge~G. \emph{Leak}: near-verbatim recall of $100$ memorised passages
with the memoriser as risky model.}\label{tab:cpk}
\setlength{\tabcolsep}{4pt}
\begin{tabular}{@{}rrrlrr@{}}
\toprule
$C$ & $\kappa$ & risky & gain, judge~B & judge~G & leak \\
\midrule
$4.16$ & $-0.02$ & $0.0\%$ & $+0.0000$ $[+0.0000, +0.0000]$ & $+0.0000$ & $0.0000$ \\
$33.27$ & $29.11$ & $0.0\%$ & $+0.0000$ $[+0.0000, +0.0000]$ & $+0.0000$ & $0.0000$ \\
$83.18$ & $79.02$ & $26.8\%$ & $+0.0395$ $[+0.0245, +0.0555]$ & $+0.0510$ & $0.0000$ \\
$159.83$ & $155.67$ & $99.6\%$ & $+0.0865$ $[+0.0620, +0.1105]$ & $+0.1335$ & $0.0000$ \\
\bottomrule
\end{tabular}
\end{table}

\paragraph{At $1{,}000$ tokens.}\label{app:long}
Every judged arm above stops at $200$ tokens. We asked whether whole-output selection still buys utility at
$1{,}000$ or needs installments, whose certificate grows with the output, and how either compares with the
meters at matched nats. We ran every mechanism at $T_{\max} = 1000$ on the $500$ prompts and judged it
uncut against the risky model's own $1{,}000$-token draw, registered in advance
(\texttt{results/\allowbreak onset\_\allowbreak prediction\_\allowbreak long\_\allowbreak outputs.md};
Table~\ref{tab:long}). One choice among $64$ whole drafts still gains $+0.063$ $[+0.043, +0.083]$ over the
anchor at the same $\log 64$. At the same whole-output certificate it beats the per-token pathwise meter by
$+0.062$ $[+0.0395, +0.084]$ and the budget granted up front by $+0.0305$ $[+0.0055, +0.056]$. Across certificate types, it also beats the windowed meter at
$W = 4.16$ nats a span$^{c}$ by $+0.0445$ $[+0.020, +0.0695]$; that comparison is not matched, since the windowed meter's
whole-output certificate is $83.18$ nats. It also beats anchored decoding, whose KL certificate is a third type, at $k=0.1$ ($100$ nats) by
$+0.0775$ and at $k=0.5$ ($500$ nats) by $+0.104$, all confirmed under judge~G. Installments of $8$ draws
per $100$ tokens, certified at $20.79$ nats for the output and $4.16$ for any window$^{b}$ (conditional on
the served text up to the last block boundary), gain $+0.0185$
$[-0.0005, +0.037]$. Against the meters at their own $20.79$ nats they are unresolved under~B, $+0.0165$
$[-0.0045, +0.038]$ pathwise and $+0.019$ $[-0.001, +0.0395]$ KL, where we had predicted wins that only
judge~G confirms. They lose to one choice among $64$ by $0.0445$ $[0.026, 0.0635]$ (G: $0.059$), so long
outputs do not need them. The anchor's drafts are short: best-of-$64$ serves a mean $203$ tokens, and
$32.6\%$ of its outputs are longer than $200$, while the risky model always runs to the cap. On the $163$
prompts where the served draft is longer than $200$ tokens its gain is $+0.092$ $[+0.057, +0.127]$. At
this length no meter improves on the anchor by more than $0.033$ under~B, and anchored decoding falls
below it at $k = 0.1$, $0.5$ and $10$. The judges part on the risky model's own $1{,}000$ tokens: G reads
them $+0.1095$ over the anchor and B $-0.0175$.

\begin{table}[ht]
\centering\footnotesize
\caption{Every mechanism at $T_{\max} = 1000$, $500$ prompts, judged uncut against the risky model's own
$1{,}000$-token draw. Certified nats for the output and for any $50$-token span (---: none beyond the output's). Span types:
$^{c}$, conditional on all served text before the span (windowed meter); $^{b}$, conditional on the
served text up to the last block boundary at or before the span's start (installments; for one whole-output
choice the only boundary is the start of the output). The types are not directly comparable. Output
certificates are max-divergence except anchored decoding's KL. \emph{Tokens}: mean
served length. Gain over the arm's own zero-budget control (the anchor alone, or the pool's single draw),
judge~B with $95\%$ interval, and judge~G (\texttt{results/\allowbreak long\_\allowbreak outputs.csv}).}\label{tab:long}
\setlength{\tabcolsep}{3.5pt}
\begin{tabular}{@{}lrrrlr@{}}
\toprule
mechanism & span & output & tokens & gain, judge~B & judge~G \\
\midrule
best of $64$, whole output & $4.16^{b}$ & $4.16$ & $203$ & $+0.0630$ $[+0.0430, +0.0830]$ & $+0.2115$ \\
best of $16$ & $2.77^{b}$ & $2.77$ & $163$ & $+0.0305$ $[+0.0105, +0.0510]$ & $+0.1665$ \\
installments, $L=100$, $n=8$ & $4.16^{b}$ & $20.79$ & $159$ & $+0.0185$ $[-0.0005, +0.0370]$ & $+0.1525$ \\
pathwise meter & --- & $4.16$ & $220$ & $+0.0010$ $[-0.0095, +0.0115]$ & $+0.0070$ \\
 & --- & $20.79$ & $339$ & $+0.0020$ $[-0.0110, +0.0150]$ & $-0.0080$ \\
budget up front & --- & $4.16$ & $275$ & $+0.0325$ $[+0.0160, +0.0490]$ & $-0.0075$ \\
windowed meter & $4.16^{c}$ & $83.18$ & $499$ & $+0.0185$ $[+0.0015, +0.0355]$ & $-0.0555$ \\
anchored decoding, $k=0.0208$ & --- & $20.79$ & $215$ & $-0.0005$ $[-0.0095, +0.0085]$ & $+0.0005$ \\
anchored decoding, $k=0.1$ & --- & $100$ & $461$ & $-0.0145$ $[-0.0280, -0.0010]$ & $-0.0375$ \\
anchored decoding, $k=0.5$ & --- & $500$ & $831$ & $-0.0410$ $[-0.0575, -0.0240]$ & $-0.0365$ \\
anchored decoding, $k=10$ & --- & $10^4$ & $1000$ & $-0.0175$ $[-0.0330, -0.0015]$ & $+0.1095$ \\
\bottomrule
\end{tabular}
\end{table}

\paragraph{Where extraction begins.}
Table~\ref{tab:onset} gives the nine pairs Figure~\ref{fig:horns}(a) plots. Every pair clears the admission
criterion (sampled $k=-1$ recall $\ge 0.10$), and $k=0$, the anchor alone, reads $0.000$ at all nine. Two
columns are reported because they are not controlled: the memorisers span a factor of $5.1$ in their own
recall, and five of the nine did not converge. Part of the residual tracks the words a $20$-token seed
hands the adversary (Spearman $-0.958$), a caution for seeds fixed in tokens
\citep{carlini2023quantifying}. Every sweep decodes $T$ tokens for every passage, the longest target in
it: $296$ at the audited pair, $306$ to $468$ elsewhere and $650$ in the KL3M tokenizer's tokens. The median
target is $0.84$ to $0.89$ of $T$ long, so its own certificate is void from that fraction of $s(x)$,
below the onset at eight of nine pairs (by $0.010$ to $0.273$) and $0.004$ above Pleias-1.2B's; a $50$-token window's is void from $50/T$, $0.08$ to $0.17$ of $s(x)$
(\texttt{results/\allowbreak onset\_\allowbreak window\_\allowbreak threshold.csv}).

\begin{table}[h]
\centering\footnotesize
\caption{Onset of extraction on nine pairs, as a fraction of the anchor's surprisal rate $s(x)$ (nats per
token; $s_r$ the memoriser's, on the same tokens). Onset is the budget at which mean single-query recall
first reaches $0.01$, interpolated inside its bracket; $n$ protected passages. \emph{mem.}: the memoriser's
own sampled recall at $k=-1$. \emph{ep.}: epochs run against the cap, so a run below its cap reached its
stop-loss. $\dagger$: baselines measured in a second invocation reproducing the sweep's protocol line.}
\label{tab:onset}
\setlength{\tabcolsep}{2pt}
\begin{tabular}{@{}lcccccccc@{}}
\toprule
pair & $s(x)$ & $s_r$ & $n$ & mem. & ep. & onset & 95\% CI & onset\,/\,$s(x)$ \\
\midrule
KL3M 1.7B $+$ mem.\ KL3M 1.7B       & $2.21$ & $0.781$ & $100$ & $0.411$ & $40/40$ & $2.58$ & $[2.52, 3.32]$ & $1.166$ \\
Comma 7B $+$ mem.\ Comma 7B$^{\dagger}$ & $2.39$ & $0.179$ & $100$ & $0.719$ & $12/12$ & $2.13$ & $[2.00, 2.92]$ & $0.892$ \\
KL3M 520M $+$ mem.\ KL3M 520M       & $2.42$ & $0.215$ & $100$ & $0.519$ & $11/40$ & $2.54$ & $[2.45, 3.00]$ & $1.053$ \\
Phi-3.5 mini $+$ mem.\ Phi-3.5 mini & $2.84$ & $0.009$ & $100$ & $0.566$ & $40/40$ & $2.63$ & $[2.27, 3.08]$ & $0.926$ \\
Pleias 1.2B $+$ mem.\ Pleias 1.2B   & $3.21$ & $0.365$ & $100$ & $0.909$ & $30/40$ & $2.82$ & $[2.53, 3.08]$ & $0.878$ \\
TinyComma 1.8B $+$ mem.\ Llama-8B$^{\dagger}$ & $3.24$ & $0.194$ & $100$ & $0.492$ & $12/12$ & $2.87$ & $[2.72, 3.45]$ & $0.887$ \\
open-calm 1B $+$ mem.\ open-calm 1B & $3.36$ & $0.416$ & $100$ & $0.181$ & $40/40$ & $3.45$ & $[3.25, 4.97]$ & $1.027$ \\
open-calm 3B $+$ mem.\ open-calm 3B & $3.38$ & $0.314$ & $100$ & $0.924$ & $21/40$ & $3.36$ & $[3.24, 3.63]$ & $0.993$ \\
Pleias 350M $+$ mem.\ Pleias 350M   & $3.55$ & $0.326$ & $458$ & $0.875$ & $16/40$ & $3.27$ & $[3.07, 3.83]$ & $0.920$ \\
\bottomrule
\end{tabular}
\end{table}

\emph{Two refined predictions, both refuted.} The rate a decoder must permit before it can pay for $x$
at all,
\begin{equation}
r(x) \;=\; s_s(x) - s_r(x),
\label{eq:req}
\end{equation}
predicts an onset ratio of $1 - s_r/s_s$ that must move with memorisation. Committed for the last five
pairs before any was swept, it is refuted under all three registered rules: every miss low, held-out errors
of $0.365$, $0.507$ and $0.296$ nats, and a directional claim that inverts. Eq.~\eqref{eq:req} is right
about what a decoder must \emph{afford} (Table~\ref{tab:imitation}'s crossover) and wrong about where
leakage \emph{begins}. The running maximum of Proposition~\ref{prop:outrun} fails in the complementary
direction: warping the anchor's logits to $\tau = 0.4$ moves $s(x)$ by $67\%$ and $62\%$ on two pairs and
the onset follows at an elasticity of $+0.72$ $[+0.41, +0.84]$ and $+0.61$ $[+0.40, +0.96]$, while
$k_{\mathrm{crit}}$ misses by $21.9\%$ against a no-change null's $12.3\%$.

\emph{The level is the claim, not the ordering.}\label{app:collapse} Holding one pair, its corpus, anchor
and grid fixed and varying only the memoriser's fine-tuning length moves the onset ratio from $0.8756$ to
$1.3477$; re-seeding at a fixed epoch count moves it by $0.033$ to $0.080$ where the fine-tune converged
and by $0.2597$ where the memoriser is marginal. \textbf{The within-pair spread is $1.64\times$ the entire
between-pair spread} of $0.8784$ to $1.1658$, and whether a pair's interval excludes $1$ flips inside that
single pair, from $[0.68, 0.98]$ to $[1.01, 1.59]$. On two further corpora (Table~\ref{tab:gutenberg}) two
of three pairs vary by $0.088$ and $0.066$, inside their own bootstrap widths of $0.227$ and $0.287$, and
the third by $0.436$ against $0.171$; on the third corpus the ordering inverts, Pleias-1.2B reading $1.314$
above the fine-tokenizer pair, whose interval no longer excludes $1$. Every onset in these arms lands
between $0.80$ and $1.35$ of $s(x)$, which is what Proposition~\ref{prop:threshold} claims; which pairs
begin above $1$ is confounded with training variables the table does not hold fixed.

\begin{table}[ht]
\centering\footnotesize
\caption{Three pairs re-measured on two further protected corpora, anchor, architecture, settings, seed and
grid held fixed. \emph{novels}: the pair's ratio on the sixteen CopyBench novels (Table~\ref{tab:onset});
\emph{no-cross}: bootstrap resamples that never reach the onset on the grid.}\label{tab:gutenberg}
\setlength{\tabcolsep}{4pt}
\begin{tabular}{@{}lcccccc@{}}
\toprule
pair & onset & onset\,/\,$s(x)$ (95\% CI) & no-cross & Eq.~\eqref{eq:req} & other corpus & novels \\
\midrule
\multicolumn{7}{@{}l}{\emph{600 excerpts of 50 public-domain books}} \\
KL3M-520M    & $2.608$ & $1.102$ $[1.07, 1.42]$ & $0.0\%$ & $2.218$ & --- & $1.053$ \\
Pleias-1.2B  & $2.513$ & $0.895$ $[0.85, 1.17]$ & $0.0\%$ & $2.463$ & --- & $0.878$ \\
Phi-3.5-mini & $2.704$ & $0.949$ $[0.79, 1.33]$ & $0.5\%$ & $2.813$ & --- & $0.926$ \\
\midrule
\multicolumn{7}{@{}l}{\emph{600 passages over all 31 books of BookMIA's \texttt{attack\_train} split}} \\
KL3M-520M    & $2.465$ & $1.014$ $[0.97, 1.13]$ & $0.0\%$ & $2.197$ & $1.102$ & $1.053$ \\
Pleias-1.2B  & $4.006$ & $1.314$ $[1.01, 1.59]$ & $0.4\%$ & $2.614$ & $0.895$ & $0.878$ \\
Phi-3.5-mini & $2.607$ & $0.992$ $[0.87, 1.37]$ & $0.0\%$ & $2.619$ & $0.949$ & $0.926$ \\
\bottomrule
\end{tabular}
\end{table}

\paragraph{The uncertified interval is the opening.}\label{app:opening}
On the $758$ passages under four safe models, $k_{\mathrm{crit}}/s$ has a median of $4.25$ to $6.10$, but
the running maximum in Eq.~\eqref{eq:rates} is attained at the \emph{first} token of the work in $87.7\%$
to $90.5\%$ of cases, because the bucket must pay for the opening out of one step's allowance. An adversary
who supplies one genuine prefix token removes it: the ratio falls to $1.32$--$1.48$ and stays there at $5$,
$10$ and $20$ supplied tokens. Per token instead of per character the whole-work median is $3.88$--$5.98$
and the ratios after a supplied prefix are unchanged ($1.35$--$1.41$). This is why oracle-prefix windows
recover $0.0965$ of a passage at $k=3$ against $0.0116$ for a single query (Table~\ref{tab:orders}).

\paragraph{Vacuity on a hundred books.}\label{app:bookmia}
The vacuity verdict needs no memoriser, only forward passes of the anchor, so we re-measured it on BookMIA
\citep{shi2024detecting} with the $758$ passages as a positive control (Table~\ref{tab:bookmia}). $65$ of
$4{,}935$ seen passages are surprising enough that a $600$-nat budget does not cover them, which is the
weaker reading registered rather than a replication; the unseen half and the control read $1.0000$. The
halves are not exchangeable under an anchor that saw neither (medians $6.0\%$ apart against a specified
$5\%$), so no arm treats them as a matched pair. A first pass scored vacuity as $k \ge s(x)$ per token and
read $0.326$ on the control: that criterion agrees with $K \ge S(x)$ only when the cap and the work have
the same length, and a deployment caps at $200$ tokens against a protected span near $60$.

\begin{table}[ht]
\centering\footnotesize
\caption{Vacuity on BookMIA at the published $K = 600$. The seen and unseen halves are not a matched pair
under an anchor that saw neither.}\label{tab:bookmia}
\setlength{\tabcolsep}{4pt}
\begin{tabular}{@{}lccccc@{}}
\toprule
corpus & passages & books & vacuous, $K = 600$ & median $S(x)$ & median $s(x)$/tok \\
\midrule
CopyBench (control) & $758$   & $16$ & $1.0000$ & $204.8$ & $3.197$ \\
BookMIA, seen       & $4{,}935$ & $50$ & $0.9868$ & $182.6$ & $3.114$ \\
BookMIA, unseen     & $4{,}935$ & $50$ & $1.0000$ & $192.1$ & $3.313$ \\
\bottomrule
\end{tabular}
\end{table}

\paragraph{A pair where memorisation is the only variable.}\label{app:second-anchor}
A vocabulary mismatch ($64{,}000$ against $128{,}256$ tokens) appears to rule out fusing a
second anchor with the risky model. The obstacle was an implementation check that the
embedding table have exactly $\mathrm{len}(\text{tokenizer})$ rows, which the Common Pile $7$B checkpoint
fails by padding $64{,}000$ tokens to $64{,}256$ rows; masking the untrained rows out of the solve removes
it. We fine-tune that $7$B model until it reproduces the protected passages ($0.915$ recall on its
training split, $0.000$ held out) and fuse it against the unmodified model it came from, so anchor and
risky model share every weight but those memorisation moved. Across $7$ budgets, both baselines, $5$
strategies and $31{,}738$ budgeted queries there is no invariant violation. The anchor alone reproduces
$0.0000$ under every strategy; at $k=20$ the decoder is the identity (single-query recall $0.7189$, equal
to the unconstrained model's to four decimals); recall is $0.000$ at every budget up to $k=1$, $0.048$ at
$k=3$ and $0.197$ at $k=5$, beginning near $s(x) = 2.39$ and far below $k_{\mathrm{crit}} = 11.32$.

\paragraph{A better anchor widens the margin.}\label{app:scaling}
Ten openly licensed safe models ($0.17$B to $7$B, three corpora) score the same $758$ passages and the same
$300$ ordinary risky-model generations, in nats per character so no shared vocabulary is needed. Both the
passages' rate $s(x)$ and the ordinary output's divergence $c_{\mathrm{use}}$ fall as the anchor improves,
and $c_{\mathrm{use}}$ falls faster, so the margin $s(x)/c_{\mathrm{use}}$ rises with capability within
every family, for all sixteen novels individually (Table~\ref{tab:marginspan}), and with training tokens at
fixed size ($4.63$ to $5.02$ from the $1$T to the $2$T checkpoint of one $7$B model). Pooled across
families it is not monotone in parameter count, because the families sit at different levels.

\begin{table}[ht]
\centering\footnotesize
\caption{The margin $s(x)/c_{\mathrm{use}}$ across the widest capability span in each family, paired over the
same sixteen novels.}\label{tab:marginspan}
\begin{tabular}{@{}llrrrr@{}}
\toprule
corpus & span & $\Delta c_{\mathrm{use}}$ & $\Delta s(x)$ & margin & sign test \\
\midrule
Common Corpus & $0.35 \to 3$B   & $-25.8\%$ & $-9.8\%$  & $3.45 \to 4.19$ & $16/16$, $p = 3.1\times10^{-5}$ \\
KL3M          & $0.17 \to 3.7$B & $-31.3\%$ & $-16.0\%$ & $1.94 \to 2.37$ & $16/16$, $p = 3.1\times10^{-5}$ \\
Common Pile   & $1.8 \to 7$B    & $-28.6\%$ & $-12.0\%$ & $4.07 \to 5.02$ & $16/16$, $p = 3.1\times10^{-5}$ \\
\bottomrule
\end{tabular}
\end{table}

\section{The four repairs, in full}\label{app:repairs}

\begin{table}[ht]
\centering\footnotesize
\caption{\textbf{The four repairs a per-token meter has left}, each measured at the audited pair with
bands fixed in advance. \citet{he2026anchored} ablate the first and last
themselves, as their $\infty$-R\'enyi decoder and their Global variant.}
\label{tab:repairs}
\setlength{\tabcolsep}{3pt}
\begin{tabular}{@{}lll@{}}
\toprule
repair & measured at the audited pair, $k=3$ unless stated & result \\
\midrule
sharper charge ($\alpha{:}\,1 {\to} 8$) & oracle recall $0.0965 \to 0.0012$; binds on $99.6\%$ of steps vs $0.4\%$ & the anchor \\
lower cap ($k = 0.5$) & no judge tells it from its anchor ($-1.03\sigma$, $+1.52\sigma$; $600$ pairs) & the anchor \\
better schedule & $3$ to $13\%$ more fidelity at $k \le 1$, under $1.4\%$ at $k=3$, same spend & percent \\
budget up front & $+0.0545$ $[+0.0315, +0.0775]$ at $64$ nats, selection $+0.1045$ at $\log 64$ & loses \\
\bottomrule
\end{tabular}
\end{table}

\paragraph{A sharper charge.}
Table~\ref{tab:orders} runs four R\'enyi orders at the authors' $k=3$. Every one publishes $K=3T$ and is
vacuous for $100\%$ of the $758$ protected passages. Raising the order cuts oracle recall
$0.0965 \to 0.0012$, an $80\times$ cut, by ceasing to serve the risky model: at $\alpha = 8$ the constraint
touches $99.6\%$ of ordinary steps. At matched \emph{utility} rather than matched budget
(Table~\ref{tab:orderlaw}, fidelity to $p_r$ on the geodesic as the utility axis, no judge) the higher order
usually helps at $k=1$, by a factor of $1.0$ on one pair and $2.5\times10^{7}$ on another, and on four pairs
$\alpha{=}8$ makes the protected tokens \emph{more} likely, by up to $1.3\times10^{6}$ on the mechanism's own
configuration; at $k=3$ matching forces every order past its vacuity threshold, and on nine of twelve pairs
the ranking reverses.\label{app:matched}

\begin{table}[h]
\centering\footnotesize
\caption{Four decoders at $k=3$, all publishing $K=3T$ and all vacuous for every protected passage. Attack
columns: mean near-verbatim recall over $100$ passages from the memorising $8$B model; price columns: $150$
ordinary prompts.}
\label{tab:orders}
\setlength{\tabcolsep}{5pt}
\begin{tabular}{@{}lccccc@{}}
\toprule
& \multicolumn{2}{c}{attack recall} & \multicolumn{3}{c}{price on ordinary traffic} \\
\cmidrule(lr){2-3}\cmidrule(lr){4-6}
order & single query & oracle windows & risky unchanged & steps touched & distinct-3 \\
\midrule
$\alpha = 1$ (KL) & $0.0116$ & $0.0965$ & $99.3\%$ & $0.4\%$  & $0.9910$ \\
$\alpha = 2$      & $0.0010$ & $0.0545$ & $97.2\%$ & $2.5\%$  & $0.9924$ \\
$\alpha = 4$      & $0.0000$ & $0.0042$ & $9.4\%$  & $90.3\%$ & $0.9850$ \\
$\alpha = 8$      & $0.0000$ & $0.0012$ & $0.08\%$ & $99.6\%$ & $0.9872$ \\
\bottomrule
\end{tabular}
\end{table}

\begin{table}[ht]
\centering\footnotesize
\caption{Factor by which an exact $50$-token window becomes \emph{less} likely under order $\alpha$, at the
budget $k_\alpha$ where it buys exactly the fidelity $\alpha{=}1$ buys at the published $k$; below $1$ the
order leaks more at equal utility. Twelve pairs in seven families, \texttt{bfloat16} with a
\texttt{float32} control per pair; $s(x)$ in nats per token.}\label{tab:orderlaw}
\setlength{\tabcolsep}{4pt}
\begin{tabular}{@{}lccccccc@{}}
\toprule
& & \multicolumn{3}{c}{published $k = 1$} & \multicolumn{3}{c}{published $k = 3$} \\
\cmidrule(lr){3-5}\cmidrule(lr){6-8}
pair & $s(x)$ & $\alpha{=}2$ & $\alpha{=}4$ & $\alpha{=}8$ & $\alpha{=}2$ & $\alpha{=}4$ & $\alpha{=}8$ \\
\midrule
Pleias 350M    & $3.50$ & $5.0{\cdot}10^{4}$ & $2.5{\cdot}10^{7}$ & $1.4{\cdot}10^{7}$ & $67.5$ & $28.7$ & $2.4$ \\
Pleias 1.2B    & $3.12$ & $3.6{\cdot}10^{4}$ & $1.3{\cdot}10^{7}$ & $1.9{\cdot}10^{7}$ & $67.3$ & $113$ & $27.2$ \\
Comma 7B       & $2.32$ & $4.3{\cdot}10^{3}$ & $1.9{\cdot}10^{5}$ & $9.0{\cdot}10^{3}$ & $5.6$ & $2.2$ & $0.26$ \\
KL3M 170M      & $2.61$ & $3.3{\cdot}10^{3}$ & $1.8{\cdot}10^{4}$ & $455$ & $4.1$ & $0.37$ & $0.12$ \\
TinyComma 1.8B & $3.15$ & $3.0{\cdot}10^{3}$ & $5.3$ & $8.0{\cdot}10^{-7}$ & $0.67$ & $1.6{\cdot}10^{-3}$ & $1.3{\cdot}10^{-4}$ \\
KL3M 3.7B      & $2.20$ & $1.4{\cdot}10^{3}$ & $8.6{\cdot}10^{3}$ & $909$ & $1.2$ & $0.37$ & $0.18$ \\
KL3M 520M      & $2.42$ & $1.3{\cdot}10^{3}$ & $1.0{\cdot}10^{3}$ & $1.5$ & $1.4$ & $0.073$ & $7.3{\cdot}10^{-3}$ \\
Llama-3.2 3B   & $2.81$ & $1.1{\cdot}10^{3}$ & $828$ & $1.6$ & $2.8$ & $0.30$ & $0.048$ \\
KL3M 1.7B      & $2.25$ & $638$ & $7.4{\cdot}10^{4}$ & $4.4{\cdot}10^{5}$ & $46.4$ & $175$ & $311$ \\
Llama-3.2 1B   & $2.78$ & $158$ & $9.4$ & $8.7{\cdot}10^{-3}$ & $0.84$ & $0.048$ & $7.8{\cdot}10^{-3}$ \\
Qwen2.5 7B     & $2.62$ & $71.8$ & $1.0$ & $1.7{\cdot}10^{-4}$ & $0.47$ & $0.032$ & $3.7{\cdot}10^{-3}$ \\
Phi-3.5 mini   & $2.71$ & $47.5$ & $7.8$ & $0.078$ & $0.15$ & $0.011$ & $2.7{\cdot}10^{-3}$ \\
\bottomrule
\end{tabular}
\end{table}

\paragraph{A lower cap.}
At $k=0.5$ neither outside judge separates the meter from its own safe model over $600$ comparisons each
($-1.03\sigma$, $+1.52\sigma$); both separate only at $k \ge 10$ ($-6.97\sigma$, $-3.29\sigma$).

\paragraph{A better schedule.}
On the geodesic the charge and the fidelity bought are closed forms in $\psi(u) = \log Z(u)$,
\begin{equation}
C(\theta) = \theta\psi'(\theta) - \psi(\theta), \quad C'(\theta) = \theta\psi''(\theta), \qquad
G(\theta) = \theta m - \psi(\theta), \quad G'(\theta) = m - \psi'(\theta),
\label{eq:charge-gain}
\end{equation}
with $m = \psi'(1)$, so $C'(0) = 0$ while $G'(0)$ is the Jeffreys divergence between the two models
\citep{monteiropaes2026limits}: the first nat bought at a step is worth the most. The bucket takes whatever
price $G'/C'$ its allowance lands on; the best allocation of the same total equalises it. Computed from two
teacher-forced passes over $30$ targets of three kinds (memorised passages, a held-out novel, the risky
model's own samples), the optimal reallocation buys $3$ to $13\%$ more fidelity at $k \le 1$ and under
$1.4\%$ at $k=3$, against an offline optimum no causal scheduler can beat \citep{tomasi2026primaldual}; the
three target kinds agree to within two points. The bucket already captures $46.7$--$54.5\%$ of the
$\theta = 1$ fidelity at $k=0.25$ and $92.7$--$98.9\%$ at $k=3$, so three orders of magnitude are not there
to recover.

\paragraph{The budget granted up front.}
The one causal policy Proposition~\ref{prop:sparse} positively permits spends a budget granted at the
start. Judged order-averaged against the committed opponent it gains $-0.0015$ $[-0.0200,+0.0170]$ at
$\log 8$ nats, $+0.0215$ $[+0.0015,+0.0415]$ at $\log 64$ and $+0.0545$ $[+0.0315,+0.0775]$ at $64$ nats:
the causal horn is not empty, but at matched budget selection leads by $+0.083$ $[+0.052,+0.1135]$ and no
budget on the grid reaches selection's $+0.1045$ at $\log 64$. Choosing \emph{where} to spend does not help:
a rule that spends at step $t$ only if the full-tilt demand reaches $\tau$ nats reserves as designed (median
spending step $3 \to 17$) and gains $+0.0215$, $+0.0150$, $+0.0110$, $+0.0045$ and $+0.0070$ at
$\tau = 0,1,2,4,8$, falling except for a tick at the last, because a higher bar releases the budget less
often ($0\%$ to $86.7\%$ of trajectories never spend) \emph{and} buys less when it does. An earlier arm,
judged single-order and not compared with these, gave the front-loaded rule $-0.008$ $[-0.056, +0.041]$
at $\log 8 = 2.0794$ nats against selection's $+0.054$ at the same budget.

\paragraph{How far each mechanism sits from Lemma~\ref{thm:nfl}'s frontier.}\label{app:frontierratio}
Read on the repaired text, order-averaged, with each arm's own control as the safe law of $U$, selection at
$n=64$ sits $35.4\times$ the frontier on its $3.175$-nat bound and the meter at $k=10$ $7{,}384\times$ on its
$171.3$ nats spent (\texttt{results/\allowbreak frontier\_\allowbreak distance\_\allowbreak h2h\_\allowbreak deecho.csv});
the $32\times$ and $11{,}646\times$ an earlier version gave were read on text that carried the prompt's tail.
Proposition~\ref{prop:imitation}'s conservative ratio needs $\pi$, the chance that the anchor attains the
top utility. On Figure~\ref{fig:selection}(d)'s instrument the anchor wins both presentation orders on
$2.2\%$ of prompts, $4.0\%$ of questions, $2.0\%$ of biographies and none of the stories, whose best anchor
level is $0.75$: $\log(1/\pi)$ is $3.82$ pooled, $3.22$ and $3.91$ for the first two classes, and infinite for
stories, where the bound says nothing. The meter's $171.3$ nats are at least $44.9\times$ what winning every
comparison needs, $53.2\times$ and $43.8\times$ by class. Because $\Lambda^*_s(u) \le \log(1/\pi)$ for any
utility the decoder does not lower, the proposition's second clause follows from its first, a spend of at
least $\gamma T$; it restates the spend in units of what the utility requires rather than adding a bound. The single-order pass this paragraph once used
gave $\log(1/\pi) = 1.30$ and $132\times$: there a win in the one order shown came on over a quarter of
prompts, and a win in both orders is the stricter event. Every figure here pools the prompts, so it bounds the prompt-averaged
divergence, which each per-prompt certificate implies but does not equal; one anchor draw per prompt
cannot estimate a per-prompt $\pi$. At the level each arm reaches rather than the top one,
$\Lambda^*_s$ is $0.00255$, $0.0115$ and $0.0232$ nats at $k=1$, $3$ and $10$, a ratio of $52{,}864$,
$14{,}357$ and $7{,}384$; at $k=0.5$ the meter gains nothing and the ratio is unbounded
(\texttt{results/\allowbreak frontier\_\allowbreak ratio.csv}).

\paragraph{Two anchors closer to the risky model.}\label{app:frontier3}
Neither anchor here is a safe model, so no certificate, leakage or $s(x)$ number is taken from either; the pairs test whether the reversal is the mechanism's or the anchor's. A token-level meter needs a shared vocabulary, and no other openly licensed anchor has one with this risky model. At Llama-3.2-1B the meter is useful,
$+0.051$, $+0.056$ and $+0.055$ under judge~B at $k=1,3,20$ for $88.1$ to $123.4$ nats, and selection at
$n=8$ gains more, $+0.076$ $[+0.031, +0.122]$ and $+0.154$ $[+0.104, +0.200]$ on two judges, for $1.204$
nats. At Llama-3.2-3B-Instruct, the closest meterable anchor and the one the protocol named the hardest
case, the meter resolves at no budget on either judge (all eight point estimates negative, spends $54.8$ to
$60.6$ nats over a $40\times$ range of caps) while selection gains $+0.150$ $[+0.106, +0.194]$ and $+0.273$
$[+0.227, +0.323]$. A close anchor flattens the meter onto the anchor. The arms each pass holds twice (the
meter's $k=0$ and selection's first draw) differ by at most $0.034$, and the same size appears independently
at Comma-7B ($0.039$), so \textbf{the cross-pass floor on a judged gain is about $0.04$}, and gains from
different passes are never compared here.

\section{Limitations in full}\label{app:limitations}

\paragraph{One protected corpus.}
Every extraction measurement runs on sixteen English genre novels from CopyBench and BookMIA, plus the
four French and German works of the multilingual arm. The margin's sign test, $p = 3.1\times10^{-5}$, is
the smallest value sixteen paired items can produce: it says the trend holds for every novel we have, not
that it was independently replicated. Non-literal reproduction and other genres are untested. On two
further corpora the level of the onset survives, every onset within about $1.3\times$ of $s(x)$, and the
ordering of pairs does not (Appendix~\ref{app:collapse}), so an onset ratio is reportable to its order of
magnitude and no more finely. Varying only a memoriser's fine-tuning length moves one pair's ratio from
$0.8756$ to $1.3477$, $1.64\times$ the nine-pair spread; re-seeding moves it by $0.033$, $0.066$ and $0.080$
where the fine-tune converged and by $0.260$ over three seeds and $0.347$ over four where it did not; and
neither intervention that would make convergence a cause could be built (a loss floor of $0.0222$ against
a $0.02$ stop-loss; a memoriser that reproduces nothing after a $17\%$ change in learning rate). The
nine-pair table's memorisers are matched on none of training length (a factor of $5.1$ in sampled $k=-1$
recall), batch size, accumulation, stopping rule or convergence. The seed-word gradient $\rho = -0.958$ keeps
its sign in every re-draw under the largest re-seed noise measured and $p<0.05$ in $92.3\%$ of them, at a
median of $-0.849$: a rank ordering, not a coefficient.

\paragraph{More passages do not narrow the onset interval.}
The bootstrap interval on onset\,/\,$s(x)$ has a mean width of $0.285$ across the pairs at $100$ passages
each; on one pair under one metric it is $0.219$ at $n=100$ and $0.215$ at $n=458$, where independent
sampling noise predicts $0.102$. The variance is between works, not windows, so several registered bands
about $0.1$ wide land undecided even where the shift is large and in the predicted direction. More distinct
works would fix it; more windows will not.

\paragraph{The onset's range has one cause we can name and one we cannot.}
$s(x)$ spans $1.87\times$ across the nine pairs, short of the range a data collapse normally rests on. Both
KL3M pairs exclude $1$ and differ from the rest in tokenizer, target length in tokens and, for one, memoriser
strength; a pair built to remove the last memorises thoroughly ($s_r/s_s = 0.09$ against $0.35$) and still
reads $1.053$. The tokenizer remains confounded with what it moves at once: the adversary's seed ($7.5$
words against $13.9$--$15.0$), the decode steps before the passage ($445$ against $197$--$230$) and the
target length.

\paragraph{Lemma~\ref{thm:nfl} is a converse.}
It lower-bounds the budget any policy needs; we do not bound the gap between that optimum and a causal
token-bucket decoder, which is a factor of $5$ to $32$ between the divergence on the risky model's own
rollout and the divergence paid on a budgeted one.

\paragraph{What selection anchoring cannot do.}
\emph{The support ceiling.} $q(y) \le n\,p_s(y)$ is also a quality ceiling: no $n$ produces text outside the
anchor's support. At $n=8$ on the judged workload it does not bind: the oracle among the same eight
candidates reaches $u = 0.807$ where the deployed scorers reach $0.52$. Across anchors it is confirmed on
AlpacaEval: the audited anchor gains $+0.074$ $[+0.042, +0.108]$ and $+0.020$ $[-0.011, +0.052]$ under two
judges, Comma-7B $+0.075$ $[+0.044, +0.106]$ and $+0.152$ $[+0.119, +0.185]$; MT-Bench's $80$ prompts
resolve nothing at a paired half-width of $0.10$. Within one anchor, across AlpacaEval's five sources and
two MT-Bench families, the correlations of $-0.79$ and $-0.70$ between gain and control level are not
separable from a null in which selection does nothing ($-0.36 \pm 0.34$ and $-0.34 \pm 0.35$, $P = 0.09$
and $0.17$).

\emph{Empty drafts.} The audited anchor's first draw is empty on $14.6\%$ of prompts in its $n=8$ arm and
$12.8\%$ in its $n=64$ arm, against $0.0\%$ at both Pleias anchors, $0.2\%$ at KL3M-1.7B, $18.4\%$ at
Comma-7B and $31.6\%$ at Comma-7B-1T, all three Comma anchors above the $5\%$ the admission criterion
allows, recorded rather than exempted. On the $427$ prompts whose $n=1$ draw is non-empty the audited
anchor's $n=8$ gain is $+0.079$ $[+0.028, +0.128]$ against $+0.073$ on all $500$, and every interval that
excluded zero still does but one, Comma-1T's under judge~B, whose lower end was $+0.001$.

\emph{Overoptimisation.} To $n=512$ no judged ladder turns over, against registered bands that allowed it
to; under a task metric the $7$B reward turns over on TriviaQA and CoTaEval (Table~\ref{tab:scorerladder}),
and nothing in Proposition~\ref{prop:selection} prevents it on a judged workload at larger $n$.

\emph{The scorer is a second model and a stated requirement.} It cannot break the certificate but can
degrade utility arbitrarily, so served utility and the certificate must be monitored separately. Every
reward turn-over here is measured at \texttt{Qwen2.5-7B-Instruct} and none survives a larger scorer, so on
knowledge-intensive tasks selection under a pointwise reward carries a scorer-size requirement, and a
deployer who ignores it can be worse off at every $n$ than serving the anchor once. On the judged workload
the two larger scorers lift nothing ($-0.0285$ and $-0.0085$ against the $7.6$B rung), and no latency arm
ran above $7$B.

\emph{The bound is relative to $p_s$.} A safe model that has seen the work carries its contamination
through, multiplied by $n$ (Table~\ref{tab:contam}); the $0.0000$ of Section~\ref{sec:experiments} is $n$
times a base rate that is zero because the anchor never saw these novels.

\emph{No frontier-quality anchor.} Comma-7B is the largest openly licensed anchor to pass our vetting
check. We do not know where a vetted anchor of frontier quality would come from: at web scale it is hard
to show that no protected work entered through derived copies, and an anchor distilled from the risky
model would inherit its risk. Every utility result here is therefore for anchors far below frontier scale,
and none covers long agentic rollouts.

\paragraph{What the certificates do not cover.}
The per-work audit bounds a near-verbatim event at $95\%$ confidence for one fixed prompt distribution, the
work's own prefix, and only given zero hits among the clean anchor draws; other prompts need their own
audit. Pooled adaptive queries are covered only by $m\log n$ composition, which at $n=64$ stays informative
about a window for $38$ queries. A vetted scorer removes that horizon only as an operational condition that
the certificate cannot verify (Section~\ref{sec:compose}, Appendix~\ref{app:perwork}). In installments the
whole-output bound $B\log n$ holds, but a window is certified only conditional on the served text up to the
last block boundary at or before its start. An earlier version of this work stated it conditional on all
preceding text, which is false inside a block (Proposition~\ref{prop:installments},
Remark~\ref{rem:installments}). The windowed meter's per-span certificate is of a different type, so
comparisons between it and installments at equal nats per span are across certificate types.

\paragraph{The metered decoder as its authors run it.}
\citet{he2026anchored} pair base risky models (\texttt{Llama-3.1-70B}, \texttt{Qwen2.5-72B},
\texttt{Llama-4-Scout}) with openly licensed anchors, decode books at temperature $0.7$ with a repetition
penalty of $1.1$, fuse mismatched vocabularies through AnchoredByte, and score utility by Prometheus-2
fluency and FActScore precision. We ran their token-level pair at temperature $1.0$ with no penalty, the
$8$B-Instruct without its chat template (and, separately, with it), AnchoredByte only at
their recommended pair, and their two metrics with open models in place of \texttt{gpt-4.1-mini}. None of
this changes a certificate or a leakage number; each can change a judged one. No human rated anything and no
lawyer assessed any output.

\paragraph{Two judges, one workload.}
Two judges at $600$ comparisons per arm agree on the sign everywhere and on neither endpoint's magnitude:
at $k=1$ one reads $-3.66\sigma$ against the anchor and the other $+0.11\sigma$, at $k=10$ $-6.97\sigma$
against $-3.29\sigma$. Served the unconstrained risky model, the first separates it at $-6.09\sigma$ and the
second at $-2.49\sigma$, and an earlier run of the same $500$ comparisons gave $-5.47\sigma$ and
$-1.49\sigma$, so about a $\sigma$ of run-to-run drift is the floor beneath which no separation here should
be read.

\paragraph{The two leakage measurements are not on one axis.}
The meter's onset and selection's zero come from two pipelines whose memoriser-alone rows disagree ($0.4921$
against $0.3925$; at a matched $300$ tokens $0.4921$ against $0.4428$, a residual of $0.0493$ against a
registered gate of $0.02$). The residual has two causes the arm cannot separate: changing the token budget changes the draw as well as the length, and the meter decodes each passage's own target length where selection decodes a flat $300$. No joint table is built, every leakage number is read against a control from
its own pipeline, and the comparison between the mechanisms is about what they spend, not what they leak.

\section{Related work in full}\label{app:related}
Table~\ref{tab:related} sorts the defences by what their guarantee is about.

\begin{table}[ht]
\centering\footnotesize
\caption{Copyright defences at or near inference time, by what the guarantee is about. The first three
rows bound the served text's law against a model (CPR does so for images); the others guarantee something
about a named string, an $n$-gram list, an index, a datastore or a takedown list.}
\label{tab:related}
\begin{tabular}{@{}p{0.27\linewidth}p{0.45\linewidth}p{0.22\linewidth}@{}}
\toprule
defence & what the guarantee is about & measured here \\
\midrule
anchored decoding \citep{he2026anchored} & the served law against a safe model, charged per token ($K = kT$) & audited throughout \\
selection anchoring (this paper) & the served law against a safe model, charged once per draw (pathwise, $\log n$) & Sections~\ref{sec:method} and~\ref{sec:experiments} \\
CP-$\Delta$, CP-$k$ \citep{vyas2023provable}, CP-Fuse \citep{abad2025cpfuse} & the served law against shard models, per query (CP-$k$ once per output), given a disjoint split & leakage (below); CP-$k$'s rule, run with one safe model (Appendix~\ref{app:cpk}) \\
$n$-gram blocklist, MemFree \citep{ippolito2023preventing} & listed $n$-grams are never completed; no bound on the served law & leakage and utility (below) \\
TokenSwap \citep{prashant2025tokenswap} & a memorised string's probability decays in its length & leakage and utility (below) \\
TRBS \citep{fu2026duplicate} & beams reranked against an index of the corpus & not measured (a code method) \\
BloomScrub \citep{zhang2025bloomscrub} & takedowns screened by a Bloom filter, with abstention & not measured \\
SILO \citep{min2024silo} & legal risk isolated in a nonparametric datastore & not measured \\
CPR \citep{golatkar2024cpr} & near-access-freeness through retrieval, for diffusion models & not measured \\
\bottomrule
\end{tabular}
\end{table}

\paragraph{Access-free guarantees and their discontents.}
\citet{cohen2025blameless} show that per-query access-freeness does not compose across queries, and
\citet{cohen2026barriers} prove barriers for counterfactual credit attribution whose non-composition is
autoregressive: imposing it on the next-token predictor does not make the sequence-level model satisfy it.
Ours is the same shape for a different quantity, priced in nats: a per-step charge honoured at every step
leaves the event $\{\text{output}=x\}$ unprotected once the budget reaches $S(x)$. \citet{segal2026provably}
report that CP-style guarantees can be numerically uninformative in realistic fine-tuning and demonstrate
extraction against NAF-protected models; their diagnosis is that memorisation puts sharp spikes on a few
tokens, and their repair charges relative rather than raw probabilities, a change in \emph{what} is
metered. Proposition~\ref{prop:threshold} closes a different lever: for every order of the same per-step
charge the certificate becomes vacuous at the same budget, so a stronger charge buys precision below the
threshold and never moves it. Neither result subsumes the other. That a KL bound does not imply a
worst-case event bound is classical \citep{sason2016fdivergence,vanerven2014renyi}; what we add is which
budgets survive the requirement that the certificate be informative about a specific work and the
mechanism remain usable.

\paragraph{Consensus sampling, the closest neighbour.}\label{app:kalai}
\citet{kalai2026consensus} aggregate $k$ generative models, an unknown subset of them safe, and serve a
sample only where enough of them assign it comparable probability, abstaining otherwise. The instinct is
ours, certifying the output relative to models that did not misbehave, and the premises are opposite: they
need several models and no knowledge of which are safe and pay by abstaining; we need one named safe model,
never abstain, and pay in draws. Both certificates inherit what their safe models hold, and ours is
checkable in advance by vetting (Appendix~\ref{app:vetting}). A deployer with a vetted anchor and several
risky models could run both.

\paragraph{TokenSwap, measured: cheaper than selection and just as suppressive.}\label{app:tokenswap}
TokenSwap \citep{prashant2025tokenswap} substitutes the probabilities of high-frequency grammatical tokens
from a small auxiliary model and proves that a memorised sequence's probability decays exponentially in its
length; TRBS \citep{fu2026duplicate} reranks beams against an index of a code corpus. Both guarantee
something about a catalogue supplied, not about the served law relative to a safe model. We reimplemented
TokenSwap from its Algorithm~1 with the published $110$-word $G$ on our $100$ protected passages and $850$
ordinary prompts. It suppresses completely (recall $0.0000$, ROUGE-L $\ge 0.5$ on $0$ of $100$ and $\ge 0.3$
on $0$, against $45$ and $69$ for the memoriser) and, judged against its own rule-off control, costs
$-0.1040$ $[-0.1255, -0.0820]$, where selection at $n=64$ costs $-0.1655$ $[-0.2000, -0.1320]$: TokenSwap is
ahead of selection by $+0.0615$ $[+0.0285, +0.0950]$. The rule-off control is the risky model served
through its chat template, judged against the committed templateless opponent, so selection's cost here
is its distance from that model, which gains $+0.272$ over the anchor to selection's $+0.1065$: less than
half the gap, as in Table~\ref{tab:served}'s chat-template block. \textbf{The mechanism this paper is about is the most
expensive of the three and the only one that publishes a bound on the served law.} The margin depends on
the auxiliary: at \texttt{DistilGPT-2}, the $82$M model its authors use, the two tie ($-0.0325$
$[-0.0645, +0.0005]$). And the suppression is fragile in a way $|G|$ does not reveal: with
\texttt{KL3M-170m} as the auxiliary recall returns to $0.2113$, with $23$ of $100$ passages over ROUGE-L
$0.5$, because that tokenizer keeps $171$ of $431$ $G$ token ids but $1.0\%$ of the probability mass, so the
swap fires on $1.8\%$ of steps instead of $25$--$29\%$. All four candidate auxiliaries read $0.0000$ on the
protected passages alone.

\paragraph{The blocklist, measured.}\label{app:blocklist}
Refusing any continuation that completes an $n$-gram of a listed work \citep{ippolito2023preventing} is the
incumbent a deployer would otherwise run. \textbf{It refuted the argument we built it to support}: collateral
on ordinary text does not grow with the catalogue. At $n=10$ the fraction of ordinary tokens blocked is
$0.000\%$ at $758$ passages, $0.140\%$ at $4{,}935$ and $0.140\%$ again at $9{,}870$, where the index has
doubled to $2.24$M $n$-grams ($0.000\%$, $0.147\%$, $0.147\%$ at $n=8$). Run as an exact rejection-sampling
decoder against the memoriser on its own training split, it binds on $1.035\%$ of steps and takes recall
from $0.4192$ to $0.0201$; on ordinary prompts it fires on $0$ of $850$, so its judged utility is the
unconstrained model's exactly ($+0.272$ $[+0.247, +0.2965]$ over the anchor with and without it, the
chat-templated model's gain). We predicted it would leak
the non-literal event and it did not: $0$ of $100$ passages reach ROUGE-L $\ge 0.5$ under the rule, against
$52$ without it. On a \emph{listed} work the blocklist costs no utility, which selection cannot approach,
and it does not suppress more ($0.0201$ against its own $0.4192$; selection $0.0000$ against $0.3925$). The
difference is in kind: the blocklist must enumerate the works and is silent on one nobody listed, and it
publishes no number about the served law, so nothing about it composes across queries, where $\log n$ adds
exactly. Selection costs $21.8\times$ a metered decode in a server that runs only the anchor
(Appendix~\ref{app:anchoronly}); a blocklist costs a hash lookup. A deployer can run both.

\paragraph{CP-Fuse, measured on leakage.}\label{app:cpfuse}
We rebuilt CP-Fuse from \citet{abad2025cpfuse}'s description as two LoRA copies of the
risky model, each fine-tuned on one of two disjoint halves of the protected passages and fused by the
balancing rule. Each shard reproduces its own half (recall $0.6905$ and $0.6965$) and not the other's
($0.0000$); the fusion reproduces neither, $0.0000$ from a single query and $0.0209$ and $0.0199$ under
oracle-prefix windows (\texttt{results/\allowbreak cpfuse\_\allowbreak audit\_\allowbreak rebuild.csv}). A
$50/50$ mixture without the balancing already holds single queries to $0.0295$ and $0.0000$, so most of the
suppression is the disjoint split. An earlier build with its own fine-tunes has the same shape,
$0.7538$ and $0.7988$ on the own shard against $0.0000$ fused
(\texttt{results/\allowbreak cpfuse\_\allowbreak audit.csv}). The utility half was fixed in advance and withdrawn before it ran: on
ordinary prompts CP-Fuse is the risky model with a device attached, and judged against that model a win is
arithmetic, not evidence. Its guarantee holds per query given a split only the trainer can attest, and
per-query access-freeness does not compose \citep{cohen2025blameless}.

The accounting tools we use
\citep{feldman2020renyifilter,whitehouse2023fully,rogers2016odometers,loynes1962stability,cruz1991calculus}
and the extraction protocol \citep{carlini2021extracting,carlini2023quantifying,cooper2025extracting} are
standard.

\paragraph{Other defences.}
BloomScrub certifies takedowns by Bloom-filter screening with abstention \citep{zhang2025bloomscrub}; SILO
isolates legal risk in a nonparametric datastore \citep{min2024silo}; others refuse \citep{liu2024shield} or
randomise the decoder \citep{chen2024randomization}; for diffusion models retrieval carries a NAF guarantee
\citep{golatkar2024cpr} in response to replication in image models \citep{somepalli2023diffusion}.
\citet{alshehyari2026copyshield} benchmark output-, behavioural- and representation-level defences; the level
their benchmark omits is the certified one, where a low measured leakage and a vacuous certificate can
coexist. VA3 \citep{li2024va3} amplifies a probabilistic copyright guarantee for text-to-image models through
repeated interaction: choosing among $m$ samples raises a per-sample probability $p$ to $1-(1-p)^m \le mp$,
the attacker's side of Proposition~\ref{prop:selection}. Training-time mitigations act on the weights
instead: approximate unlearning \citep{eldan2023whos}, whose completeness and robustness are studied by
\citet{shi2025muse}, \citet{deeb2024unlearning} and \citet{lucki2025adversarial}. They are complementary to
an inference-time certificate, and we measure none of them.

\paragraph{Self-consistency.}
Self-consistency \citep{wang2023selfconsistency} serves the majority answer over $n$ sampled paths; a mode
is a score, so it satisfies $q(y) \le n\,p_s(y)$ and carries a pathwise certificate of exactly $\log n$
against the model it samples. As deployed it samples web-trained models, relative to which $\log n$
certifies nothing about copyright; over a safe model the same inequality is a copyright guarantee its users
already have. The best-of-$n$ literature prices $\log n$ as a \emph{cost}, the budget spent before reward
overoptimisation \citep{gao2023scaling,beirami2025bestofn} or the one an attacker spends to jailbreak
\citep{hughes2024bestofn}; we read it as a guarantee, compare it with a metered decoder on one divergence
axis, and measure it. The mechanism is theirs. So is soft best-of-$n$ \citep{verdun2025soft}, which draws the
served index from a softmax of the rewards and so interpolates between the anchor and best-of-$n$; it serves
one of the draws, so Proposition~\ref{prop:selection} covers it at the same $\log n$.

\end{document}